\documentclass{ieeeaccess}
\usepackage{amsmath,amssymb,amsfonts,subcaption}
\usepackage{ulem,url}
\usepackage{graphicx}
\usepackage{textcomp}
\newcommand{\corr}[1]{\left\langle#1\right\rangle} 
\usepackage[T1]{fontenc}
\usepackage[utf8]{inputenc}
\usepackage{cite}

\def\corr#1{\left\langle #1 \right\rangle}
\newcommand{\tr}{\operatorname{tr}}

\newcommand{\be}{\begin{equation}}
\newcommand{\bea}{\begin{eqnarray}}
\newcommand{\ba}{\begin{equation}\begin{aligned}}

\newcommand{\ee}{\end{equation}}
\newcommand{\eea}{\end{eqnarray}}
\newcommand{\ea}{\end{aligned}\end{equation}}

\usepackage{fancyhdr}

\doi{10.1109/ACCESS.2020.3000901}
\def\BibTeX{{\rm B\kern-.05em{\sc i\kern-.025em b}\kern-.08em
    T\kern-.1667em\lower.7ex\hbox{E}\kern-.125emX}}
    
\begin{document}

\history{This article has been accepted for publication in a future issue of this journal, but has not been fully edited. Content may change prior to final publication. Citation information: DOI 10.1109/ACCESS.2020.3000901, IEEE Access}

\title{Is Deep Learning a Renormalization Group Flow?}
\author{\uppercase{Ellen de Mello Koch}\authorrefmark{1},
\uppercase{Robert de Mello Koch\authorrefmark{2}, and Ling Cheng}.\authorrefmark{1},}
\address[1]{School of Electrical and Information Engineering, University of the Witwatersrand, Wits, 2050, South Africa}
\address[2]{School of Physics and Telecommunication Engineering, South China Normal University, Guangzhou 510006, China and National Institute for Theoretical Physics and the School of Physics and Mandelstam Institute for Theoretical Physics, University of the Witwatersrand, Wits, 2050, South Africa.}
\tfootnote{}


\corresp{Corresponding author: Robert de Mello Koch (e-mail: robert@neo.phys.wits.ac.za).}

\begin{abstract}
    Although there has been a rapid development of practical applications, theoretical explanations of deep learning are in 
    their infancy. 
    Deep learning performs a sophisticated coarse graining.
    Since coarse graining is a key ingredient of the renormalization group (RG), RG may provide a useful theoretical framework 
    directly relevant to deep learning.
    In this study we pursue this possibility.
    A statistical mechanics model for a magnet, the Ising model, is used to train an unsupervised restricted Boltzmann machine (RBM). 
    The patterns generated by the trained RBM are compared to the configurations generated through an RG treatment of the
    Ising model.
    { Although we are motivated by the connection between deep learning and RG flow, in this study we focus
    mainly on comparing a single layer of a deep network to a single step in the RG flow.}
    We argue that correlation functions between hidden and visible neurons are capable of diagnosing RG-like coarse graining.
    Numerical experiments show the presence of RG-like patterns in correlators computed using the trained RBMs.
    The observables we consider are also able to exhibit important differences between RG and deep learning.
    \end{abstract}


\maketitle
\section{Introduction}\label{Intro}
The power of machine learning and artificial intelligence is established: these 
are powerful methods that already outperform humans in specific tasks \cite{jordan2015machine,deng2014deep,bengio2009learning,hinton2006reducing}. 
Much of the research carried out in machine learning is of an applied nature.
It establishes the practical utility of the method but does not construct an understanding of how deep learning works or even if such an understanding is possible \cite{le2010deep,le2008representational,bengio2007scaling,paul2014does,bengio2012unsupervised,zhang2016understanding}.
Consequently, deep learning remains an impressive but mysterious black box.
A possible starting point for a theoretical treatment suggests that deep learning is a form of coarse 
graining \cite{bengio2009learning,bengio2007greedy,lin2017does}.
Since there are more input than output neurons this is almost certainly true.
The real question is then if this is a useful observation, one that might shed light on how deep learning works.
This is the question we take up in this paper.

We argue that understanding deep learning as a form of coarse graining is a useful observation, and make the case
by adopting and adapting several ideas from theoretical physics.
Specifically, in theoretical physics there is a sound framework to carry out coarse graining, known as the 
renormalization group (RG) \cite{kogut1974renormalization}. 
RG provides a systematic way to construct a theory describing large scale features from an underlying microscopic 
description, which can be understood as recognizing sophisticated emergent patterns, a routine achievement of deep
learning.
Further, RG is applicable to field theories, that is, to systems with a large number of degrees of freedom so that it 
seems that RG is well positioned to deal with massive data sets.
Finally, the way in which RG works is, in contrast to deep learning, well understood and can be described in
precise mathematical language.
These features suggest that RG may provide a useful framework in which to describe deep learning
and attempts to argue that this is the case have been made in \cite{mehta2014exact,iso2018scale,funai2018thermodynamics,beny2013deep}.
We focus on unsupervised learning by a restricted Boltzmann machine (RBM). 

Two distinct possible connections to RG have been attempted, both relevant to our study.
The first \cite{mehta2014exact} is an attempt to link deep learning to the RG flow. 
The RG flow is a smooth process during which degrees of freedom are continuously averaged out, so that we flow
from the initial microscopic description to the final macroscopic description.
In deep learning one stacks layers of networks to obtain a deep network. 
The proposed connection of \cite{mehta2014exact} suggests that each layer in the stack performs a small step along the RG flow\footnote{For a critical discussion of this proposal, see \cite{lin2017does,koch2018mutual}}.
We contribute to this discussion by developing quantitative tools with which this proposal can be explored with precision.
The basic objects that appear in our analysis are correlation functions between the visible and the hidden neurons.
This allows us to decode the mechanics of the RBM's pattern generation and to compare it to what the RG is doing. 
Although there are important differences, our results indicate remarkable similarities between how the RBM
and RG achieve their results.
The second approach \cite{iso2018scale,funai2018thermodynamics} builds an RBM flow using the weight 
matrix of the neural network after training is complete.
The results of \cite{iso2018scale,funai2018thermodynamics} suggest that the RBM flow is closely related to the RG flow.
We carry out a critical examination of this conclusion.
The central tools we employ are correlation functions defined using the patterns generated by the RBM.
We give a detailed and precise argument showing that the largest scale features of RG and RBM 
patterns are in complete agreement. 
The correlation functions involved are non-trivial probes of the statistics of the generated pattern\footnote{The
 studies carried out in \cite{iso2018scale,funai2018thermodynamics} used averages of the RBM pattern. Our correlation functions provide more sensitive
probes into the structure of the pattern.} so the conclusion we reach is compelling.
We also find that if one probes smaller scale features there are important differences between the two patterns.
We will comment further on the interpretation of these results in the conclusions.

{
At this point, a comment is in order.
The word ``deep'' in ``deep learning'' indicates that many layers are stacked to produce the network.
Each layer in our network is an unsupervised RBM. 
The word ``flow'' in ``RG flow'' indicates that many small steps of coarse graining are carried out.
Each step performs a local averaging and one basic step is being repeated.
In this study we are developing methods that allow a comparison between one step of the 
renormalization group flow to one layer of the deep network.
Thus, although we spend much of our effort on comparing a single RBM layer to a single step in the RG flow,
we are interested in understanding learning achieved through the composition of many layers as an RG flow.
It is in this sense that although we study a single layer we nevertheless claim we are exploring the problem of deep
learning.}

The setting for our study is the two dimensional Ising model \cite{mccoy2014two}. 
This is a simple model of a magnet, built for many individual ``spins'' each of which should be thought of as a microscopic
bar magnet.
Each spin can be aligned ``up'' or ``down''.
Spins align at low temperatures producing a magnet. 
At high temperatures, spins are aligned randomly and there is no net magnetic field.
The spins themselves define a binary pattern (the two states are up or down) and it is these patterns that
the RBM learns.
An important motivation for this choice of model is that it is well understood. 
The theory exhibits a first order phase transition terminating at a critical point.
The theory at the critical point enjoys a conformal symmetry so that it can be solved exactly. 
It exhibits many interesting observables which we use to explore how deep learning is working \cite{carrasquilla2017machine,morningstar2017deep}.
For example, if a neural network generates a microstate of the model, we can ask what the corresponding temperature of the microstate is.
At the critical point special observables known as primary operators, can be defined.
Their correlation functions are power laws with powers that are known.
These are the natural variables which encode, completely, the long scale features of the patterns.
In this way, the Ising model gives a framework to explore deep learning both through the results of numerical experiments 
and using the complete understanding of the large scale features of the coarse grained system.
To probe whether deep learning is a type of coarse graining, we will see that this knowledge of correlations on
large length scales is a valuable tool.

Our study of an Ising magnet may seem rather far removed from more usual (and practical) applications, 
including for example image recognition and manipulation. 
However, one might be optimistic that lessons learned from the Ising model are applicable to 
these more familiar examples. 
Indeed, the energy function of the Ising model tries to align nearby spins with the result that nearby 
spins are correlated.
This is not at all unlike an image for which the color of nearby pixels is likely to be correlated \cite{saremi2013hierarchical}.

A description of deep learning in the RG framework would have important implications.
RG explains how macroscopic physics emerges from microscopic physics.
This understanding leads to an organization of the microscopic physics into features that are relevant or irrelevant, so that 
in the end the emergent patterns depend only on a small number of relevant parameters.
Carried over to the deep learning context, a similar understanding will strive to explain what features of the data and
which weights in the network are important for deep learning.
Such an understanding would have implications for what architectures are optimal and how the learning process can be
improved and made more efficient.

We now sketch the content of the paper and outline how it is organized. 
In Section \ref{background} we give a quick review of RBMs, RG and the Ising model, providing the background needed
to follow subsequent arguments.
In Section \ref{RBMflow} we consider the RBM flows defined using the matrix of weights learned by the network \cite{iso2018scale,funai2018thermodynamics}. 
By studying correlation functions of primary operators of the Ising conformal field theory, we argue that although
the RBM and RG patterns agree remarkably well on the largest scales of the pattern they differ on the short scale structure.
In Section \ref{MSflow} we examine the possibility that deep learning reconstructs an RG flow, with each layer 
of the deep network performing one step of the flow. 
Our discussion begins with a critical look at the argument given in \cite{mehta2014exact} that claims that 
deep learning is mapped onto the RG flow. 
The argument shows a system of equations that is obeyed by both the RBM and a variational realization of the RG flow. 
Our basic conclusion is that the argument of \cite{mehta2014exact} only shows that aspects of the RBM learning are consistent with 
the structure of the RG transformation. 
Indeed, we explicitly construct examples that satisfy the equations derived by \cite{mehta2014exact} that certainly do not perform 
an RG flow or arise from an RBM. 
Nevertheless, the arguments of \cite{mehta2014exact} are compelling and we find the possible connection between deep learning 
and RG fascinating and deserving of further study. 
Towards this end we couch some of the qualitative observations of \cite{mehta2014exact} as statements about the behavior of 
well chosen correlation functions. 
The form of these correlators, puts certain qualitative observations of 
Section IV.B. of \cite{mehta2014exact} onto a firm quantitative footing.
Finally we study the RG flow of the temperature.
This turns out to be interesting as it reveals a further difference between the RBM patterns and RG.
In the final Section of this paper, we discuss our results and suggest open directions that can be pursued.

\section{RBM, RG and Ising}\label{background}

In this section we introduce the background material used in our study.
The first subsection reviews RBMs emphasizing both the structure of the network and its implementation.
{Following this, the RG is reviewed, with an emphasis on aspects relevant to deep learning.}
This section concludes with a review of the Ising model, motivating why the model is considered.

\subsection{Restricted Boltzmann Machines}\label{RBM}

\begin{figure}
\includegraphics[width=0.5\textwidth]{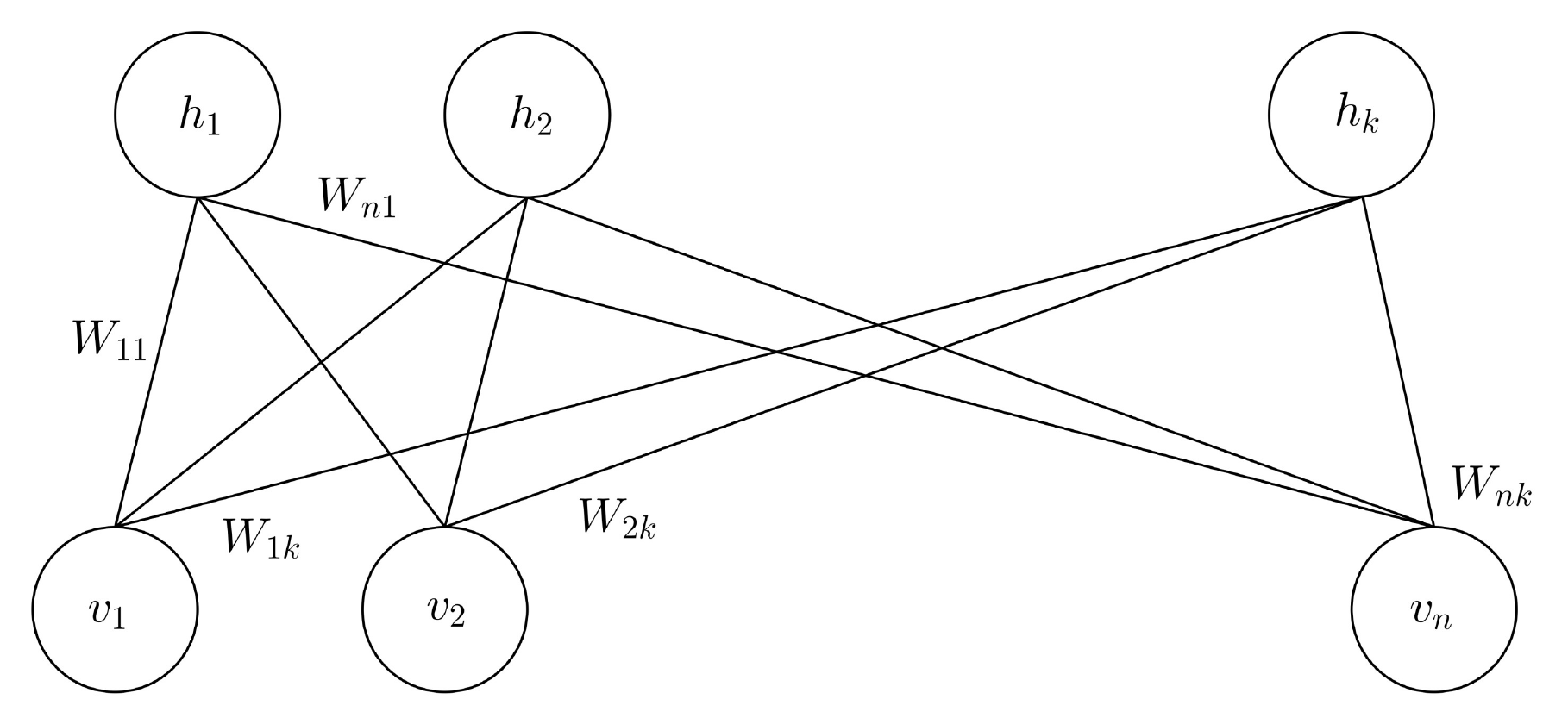}
\caption{An RBM network with visible nodes $v_i$ and hidden nodes $h_a$ where $N_v=n$ and $N_h=k$. Connections between visible and hidden nodes are each associated with a weight $W_{ia}$.}
\label{fig:rbm}
\end{figure}

RBMs perform unsupervised learning to extract features from a given data set \cite{smolensky1986information,freund1992unsupervised,hinton2002training}.
{They have a visible (input) and a hidden (output) layer.}
The {visible} layer is made up of visible nodes, $v_i$ with $i=1,2\cdots,N_v$ and the {hidden} 
layer is made up of hidden nodes, $h_a$ with $a=1,2,\cdots,N_h$ as illustrated in Figure \ref{fig:rbm}.

The {visible} nodes are set with values of $\pm 1$ and the trained network generates a corresponding
pattern by setting the output nodes to $\pm 1$.
The values of the {hidden} neurons are obtained by evaluating a non-linear function on a linear
combination of the {visible} neurons, perhaps offset by a constant bias.
{ The nonlinear function we use here is the hyperbolic tangent which can be seen in equation \eqref{eq:flowsh}.}
The specific linear combination of neurons is represented by connections between nodes, 
with a weight for each connection.  
For the RBM there are connections between every {visible} node and every {hidden} node, while 
nodes belonging to the same layer are not connected. 
The ``unrestricted'' Boltzmann machines allow connections between any two nodes in the network \cite{hopfield1982neural},
but this generality comes at a cost: training algorithms are much less efficient \cite{smolensky1986information,freund1992unsupervised,hinton2002training}.
The connection between {visible} node $v_i$ and {hidden} node $h_a$ is assigned a weight, $W_{ia}$, and visible (hidden)
nodes are assigned a bias $b_i^{(v)}$ ($b_a^{(h)}$).
Using these ingredients we define a Hamiltonian for the RBM
\bea
E=-\sum_{a}b_a^{(h)}h_a-\sum_{i,a}v_iW_{ia}h_a-\sum_{i}b_i^{(v)}v_i,
\eea
where $h_a,v_i \in \{-1,1\}$.
The RBM defines the probability distribution for obtaining configurations $\bf v$ and $\bf h$ of visible 
and hidden vectors by \cite{hinton2012practical}
\be
p({\bf v},{\bf h})=\frac{1}{\mathcal{Z}}e^{-E},
\label{eq:model}
\ee
where $\mathcal{Z}$ is the partition function, obtained by summing over all possible hidden and visible vectors
\be
\mathcal{Z}=\sum_{\{{\bf v},{\bf h}\}}e^{-E}.
\ee
As usual, to determine the marginal distribution of a visible vector, sum over the state space of hidden vectors
\be
p({\bf v})=\frac{1}{\mathcal{Z}}\sum_{\{{\bf h}\}}e^{-E}.
\ee
Similarly, the marginal distribution of a hidden vector is
\be
p({\bf h})=\frac{1}{\mathcal{Z}}\sum_{\{{\bf v}\}}e^{-E}.
\ee
The weights, $W_{ia}$ and biases $b_i^{(v)}$, $b_a^{(h)}$ are determined during training.
Training strives to match the model distribution $p({\bf v})$ to the distribution $q({\bf{v}})$ defined by the data and
it achieves this by minimizing the Kullback-Leibler (KL) divergence, which is given by

\bea
D_{KL}(q||p)&=&\sum_{{\{ v\}}} q({v})\left(\log\left(q({v})\right)-\log\left(p({v})\right)\right)\cr
&=&\sum_{\{ v\}} q({v})\log\left(\frac{q({v})}{p({v})}\right).
\eea

The KL divergence is a measure of how much information is lost when approximating the actual distribution with 
the model distribution \cite{carreira2005contrastive}.
Training adjusts $\{W_{ia},b_i^{(v)},b_a^{(h)}\}$ to minimize the KL divergence.
Gradients of the KL divergence used to update the parameters of the RBM are computed as follows
\bea
\frac{\partial D_{KL}(q||p)}{\partial W_{ia}}=&\langle v_ih_a \rangle_{data}-\langle v_ih_a \rangle_{model},
\label{eq:dklw}
\eea
\bea
\frac{\partial D_{KL}(q||p)}{\partial b_{i}^{(v)}}=\langle v_i \rangle_{data}-\langle v_i\rangle_{model},
\label{eq:dklbv}
\eea
\bea
\frac{\partial D_{KL}(q||p)}{\partial b_{a}^{(h)}}=\langle h_a \rangle_{data}-\langle h_a \rangle_{model},
\label{eq:dklbh}
\eea
where the expectation values appearing above are easily derived using (\ref{eq:model}).
They are given explicitly in Appendix \ref{EVs}.
The data set contains an enormous number $N_s$ of samples implying that the method just outlined is numerically
intractable: the sum over the whole state space of visible and hidden vectors is too expensive.
{In this study we consider networks with $N_v$ in the range of $100$ to $1024$ nodes and $N_h$ in the range of $81$ to $256$ nodes. The number of samples $N_s$ we sum over lies in the range of $2^{81}$ and $2^{1024}$.}
To make progress, {we} approximate the KL divergence by the contrastive divergence \cite{hinton2002training}.
Rather than summing over the entire state space of visible and hidden vectors, one simply sets the states of the 
visible units to the training data \cite{hinton2012practical}. 
This is an enormous simplification.
Given a set of visible vectors, the hidden vectors are sampled by setting each $h_a$ to 1 with probability
{
\be
p(h_a {= 1}|{\bf v})=\frac{1}{2}\left(1+\tanh\left(\sum_iW_{ia}{v_i}+b_a^{(h)}\right)\right).
\label{eq:hidd_recon}
\ee
Likewise, given a set of $h_a$, we are able to sample visible vectors by setting each $v_i$ to 1 with probability
\be
p(v_i {= 1}|{\bf h})=\frac{1}{2}\left(1+\tanh\left(\sum_aW_{ia}{h_a}+b_i^{v}\right)\right).
\label{eq:vis_recon}
\ee}
Expectation values for the data are computed using $\bf \hat{h}$, generated using (\ref{eq:hidd_recon})
and $\bf \hat{v}$, provided by the training data set.

To determine model expectation values, determine a sample of visible vectors {$\{\tilde{v}\}$} 
and a sample of the hidden vectors {$\{\tilde{h}\}$}, using the equations \eqref{eq:vis_recon} and \eqref{eq:hidd_recon}
as we now explain.
The set {$\bf\{\tilde{v}\}$}, is calculated using {$\{\bf\hat{h}\}$} and equation \eqref{eq:vis_recon}. 
We then determine {$\bf\{\tilde{h}\}$}, using {$\bf\{\tilde{v}\}$} and equation \eqref{eq:hidd_recon}. 
{The equations for the $Ath$ vectors in the sets $\bf\{\hat{h}\}$, $\bf\{\tilde{v}\}$ and $\bf\{\tilde{h}\}$ are thus}
\begin{equation}
\begin{split}
\hat{h}_a^{(A)}=\tanh\left(\sum_i W_{ia}\hat{v}_i^{(A)}+b_a^{(h)}\right),
\end{split}
\label{eq:flowsh}
\end{equation}
\begin{equation}
\begin{split}
\tilde{v}_i^{(A)}=&\tanh\left(\sum_a W_{ia}\hat{h}_a^{(A)}    +b_i^{(v)}\right),
\\
\end{split}
\label{eq:flowsv}
\end{equation}
\begin{equation}
\begin{split}
\tilde{h}_a^{(A)}=&\tanh\left(\sum_i W_{ia}\tilde{v}_i^{(A)}    +b_a^{(h)}\right),\\
\end{split}
\label{eq:flowsth}
\end{equation}
Expectation values of the model are approximated using these sets.
Again, summing these much smaller sets (and not the complete space of hidden and visible vectors) is an enormous simplification.

Using this approximation the expressions used to train the RBM are
\bea
\langle v_i h_a\rangle_{data}=\frac{1}{N_s}\sum_{A}\hat{v_i}^{(A)}\hat{h_a}^{(A)},
\label{eq:viha_data}
\eea
\bea
\langle v_i h_a\rangle_{model}=\frac{1}{N_s}\sum_{A}\tilde{v_i}^{(A)}\tilde{h_a}^{(A)},
\label{eq:viha_model}
\eea
\bea
\langle v_i \rangle_{data}=\frac{1}{N_s}\sum_{A}\hat{v_i}^{(A)},
\label{eq:vi_data}
\eea
\bea
\langle v_i \rangle_{model}=\frac{1}{N_s}\sum_{A}\tilde{v_i}^{(A)},
\label{eq:vi_model}
\eea
\bea
\langle h_a \rangle_{data}=\frac{1}{N_s}\sum_{A}\hat{h_a}^{(A)},
\label{eq:ha_data}
\eea
\bea
\langle h_a \rangle_{model}=\frac{1}{N_s}\sum_{A}\tilde{h_a}^{(A)},
\label{eq:ha_model}
\eea
where $A=1,2,3,\dots,N_s$ denotes samples in the training data set made up of $N_s$ samples.
These approximations achieve a dramatic speed up in training.
Although this method performs well in practice \cite{jordan2015machine,salakhutdinov2007restricted,ranzato2010factored},
it is difficult to understand when and why the approximations work \cite{hinton2012practical,tieleman2008training}. 
This approximation does not follow the gradient of any function \cite{sutskever2010convergence}.

\subsection{RG}

RG is a tool used routinely in quantum field theory and statistical mechanics \cite{kogut1974renormalization}. 
RG coarse grains by first organizing the theory according to length scales and then  
averaging over the short distance degrees of freedom.
The result is an effective theory for the long distance degrees of freedom. 
RG thus gives a systematic procedure to determine the dynamical laws governing macroscopic physics of a system 
with given microscopic laws, and it achieves this by employing coarse graining.
The analogy to deep learning should be evident: deep learning also extracts regularities from massive data sets.

{ At this point it is helpful to make a comment on what a field is.
A field is a type of observable.
A very simple example of a field could be the temperature inside a room.
The measured value of the temperature depends on exactly where in the room you make the measurement\footnote{For
example, there maybe an air conditioner in the room. Points closer to the air conditioner will be cooler.} and when you make the 
measurement\footnote{Temperature measurements at midnight in the middle of winter will typically be lower than measurements
at midday in the middle of summer.}.
Anything that can be measured everywhere and/or everywhen is an example of a field.}

To illustrate RG consider the example provided by quantum field theory.
{ To have a concrete example in mind, which is relevant to the discussion that follows, we might study a field
$\phi(x)$ describing the magnetization inside a magnet.
The value of the field $\phi(x)$ gives the value of the magnetization at position $x$.
$\phi(x)$ can be manipulated as we would normally manipulate a function of $x$.
In particular, we can take derivatives of $\phi (x)$ with respect to $x$ and we can take its Fourier transform to obtain $\phi(k)$.}
Observables ${\cal O}$ are functions (usually polynomials) of the field and its derivatives.
Examples of observables are the energy or momentum of the field.
To calculate the expected value $\langle {\cal O}\rangle$ of observable ${\cal O}$, integrate (i.e. average) over all 
possible field configurations
\bea
\langle {\cal O}\rangle =\int [d\phi] e^{-S} {\cal O}.
\eea
{To make sense of this integral one can work on a lattice.
Here we use the term ``lattice'' to denote an ordered array of points, and we imagine replacing the continuum of space
with this discrete structure, so that the set of all possible positions in space is now a discrete set. 
The integral over all possible field configurations then becomes a product of ordinary integrals, with one integral over the allowed range of the field, at each lattice site.
The range of the field is usually taken to be the real number field.}
The factor $e^{-S}$, which defines a probability measure on the space of fields, depends on the theory considered.
$S$ is called the action of the theory and is also a polynomial in the field and its derivatives, with the
coefficients of the polynomial providing the parameters of the theory, things like couplings and masses.
A theory is defined by specifying $S$.

To coarse grain, express the position space field in terms of momentum space components
\bea
\phi (x)=\int dk e^{ik\cdot x}\phi (k).
\eea
$e^{ik\cdot x}$ oscillates in position space with wavelength ${\frac{2\pi}{k}}$.
High momentum (big $k$) components have small wavelengths and encode small distance structure. 
Low momentum components have huge wavelengths and describe large distance structure.
Declare there is a smallest possible structure, implemented by cutting off the momentum modes
at a large momentum $\Lambda$ as follows
\bea
[d\phi ]=\prod_{k<\Lambda} d\phi (k).
\eea
RG breaks the integration measure into high and low momentum components
$[d\phi ]=[d\phi_<][d\phi_>]$ where
\bea
[d\phi_<]&=&\prod_{k< (1-\epsilon)\Lambda} d\phi (k)\cr
[d\phi_>]&=&\prod_{(1-\epsilon)\Lambda<k<\Lambda} d\phi (k)\label{ssplit}.
\eea
{ 
The dimensionless parameter $\epsilon$ defines the split between the two sets of components.
In the end we imagine taking $\epsilon\to 0$ as explained below.}
RG considers observables that depend only on large scale structure of the theory, i.e.
observables that depend only on $\phi_{<}$.
In this case, when computing the expected value of ${\cal O}$ we can pull ${\cal O}$ out of the integral over
$\phi_{>}$ and integrate over the high momentum components
\bea
\langle {\cal O}(\phi_{<})\rangle
&=&\int [d\phi_<] \int [d\phi_>]\,\,  e^{-S} {\cal O}(\phi_{<})\cr
&=&\int [d\phi_<]\,\, e^{-S_{\rm eff}} {\cal O}(\phi_{<}).\label{BasicRG}
\eea
This procedure of splitting momentum components into two sets and integrating over the large momenta
defines a new action $S_{\rm eff}$.
Repeating the procedure many times defines the RG flow under which $S_{\rm eff}$ changes continuously.
{ To obtain a continuous flow we should take the limit $\epsilon\to 0$, so that the procedure needs
to be repeated an infinite number of times to flow to low momentum. The parameter $\epsilon$ should be thought
of as a step size in a discrete flow. It is not a physical parameter and must be taken small enough that the results of
computations are independent of $\epsilon$.} 
After the flow, one is left with an integral over the very long wavelength modes.
This completes the coarse graining: we have a new theory defined by $S_{\rm eff}$.
The new theory uses only long wavelength components of the field and correctly reproduces the expected
value of any observable depending only on long wavelength components.
Values of the parameters of the theory, which appear in $S_{\rm eff}$, change under this transformation.
In general, many possible terms are generated and appear in $S_{\rm eff}$.
Each possible term defines a coupling of the theory.
Each coupling can be classified as marginal (the size of the coupling is unchanged by the RG flow), 
relevant (the coupling grows under the flow) or irrelevant (the coupling goes to zero under the flow).
It is a dramatic insight of Wilson that almost all couplings in any given quantum field theory are irrelevant and 
so the low energy theory is characterized by a handful of parameters.
This is a dramatic (experimentally verified) simplicity hidden in the rather complicated quantum field theory.
This simplicity explains why ``simple large scale patterns'' can emerge from ``complicated short distance data''.
The possibility that the same simplicity is at work in deep learning is a key motivation of this paper.

{ Although the equation (\ref{BasicRG}) defines the relationship between $S$ and $S_{\rm eff}$, the connection
is rather abstract and a few clarifying remarks are in order.
Consider the situation in which we started with an action $S$ and we have flowed to obtain some effective 
low energy dynamics $S_{\rm eff}$.
The action $S$ is the original action of the theory.
In the case of a magnet, this would describe the dynamics of atomic spins, where the relevant length scale is $10^{-10}$m.
The effective action would describe the dynamics of a classical magnet, where the relevant length scale may be $10^{-3}$m or even larger.
The renormalization group is the coarse graining that constructs the classical macroscopic physics from the microscopic physics.}

Conceptually, the coarse graining performed by RG is well defined.
Computationally, it is almost impossible to carry out.
To develop a useful calculation scheme, partition the momentum components into tiny sets (i.e. follow (\ref{ssplit}) with 
$\epsilon$ infinitesimal) and ask what happens when we average over a single tiny set.
Two things happen: couplings $g_i$ change {$\delta g_i = \beta_i\,\epsilon$} and the strength of the field changes 
{$\delta\phi = \gamma\,\epsilon\phi$}. 
One can prove that all observables built using $n$ fields will obey the Callan-Symanzik equation \cite{callan1970broken}
\bea
\left( \mu {\partial\over\partial\mu}+\sum_i \beta_i{\partial\over\partial g_i}+n \gamma\right)\langle{\cal O}\rangle
=0.
\eea
The parameter $\mu$ here defines the scale of the effective theory: the smallest wavelength in the effective theory is
${2\pi\over\mu}$. This equation provides a remarkable and simple description of the RG coarse graining that captures
the essential features of the long distance effective theory.
{In practice correlation functions are computed and then inserted into the Callan-Symanzik equation.
The $\beta_i$ and $\gamma$ functions are then read from the resulting equations.}

If RG (or a variant of it) is relevant to understanding deep learning, it makes concrete suggestions for the resulting theory.
For example, is there an analogue of the Callan-Symanzik equation?
One might assign beta functions $\beta_{ia},\beta_i^{(v)},\beta_a^{(h)}$ to the weights $W_{ia}$ 
and biases $b_i^{(v)},b_a^{(h)}$.
These would determine which parameters of the RBM are relevant, irrelevant or marginal.

The RG flow halts at a fixed point, described by a conformal field theory.
This field theory enjoys additional symmetries including scale invariance.
It is interesting to note that the possibility that scale invariance plays a role in deep learning has been raised in \cite{lin2017does,mehta2014exact,iso2018scale,funai2018thermodynamics,beny2013deep}.

{ Although we have focused mainly on a physical model in this article, we should point out that there are many
other applications of the renormalization group formalism.
For example, RG has been used to understand the spread of forest fires 
\cite{loreto1995renormalization}, in the modeling of the spread of 
infectious diseases in epidemiology \cite{clar1996forest}, for the prediction 
of earthquakes \cite{clar1996forest} and more generally, to any system with self organized 
criticality \cite{diaz1994dynamic}.}

\subsection{Ising model}\label{sec:ising}

The Ising model is a model for a magnet.
The two dimensional model has a discrete variable, { called a spin}, $\sigma_{\vec k}=\pm 1$ on each site 
of a rectangular lattice.
The sites are labeled by a two dimensional vector $\vec k$, which has integer components. 
{ A state of the system is given by specifying a collection of spins, 
$\{\sigma_{\vec k}\}$,
one for each site in the lattice.
To refer to a specific state of the spin system we use the notation $\sigma=\{\sigma_{\vec k}\}$.}
Spins on adjacent sites $i$ and $j$ interact with strength $J_{ij}$. 
Each spin will also interact with an external magnetic field $h_j$, with strength $\mu$.
The energy of a given configuration $\{\sigma_{\vec k}\}$ of spins is determined by the Hamiltonian
\bea
H=-\sum _{\langle i\,j\rangle }J_{ij}\sigma _{i}\sigma _{j}-\mu \sum _{j}h_{j}\sigma _{j},
\eea
where the first sum is over adjacent pairs, indicated by $\langle i\,j\rangle$. 
{ We simplify the model by setting the external field to zero $h_j=0$, 
and by choosing the couplings
in the most symmetric possible way $J_{ij}=J$. Since $J$ is an energy we can set it to 1 by choosing units appropriately.}
The probability of configuration {$\sigma=\{\sigma_{\vec k}\}$} of spins is given by the Boltzmann distribution, with inverse 
temperature $\beta\ge 0$
\bea
P_{\beta }(\sigma )={e^{-\beta H(\{\sigma_{\vec k}\})} \over Z_{\beta }},
\label{IsingDist}
\eea
where the constant $Z_\beta$, the partition function, is given by
\bea
Z_{\beta }=\sum_{\{\sigma_{\vec k}\} }e^{-\beta H(\sigma )}.
\eea
Averages of physical observables are defined by
\bea
\langle f\rangle _{\beta }=\sum _{\sigma }f(\sigma )P_{\beta }(\sigma ).
\eea
We study unsupervised learning of the Ising model by an RBM.
The visible data that is used to train the network is generated using the probability measure (\ref{IsingDist}).
{ The lattices have a total of $L_v\times L_v$ sites, with each site indexed by a position vector $\vec k$. 
We rearrange this array of spins $\sigma$ into an $N_v=L_v \times L_v$ dimensional vector by concatenating the rows 
of the given array. These components of these vectors are the training data input to the visible nodes of the RBM.}

There are good reasons to focus on the Ising model.
The model has a fixed point in its RG flow.
The fixed point is described by a well known conformal field theory (CFT) \cite{poland2019conformal}.
This fixed point is an unstable fixed point meaning that generic flows move away from the fixed point.
We must tune things carefully if we are to terminate on the fixed point.
This tuning is necessary because there is a relevant operator present in the spectrum of the conformal field theory and 
it tends to push us away from the fixed point.
{ We need to tune the temperature. If the temperature is slightly above the critical temperature,
thermal fluctuations destroy the long range correlations that are forming, whilst if we are slightly below the critical temperature,
the tendency of spins to align dominates and we find a state with all spins aligned and no fluctuations at all.
It is only exactly at the critical temperature that the system exhibits the interesting long range correlations that are 
described by the CFT.}
The papers \cite{iso2018scale,funai2018thermodynamics} argue that the RBM flow always flows to the fixed point.
This challenges conventional wisdom and it suggests a different kind of coarse graining to that employed by RG, is at work.
A distinct proposal \cite{mehta2014exact} claims that the RG flow arises by stacking RBMs to produce a classic deep learning scenario.
Each layer of the deep network performs a step in the flow.

At the Ising model fixed point, detailed checks of both proposals are possible.
There are CFT observables, known as primary operators, whose correlators are power laws of distances on the lattice.
The powers entering these power laws are known, so that we have a rich and detailed data set that the RBM must
reproduce if it is indeed performing an RG coarse graining.
This is a compelling motivation for the model.
Another advantage of the model is simplicity: it is a model of spins which take the values $\pm 1$ so it defines a simple 
model with discrete variables, well suited to numerical study and naturally accommodated in the RBM framework.
Finally, the Ising model is not that far removed from real world applications: the Ising Hamiltonian favors configurations 
with aligned neighboring spins.
Thus, at low enough temperatures ``smooth'' slowly varying configurations of spins are favored.
This is similar to data defining images for example, where neighboring pixels are likely to have the same color. 
In slightly poetic language one could say that at low temperatures the Ising model favors pictures and not speckle.

We end this section with a summary of the most relevant features of the Ising model fixed 
point.
At the critical temperature
\bea
T_{c}={\frac {2J}{k\ln(1+{\sqrt {2}})}},
\eea
where $J$ is the interaction strength and $k$ is the Boltzmann constant, 
the Ising model undergoes a second order phase transition.
{The critical temperature is given by $T_c\approx 2.269$ when $J=1$.}
There are two competing phases: an ordered (low temperature) phase in which spins align producing a macroscopic 
magnetization, and a disordered (high temperature) phase in which spins fluctuate randomly and the magnetization 
averages to zero.
At the critical point the Ising model develops a full conformal invariance and one can use the full power of conformal
symmetry to tackle the problem.
The field which takes values $\pm 1$ in the Ising model is a primary field, of dimension $\Delta={1\over 8}$.
The two and three point correlation functions of primary fields are determined by conformal invariance to be
\begin{equation}
\langle\phi(\vec x_1)\phi(\vec x_2)\rangle={B_1\over |\vec x_1-\vec x_2|^{2\Delta}},
\label{eq:2point_corr}
\end{equation}
\begin{equation}
\begin{split}
\langle\phi(\vec x_1)\phi(\vec x_2)\phi(\vec x_3)\rangle=
{B_2\over |\vec x_1-\vec x_2|^{\Delta}|\vec x_1-\vec x_3|^{\Delta}|\vec x_2-\vec x_3|^{\Delta}},
\end{split}
\end{equation}
where $B_1$ and $B_2$ are constants.
{ Since we study the Ising model on a lattice, the positions $\vec x_1$, $\vec x_2$ and $\vec x_3$ appearing
in the above correlation functions refer to sites in a lattice.}
There is also a primary operator in the Ising model (which we describe below) with a dimension $\Delta=1$.
These correlation functions must be reproduced by the RBM if it is indeed flowing to the
critical point of the Ising model.

\section{Flows derived from learned weights}\label{RBMflow}

In this section we consider the RBM flows introduced in \cite{iso2018scale,funai2018thermodynamics}.
These flows use the weight matrix $W_{ia}$, and bias vectors $b_i^{(v)}$ and $b_a^{(h)}$, obtained by training, to
define a continuous flow from an initial spin configuration to a final spin configuration.
The flow appears to exhibit a fascinating behavior: given any initial snapshot, the RBM flows towards the critical
point of the Ising model.
This is in contrast to the RG which flows away from the fixed point.
In addition, the number of spins in the configuration is a constant along the RBM flow.
In contrast to this, the number of spins in the configuration decreases along the RG flow, as high energy modes are
averaged over to produce the coarse grained description.
Despite these differences, the flow of \cite{iso2018scale,funai2018thermodynamics} appears to produce configurations ever closer to the critical
temperature and these configurations yield impressively accurate predictions for the critical exponents of the Ising magnet.
Our goal in this section is to further test if the RBM flow produces configurations at the critical point of the Ising model.
We explore the spatial dependence of spin correlations in configurations produced by the flow.
Our results prove that on large scales the Ising critical point configurations are correctly reproduced. 
However, we are also able to prove that as one starts to probe smaller scales there are definite quantifiable departures 
from the Ising predictions.

\subsection{RBM Flow}

RBM flows \cite{iso2018scale,funai2018thermodynamics} are generated using equation \eqref{eq:flowsv} together with 
the trained weight matrix, $W_{ia}$, and bias vectors, $b_i^{(v)}$ and $b_a^{(h)}$.
{Our data set $\hat{v}^{(A)}$ is labeled by an index $A$.
For each value of $A$,  $\hat{v}^{(A)}$ is a collection of spin values, one for each lattice site.
The RBM flow is generated through a series of discrete steps, with each step producing a new data 
set of the same size as the original.
Denote the data set produced after $k$ steps of flow by $\tilde{v}^{(A,k)}$, and by convention 
we identify $\tilde{v}^{(A,0)}$
with the original training set.
Apply equation \eqref{eq:flowsv} to $\hat{v}^{(A,k)}$ and then apply \eqref{eq:flowsth} 
to carry out a single step of 
the RBM flow, with the result $\tilde{v}^{(A,k+1)}$.
The flow proceeds by repeatedly applying equations \eqref{eq:flowsv} and \eqref{eq:flowsth}. 
Concretely, for a flow of length $n$, we have}
{
\begin{equation}
\begin{split}
{\tilde{v_i}^{(A,1)}}=&\tanh\left(\sum_a W_{ia}\hat{h}_a^{(A)}+b_i^{(v)}\right),\\
{\tilde{v_i}^{(A,2)}}=&\tanh\left(\sum_a W_{ia}\tilde{h}_a^{(A,1)}+b_i^{(v)}\right),\\
\vdots \\
{\tilde{v_i}^{(A,n)}}=&\tanh\left(\sum_a W_{ia}\tilde{h}_a^{(A,n-1)}+b_i^{(v)}\right).
\end{split}
\end{equation}

where 

\begin{equation}
    \begin{split}
   \tilde{h_a}^{(A)}=&\tanh\left(\sum_i W_{ia}\tilde{v}_i^{(A)}+b_a^{(h)}\right),\\
    \tilde{h_a}^{(A,1)}=&\tanh\left(\sum_i W_{ia}\tilde{v}_i^{(A,1)}+b_a^{(h)}\right),\\
    {\tilde{h_a}^{(A,2)}}=&\tanh\left(\sum_i W_{ia}\tilde{v}_i^{(A,2)}+b_a^{(h)}\right),\\
    \vdots \\
    {\tilde{h_a}^{(A,n)}}=&\tanh\left(\sum_i W_{ia}\tilde{v}_i^{(A,n)}+b_a^{(h)}\right).
    \end{split}
\end{equation}

}
Note that the length of the vector $\tilde{v}^{(A,k)}$ is a constant of the flow and consequently there is not obviously any
coarse graining implemented.

\subsection{Numerical results}\label{sec:rbm_flows_numeric}

This section considers statistical properties of configurations produced by the RBM flow.
At the Ising critical point, the theory enjoys a conformal invariance.
Using this symmetry a special class of operators with a definite scaling dimension $\Delta$ can be identified.
The utility of these operators is that their spatial two point correlation functions drop off as a known power of
the distance between the two operators, as reviewed above in equation \eqref{eq:2point_corr}.
These two point functions can be evaluated using the RBM flow configurations and, if these configurations are
critical Ising states, they must reproduce the known correlation functions.
This is one of the checks performed and it detects discrepancies with the Ising model predictions.
There are two primary operators we consider.
This first is the basic spin variable minus its average value $s_{ij}=\sigma_{ij}-\bar\sigma$.
The prediction for the two point function is \eqref{eq:2point_corr} with $\Delta_s={1\over 8}$.
This correlator falls off rather slowly, so that this two point function probes the large scale features of the
RBM flow configurations.
The RBM flow nicely reproduces this correlator and in fact, this is enough to reproduce the critical exponent
for the Ising model consistent with the results of \cite{funai2018thermodynamics}.
One should note however that our computation and those of \cite{funai2018thermodynamics} could very well have disagreed, since they
probe different things.
The critical exponent evaluated in \cite{funai2018thermodynamics} uses the magnetization computed from different flows generated by the RBM, 
at temperatures around the critical temperature.
Magnetization measures the average of the spin in the lattice.
It is blind to the spatial location of each spin.
On the other hand, the two-point correlation function is entirely determined by the spatial location of spins in a single
flow configuration.
Thus, the two point correlation function uses data at a single temperature, but uses detailed spatial dependence of the
lattice state.
We also consider a second primary operator
\begin{equation}
\epsilon_{ij}=s_{ij}\cdot (s_{i+1,j}+s_{i-1,j}+s_{i,j+1}+s_{i,j-1})-\bar{\epsilon},
\end{equation}
which has $\Delta_\epsilon=1$.
{ We have again subtracted off the average value of the operator.}
This correlation function falls off much faster and is consequently a probe of shorter scale features of the RBM configurations.
The RBM flow fails to reproduce this correlation function, indicating that the RBM configurations differ from those
of the critical Ising model. 
They have the same long distance features, but differ on shorter length scales.

\begin{figure}[t!]
\centering
\includegraphics[width=0.5\textwidth]{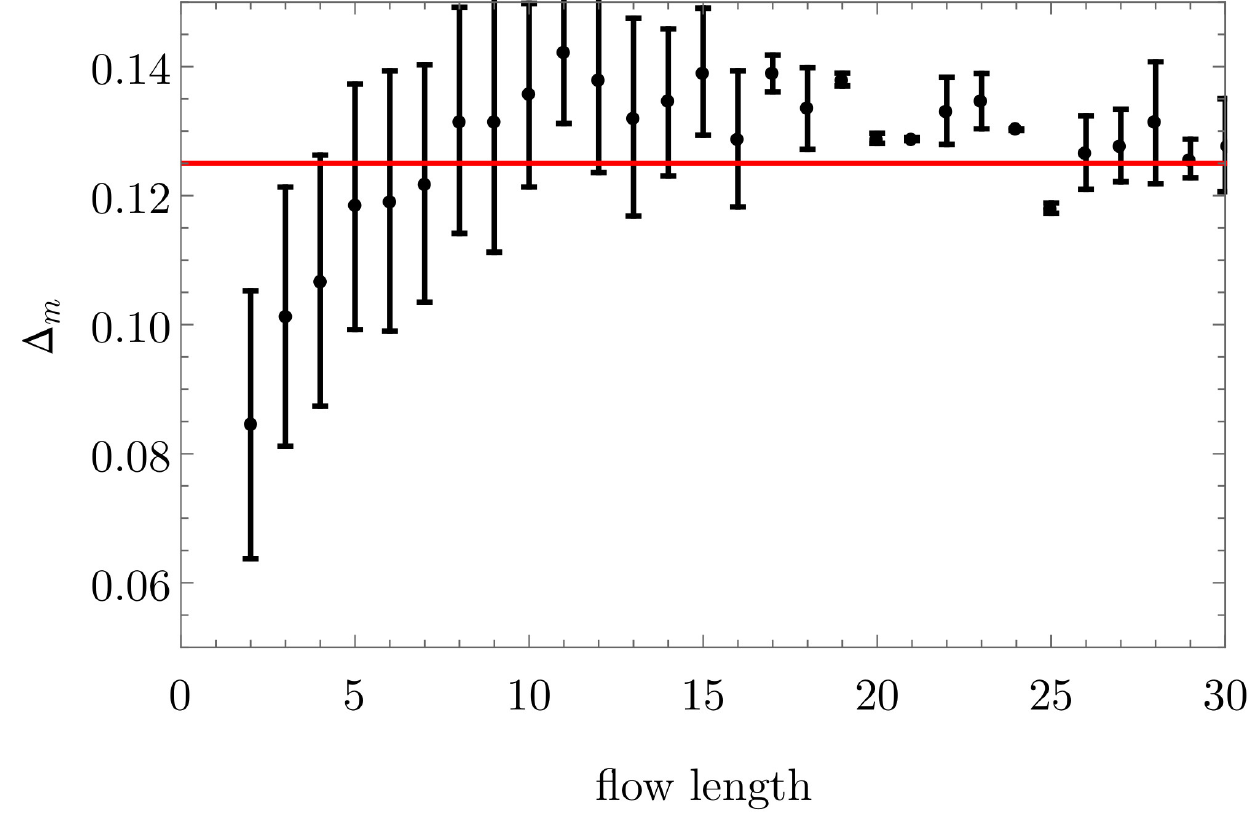}
\caption{Estimates of the scaling dimension $\Delta_m$ versus flow length, obtained using the average magnetization of flows 
at temperatures $T=2.1$, $2.2$ and $2.3$. The red line indicates the value of $\Delta=0.125$ at the critical point. 
After approximately 8 flows, $\Delta_m$ converges to this critical value. {The error bars are determined using Mathematica’s NonlinearModelFit function. Mathematica uses the Student's t-distribution to calculate a confidence interval for the given parameters with a 90\% confidence level.}}
\label{fig:scaldim_mag}
\end{figure}

\begin{figure}[h]
  \centering
\renewcommand{\thesubfigure}{a}
\subcaptionbox{\label{fig:T22}}{\includegraphics[width=0.44\textwidth]{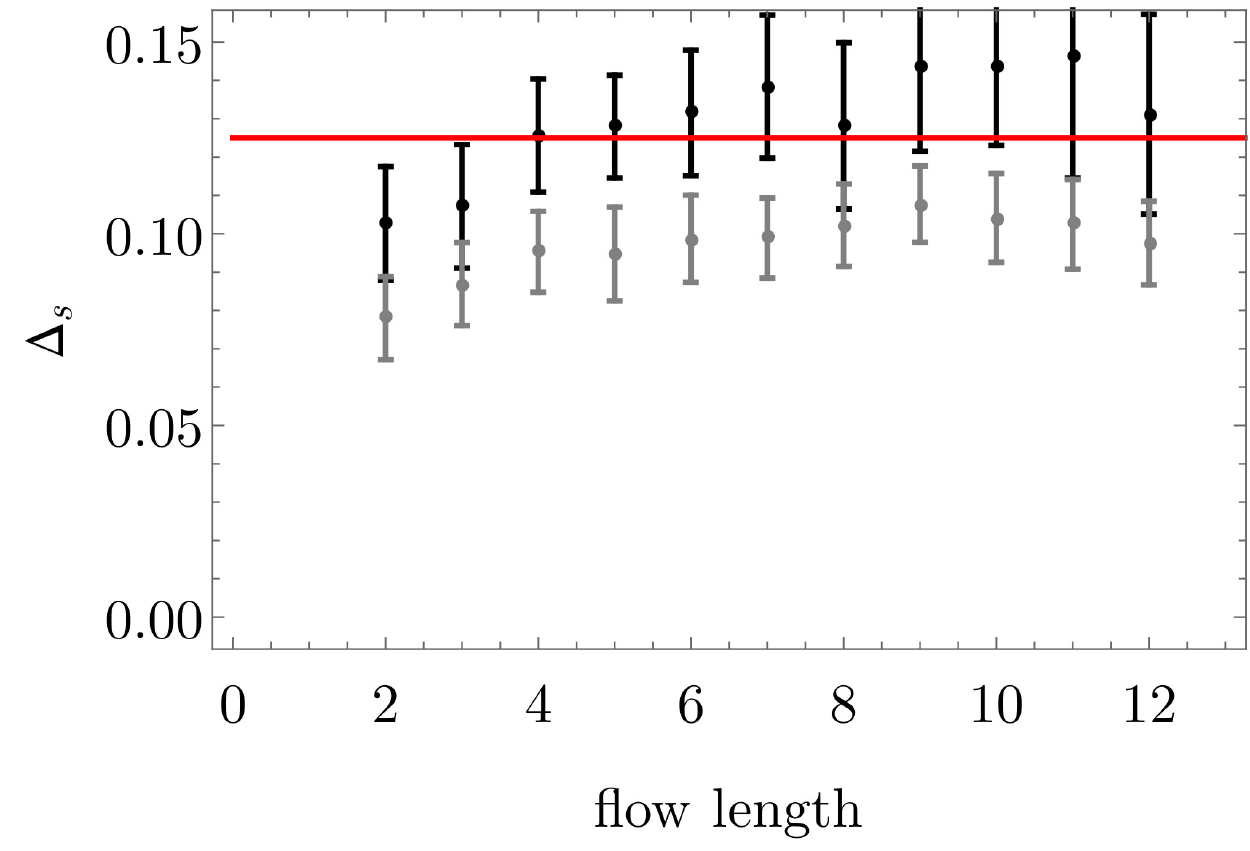}}
\renewcommand{\thesubfigure}{b}
\subcaptionbox{\label{fig:data}}{\includegraphics[width=0.44\textwidth]{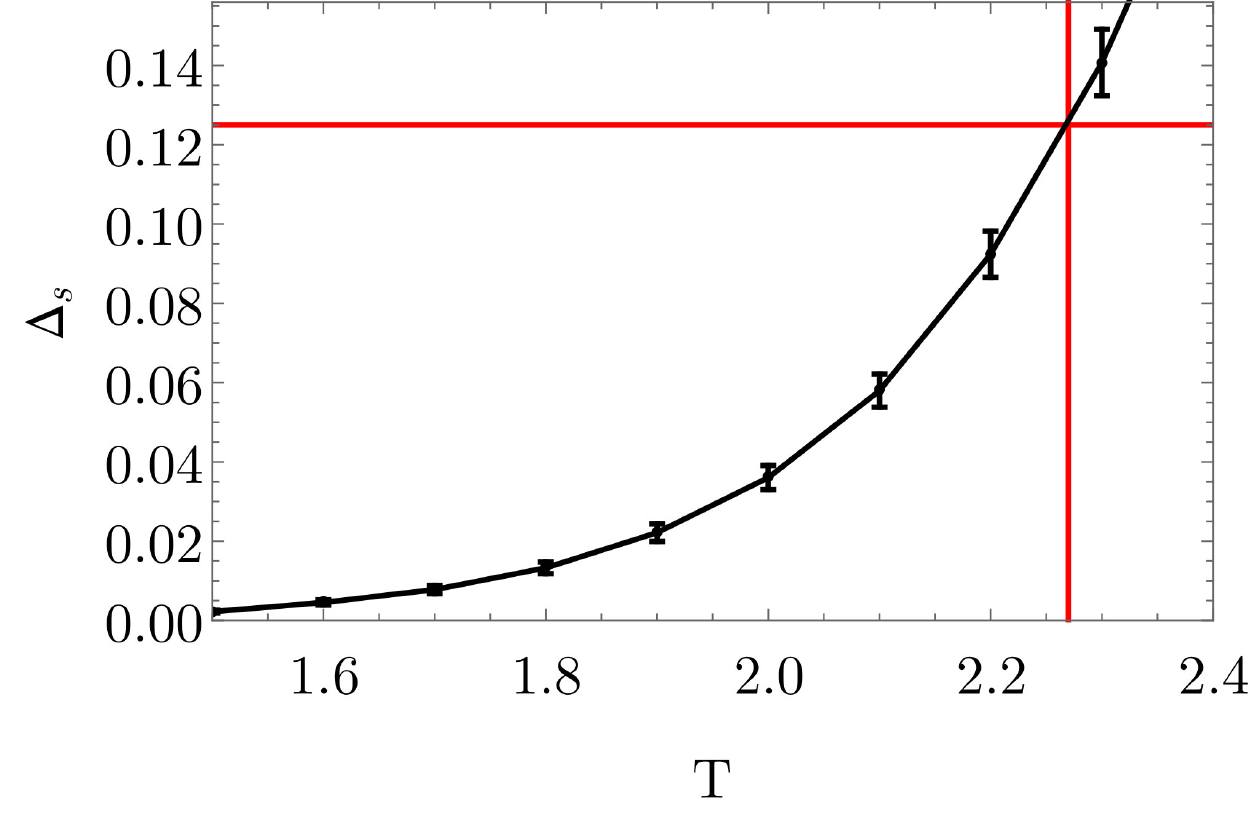}}
  \caption{Plot showing the estimated scaling dimension $\Delta_s$ versus flow length using the two-point correlation function 
for (a) flows at $T=2.2$ (in gray) and flows at $T=2.3$ (in black). Plot (b) shows the estimated scaling dimension 
versus temperature for the Ising model data used for training.
{The error bars are determined using Mathematica’s NonlinearModelFit function. Mathematica uses the Student's t-distribution to calculate a confidence interval for the given parameters with a 90\% confidence level.}}
  \label{fig:rbmflow_scaldim_2pt}
  \end{figure}
\begin{figure}[t!]
    \centering
\renewcommand{\thesubfigure}{a}
\subcaptionbox{\label{fig:eps_data100}}{\includegraphics[trim={0 0 0 0},clip,width=0.43\textwidth]{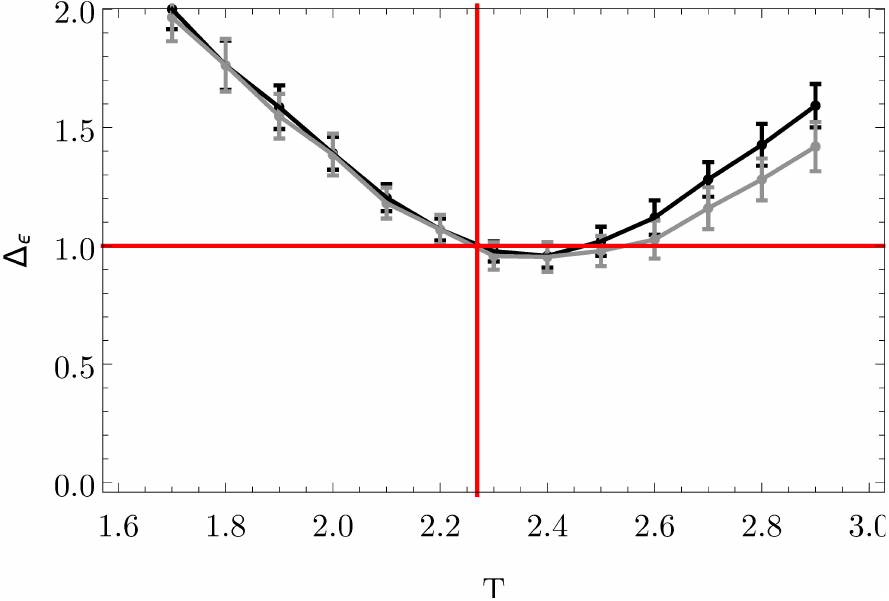}}
\renewcommand{\thesubfigure}{b}
\subcaptionbox{\label{fig:eps_22}}{\includegraphics[trim={0cm 0cm 0cm 0cm},clip,width=0.43\textwidth]{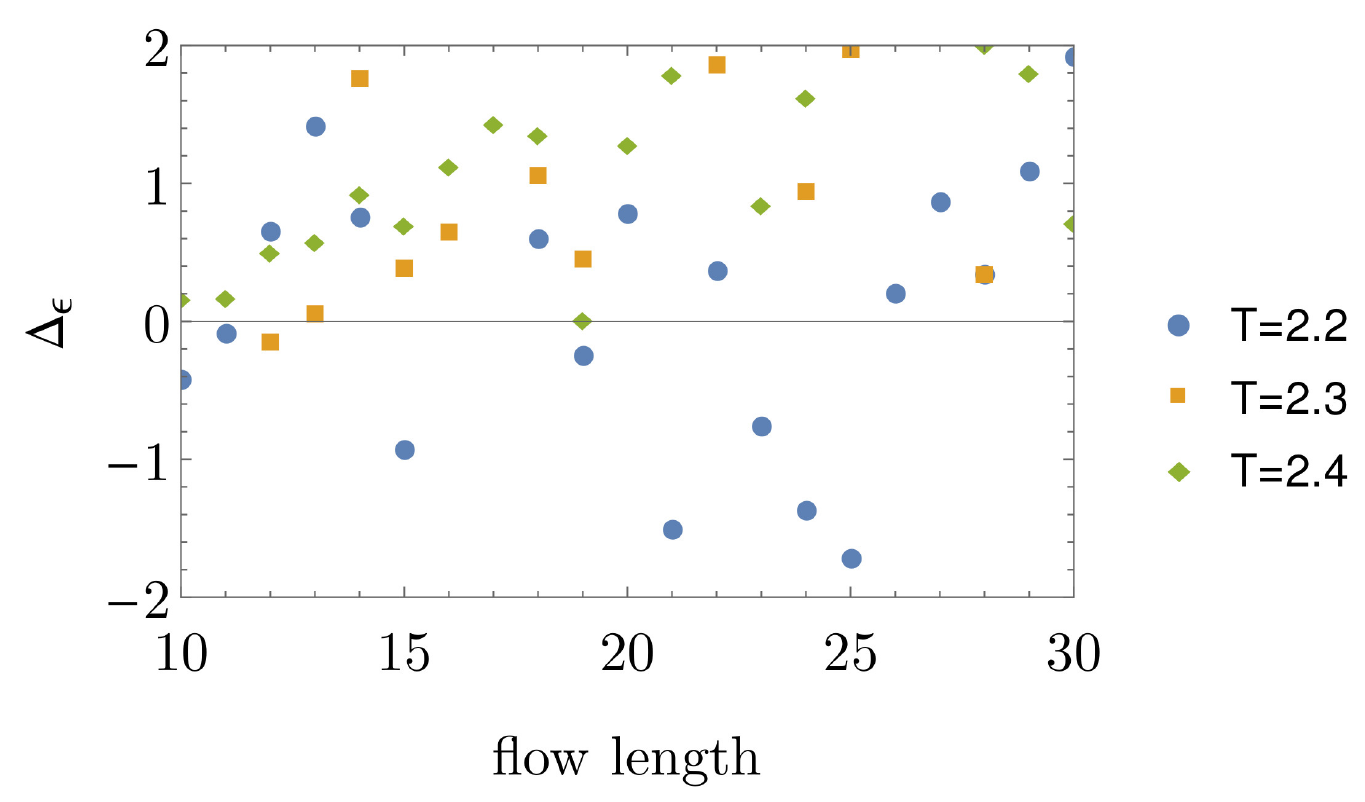}}
    \caption{Plots showing $\Delta_\epsilon$ calculated using (a) Monte Carlo Ising model data on a 10 by 10 lattice (in black) and a 9 by 9 lattice (grey), (b) RBM flows at a temperature of $T=$2.2, 2.3 and 2.4. Error bars in plot (a) indicate a 90\% confidence interval. No error bars are shown in (b) as the error bars are larger than the y range.
    {The error bars shown in (a) are determined using Mathematica’s NonlinearModelFit function. Mathematica uses the Student's t-distribution to calculate a confidence interval for the given parameters with a 90\% confidence level.}} 
    \label{fig:Delta_eps}
\end{figure}

Consider an RBM network with $100$ visible nodes and $81$ hidden nodes.
{This corresponds to an input lattice of size $10\times 10$ and an output lattice of size $9\times 9$.}
The number of visible and hidden nodes is chosen to match \cite{funai2018thermodynamics}, so that we can compare our results to existing
literature. 
{
We would like to study a lattice that is as small as possible but large enough to detect the power law fall off of the correlation functions we study. 
The power law behavior is given in the scaling dimensions $\Delta_s$ and $\Delta_\epsilon$. 
We demonstrate that when studying a lattice of size $9\times 9$ or larger we correctly determine 
the values for $\Delta_s$ and $\Delta_\epsilon$ using configurations generated from MC simulations. 
These results are shown in Figures \ref{fig:eps_data100} and \ref{fig:data}.
}

The network trains on data generated by Monte Carlo simulations which use the Boltzmann distribution 
given in equation \eqref{IsingDist} \cite{MCdatagen}.
The training data set includes $20000$ samples at each temperature, ranging from $0$ to $5.9$ in increments of $0.1$.
This gives a total of $1200000$ configurations. 
{Training uses $10000$ iterations of contrastive divergence, performed with the update 
equations \eqref{eq:flowsh} to \eqref{eq:ha_model} which are derived from equations 
\eqref{eq:dklw}, \eqref{eq:dklbv} and \eqref{eq:dklbh} \cite{dmk2020DLandRG}.} 

Once the flow configurations are generated, following \cite{iso2018scale,funai2018thermodynamics}, a supervised network is used to measure 
the temperature of each flow. 
{
The supervised network allows us to measure discrete temperatures of $T=0,0.1,\dots,5.9$.

We train a network which consists of three layers, an input layer with 100 nodes, a hidden layer with 80 nodes and an output layer with 60 nodes which correspond to the 60 temperatures we want to measure.
All nodes in the input layer are connected to all nodes in the hidden layer, and all nodes in the hidden layer are connected to all nodes in the output layer.
No connections between nodes within the same layer exist.

The $A$th sample of input data, $Z^{(1)}_{A}$ is transformed by the hidden layer using the hyperbolic tangent function $f(x)=\tanh(x)$ as follows
\begin{equation}
    Z^{(2)}_{Aa}=f(\sum_{i} Z^{(1)}_{Ai}W^{(1)}_{ia}+b_a^{(1)}).
\end{equation}
The output layer then transforms $Z^{(2)}_{A}$ into an output probability using the softmax function $g(x)$
\begin{equation}
    Z^{(3)}_{A\mu}=g(\sum_{a}Z^{(2)}_{Aa}W^{(2)}_{a\mu}+b^{(2)}_\mu)=:g(U^{(2)}_{A\mu}),
\end{equation}
where the softmax function normalizes the output to sum to 1 as follows
\begin{equation}
    g(U^{(2)}_{A\mu})=\frac{\exp(U^{(2)}_{A\mu})}{\sum_{\nu}\exp(U^{(2)}_{A\nu})}.
\end{equation}
The output of the trained network can thus be interpreted as probabilities for the temperatures we measure. 
The highest probability in the output is taken as the temperature for the given input.

We make use of the Keras library \cite{chollet2015keras} to train this network using the back-propagation algorithm \cite{rumelhart1986learning}.
The cost function used to train the network is the KL divergence as is used in \cite{funai2018thermodynamics}.
We choose a learning rate of $\epsilon=0.1$ and train the network for 3000 epochs.
In Figure \ref{fig:sup_loss} we show the validation and training loss versus the number of training epochs. 
The supervised network is trained using the same data set used to train the RBM as well as corresponding labels for each vector.
These labels are one hot encoded vectors which correspond to the temperatures $T=0,0.1,\dots,5.9$.
Here we use a split of 40 \% of the input data for validation and 60 \% of the input data for training. 
Figure \ref{fig:sup_loss} shows that the supervised network has converged to a 
loss very close to zero after 3000 epochs.

\begin{figure}[h!]
    \centering
    \includegraphics[width=0.4\textwidth]{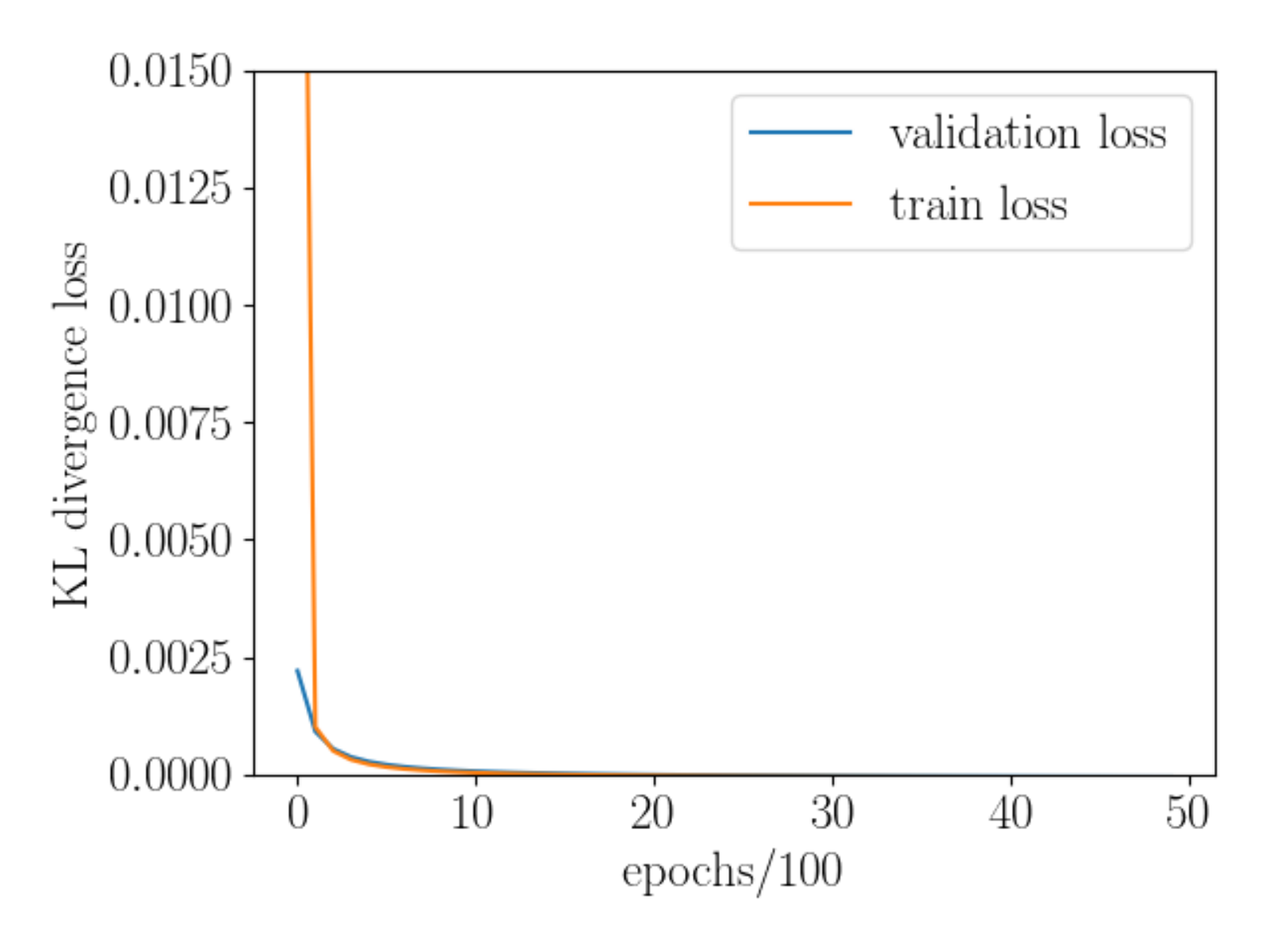}
    \caption{{Plot showing the KL divergence loss function of the supervised temperature measuring network versus the number of epochs divided by 100.
    This network consists of an input layer with 100 nodes, a hidden layer with 80 nodes and and output layer of 60 nodes.}}
    \label{fig:sup_loss}
\end{figure}

}

To estimate $\Delta$ using magnetization the study \cite{funai2018thermodynamics} selects flows at temperatures close to $T_c$, 
where the average magnetization $m$ depends on temperature as 
\begin{equation}
m\approx \frac{A|T-T_c|^{\Delta_m}}{T_c}.
\label{eq:m_approx}
\end{equation}

{
In \cite{funai2018thermodynamics} the magnetization is expanded about the critical point to give values for $A=1.22$ and $\Delta_m=1/8=0.125$
\begin{equation}
  m\approx 1.222 \frac{|T_c-T|^{1/8}}{T_c}.  
\end{equation}
}

We denote $\Delta$ obtained by this fitting as $\Delta_m$.
The fit also determines $A$. 
{The value of $T_c=2.269$ is a known theoretical value for the 2D Ising model with coupling strength $J=1$.} 
The fit uses the magnetization computed at temperatures $T=2.1$, $2.2$ and $2.3$.
A plot of $\Delta_m$ versus flow length is given in Figure \ref{fig:scaldim_mag}.
{ The error bars shown in Figure \ref{fig:scaldim_mag} (and all subsequent plots) are determined using Mathematica’s NonlinearModelFit function. The error bars show the standard error obtained from the regression. Mathematica uses the Student's t-distribution to calculate a confidence interval for the given parameters with a 90\% confidence level.}

\begin{figure}[h!]
    \centering
    \includegraphics[width=0.4\textwidth]{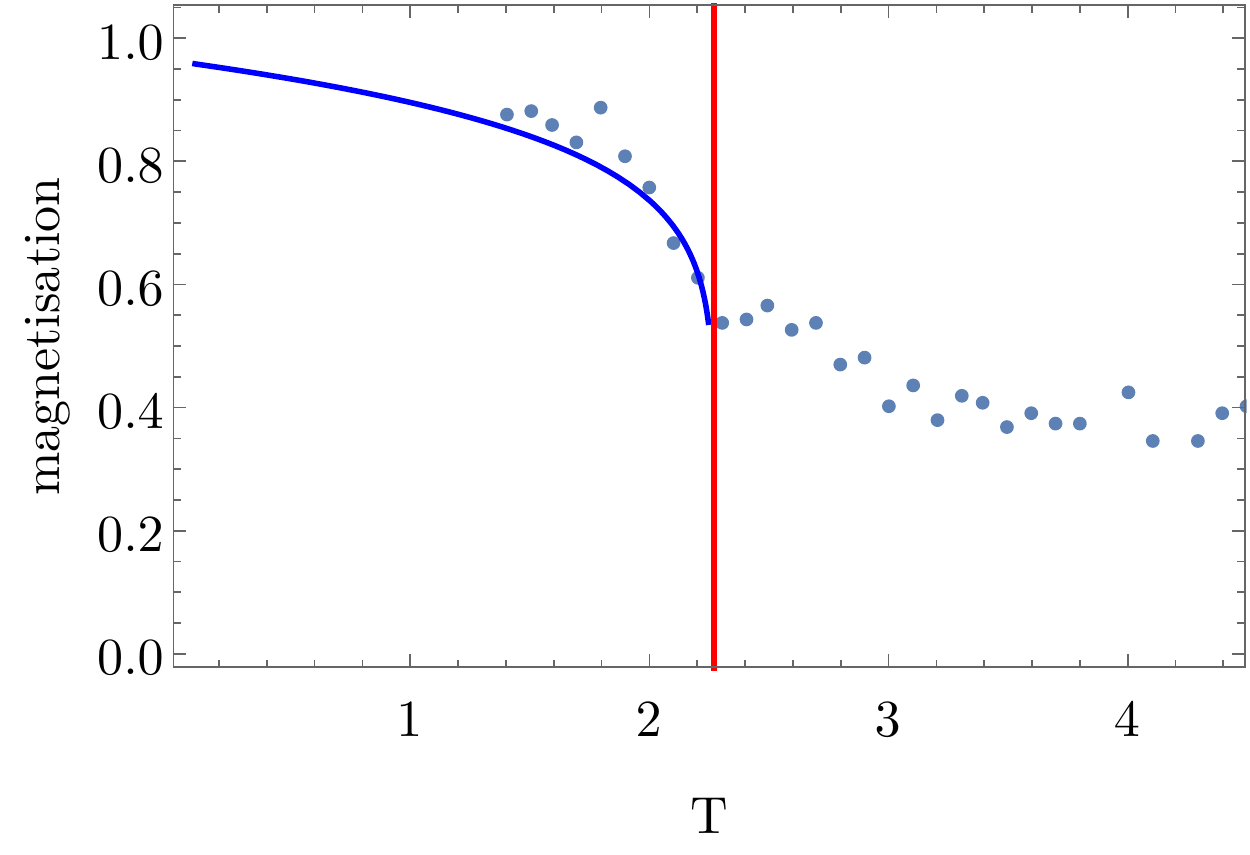}
    \caption{{Plot showing the fit for equation \eqref{eq:m_approx}. The blue line shows the function $m=0.942(T-2.269)^{0.126}$. This function is fitted using the dots which show the average magnetization for flows of length 26 at various temperatures measured using the supervised network. The vertical red line shows the critical temperature of $T_c\approx 2.269$. Equation \eqref{eq:m_approx} is fitted using data points for temperatures near $T_c$.}}
    \label{fig:mag_fit}
\end{figure}

{Our results indicate that we converge to the correct critical value $\Delta_m=0.125$ for flows of length 26. 
It is evident from Figure \ref{fig:scaldim_mag} that the flow converges to the theoretical value depicted by the red horizontal 
line.
The convergence of the flow is also reflected as a decrease in the size of the error bars as the flow proceeds.
In \cite{funai2018thermodynamics} the value for $A/T_c$ is found to be $0.931$. 
The value of $A/T_c$ which we find after convergence is $0.942$ with a 90\% confidence interval within $\pm 0.0168794$ 
of this value.
Although the values we obtain for $A$ and $\Delta_m$ are consistent with the results of
\cite{funai2018thermodynamics}, they have used a flow of length 9, at which point $\Delta_m$ is correctly determined.
As just described, we need longer flows for convergence.
The fitting of $\Delta_m$ and $A$, for a flow length of 26, is shown in Figure \ref{fig:mag_fit}.}

We now shift our focus to consider spatial two point correlation functions computed using the configurations generated
by the RBM flows.
The correlators are calculated using the flow configuration and the result is then fitted to the function in equation 
\eqref{eq:2point_corr} to estimate $\Delta$.
We denote this estimate by $\Delta_s$ as it is determined using spatial information.
{For RBM flows at the critical temperature, the prediction which is determined by theory} is $\Delta_s=0.125$ at $T=2.269$, as explained above.
{
We expect that for temperatures below the critical temperature $\Delta_s$ will be less than the theoretical value of $0.125$.
For temperatures that are above the critical temperature we expect that $\Delta_s$ will be greater than $0.125$.}
With a lattice of 10 by 10 spins, we find $\Delta_s=0.1263$ using Monte Carlo Ising model configurations. 
A plot showing this estimate can be seen in Figure \ref{fig:data}.
The point of this exercise is to demonstrate that a lattice of size 10 by 10 is large enough 
to estimate the scaling dimension of interest { and to verify the integrity of the data set used in the numerical simulations.}

Figure \ref{fig:T22} shows the scaling dimension $\Delta_s$ versus the flow length, for RBM flows at temperatures of $2.2$ (in gray) and $2.3$ (in black).
The red horizontal line indicates the scaling dimension at the critical point.
The results are intuitively appealing.
The gray points in Figure \ref{fig:T22} show estimates of $\Delta_s$ from flows slightly below the critical temperature, where the scaling
dimension is slightly underestimated. 
Below the critical temperature spins are more likely to align and so the correlator should fall off more slowly than
at the critical temperature.
This is what our results show.
The black points in Figure \ref{fig:T22} show $\Delta_s$ estimated using flows slightly above the critical temperature.
The scaling dimension is over estimated, again as expected.
Selecting flows at $T_c$ would determine the scaling dimension in between the values shown in Figure \ref{fig:T22}. 
{This gives a value very close to the theoretical value} of $\Delta_s=0.125$.

The two point correlation functions for the spin variable establish that the critical Ising states and the states produced by
the RBM flow share the same large scale spatial features.
We will now consider the two point correlation function of the $\epsilon_{ij}$ field, which probes spatial features on a smaller
scale.
Using critical Ising data generated using Monte Carlo, on a lattice of size 10 by 10 and 9 by 9, we estimate $\Delta_\epsilon$ 
at various temperatures as shown in Figure \ref{fig:eps_data100}.
{We can estimate the error for the point we are interested in, at $\Delta_\epsilon=1$ and $T=2.269$ which is depicted by the red vertical and horizontal lines shown in Figure \ref{fig:eps_data100}. 
The error bar of the estimate for the grey line ($9 \times 9$ lattice) is $\pm 0.059$ for an estimated value of $\Delta_\epsilon$ being $1.013$. 
For the black line ($10 \times 10$ lattice) we have an error bar of $\pm 0.044$ on an estimated value of $1.023$ for $\Delta_\epsilon$. 
This is determined using the average of the values as well as the errors on either side 
of the red vertical line at $T = 2.269$.}
{
Figure \ref{fig:eps_data100}, demonstrates that a lattice of size $9\times 9$ is large enough to correctly determine the $\epsilon$ scaling dimension $\Delta_\epsilon$.
}

The intersection of the red horizontal and vertical lines cross the critical temperature and prediction $\Delta_\epsilon=1$.
Interpolating the Ising data with a continuous curve, we would pass through the intersection point, as predicted.
These numerical results again demonstrate that a lattice of size 10 by 10 is large enough for the questions we consider.
The RBM flows are unable to confirm this prediction.
Indeed, the RBM flows near $T_c$ are summarized in Figure \ref{fig:eps_22}.
None of the three temperatures shown have a value of $\Delta_\epsilon$ that converges with flow length.

The fact that the RBM produces configurations that correctly reproduce the correlation function of the spin field $s_{ij}$ 
but not of the $\epsilon_{ij}$ implies that although the spatial correlations encoded into the RBM flow configurations 
agree with those of the critical Ising configurations at long length scales, the two start to differ on smaller length scales.
{This conclusion agrees with \cite{funai2018thermodynamics} which also finds differences between the RBM flow and 
RG. \cite{funai2018thermodynamics} considers $h\neq 0$ and uses different arguments to reach the conclusion.}

\section{Flows derived from deep learning}\label{MSflow}

The RBM flows of the previous section provide one possible link to RG.
An independent line reasoning, developed in \cite{mehta2014exact}, claims a mapping between deep learning and RG.
The idea is not that there is an analogy between deep learning and RG, but rather, that the two are to be identified.
The argument for this identity starts from the energy function of the RBM, which is
\begin{equation}
E(\{v_i,h_a\})=b_ah_a+v_iW_{ia}h_a+c_iv_i.
\end{equation}
This energy determines the probability of obtaining configuration $\{v_i,h_a\}$ as
\begin{equation}
p_{\lambda}(\{v_i,h_a\})=\frac{e^{-E(\{v_i,h_a\})}}{\mathcal{Z}},
\end{equation}
{where $\lambda=\{b_a,W_{ia},c_i\}$ are the parameters of the RBM model which are tuned during training.}
Marginal distributions for hidden and visible spins are defined as follows
\bea
p_{\lambda}(\{h_a\})=\sum_{\{v_i\}}p_{\lambda}(\{v_i,h_a\})=\tr_{v_i}p_{\lambda}(\{v_i,h_a\}),\cr
p_{\lambda}(\{v_i\})=\sum_{\{h_a\}}p_{\lambda}(\{v_i,h_a\})=\tr_{h_a}p_{\lambda}(\{v_i,h_a\}).
\label{KeyRBM}
\eea
The equations (\ref{KeyRBM}) are key equations of the RBM and \cite{mehta2014exact} essentially uses these to characterize the RBM.
The comparison to RG is made by employing a version of RG known as variational RG.
This is an approximate method that can be used to perform the renormalization group transformation in practice.
As explained in Appendix \ref{varrg}, the variational RG uses an operator $T(\{v_i,h_a\})$ defined as follows

\begin{equation}
\frac{e^{H_{\lambda}^{RG}(\{h_a\})}}{\mathcal{Z}}
=\tr_{v_i}\frac{e^{T(\{v_i,h_a\})-H(\{v_i\})}}{\mathcal{Z}}.
\label{frst}
\end{equation}
In this formula, $H(\{v_i\})$ is the microscopic Hamiltonian describing the dynamics of the visible spins 
and $H_{\lambda}^{RG}(\{h_a\})$ is the coarse grained Hamiltonian describing the hidden spins {where here $\lambda$ defines the parameters of the variational RG}.
{Block spin averaging is discussed in more detail in Appendix \ref{BSA}.}
The operator $T(\{v_i,h_a\})$ is required to obey 
{(see in equation \eqref{eq:varrg_sumha}})
\begin{equation}
\tr_{h_a}e^{T(\{v_i,h_a\})}=1,
\end{equation}
which obviously implies that
\begin{equation}
\tr_{h_a}e^{T(\{v_i,h_a\})-H(\{v_i\})}=e^{-H(\{v_i\})}.
\label{scnd}
\end{equation}
Notice that (\ref{frst}) and (\ref{scnd}) exactly match (\ref{KeyRBM}) as long as we identify
\begin{equation}
T(\{v_i,h_a\})=-E(\{v_i,h_a\})+H(\{v_i\}).
\end{equation}
This then implies that
\bea
\frac{e^{H_{\lambda}^{RG}(\{h_a\})}}{\mathcal{Z}}
&=&\tr_{v_i}\frac{e^{T(\{v_i,h_a\})-H(\{v_i\})}}{\mathcal{Z}}\cr
&=&\tr_{v_i}\frac{e^{-E(\{v_i,h_a\})}}{\mathcal{Z}}\cr
&=&\frac{e^{-H_{\lambda}^{RBM}(\{h_a\})}}{\mathcal{Z}},
\eea
which is the central claim of \cite{mehta2014exact}.

\begin{figure}[t!]
    \centering
\renewcommand{\thesubfigure}{a}
\subcaptionbox{\label{fig:a-i-rg}}{\includegraphics[trim={0 0 0 0},clip,width=0.23\textwidth]{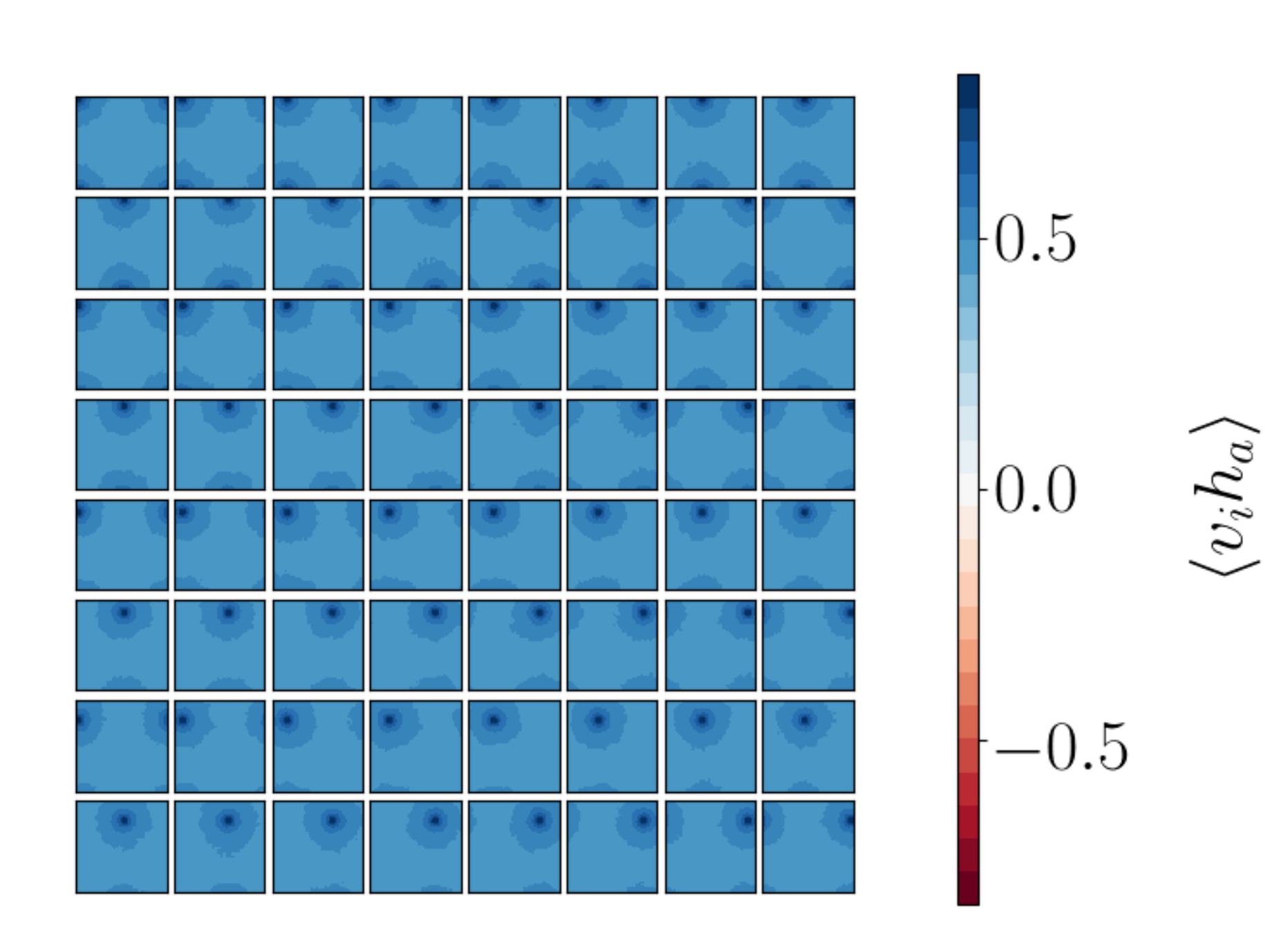}}~
\renewcommand{\thesubfigure}{b}
\subcaptionbox{\label{fig:b-i-rg}}{\includegraphics[trim={0 0 0 0},clip,width=0.23\textwidth]{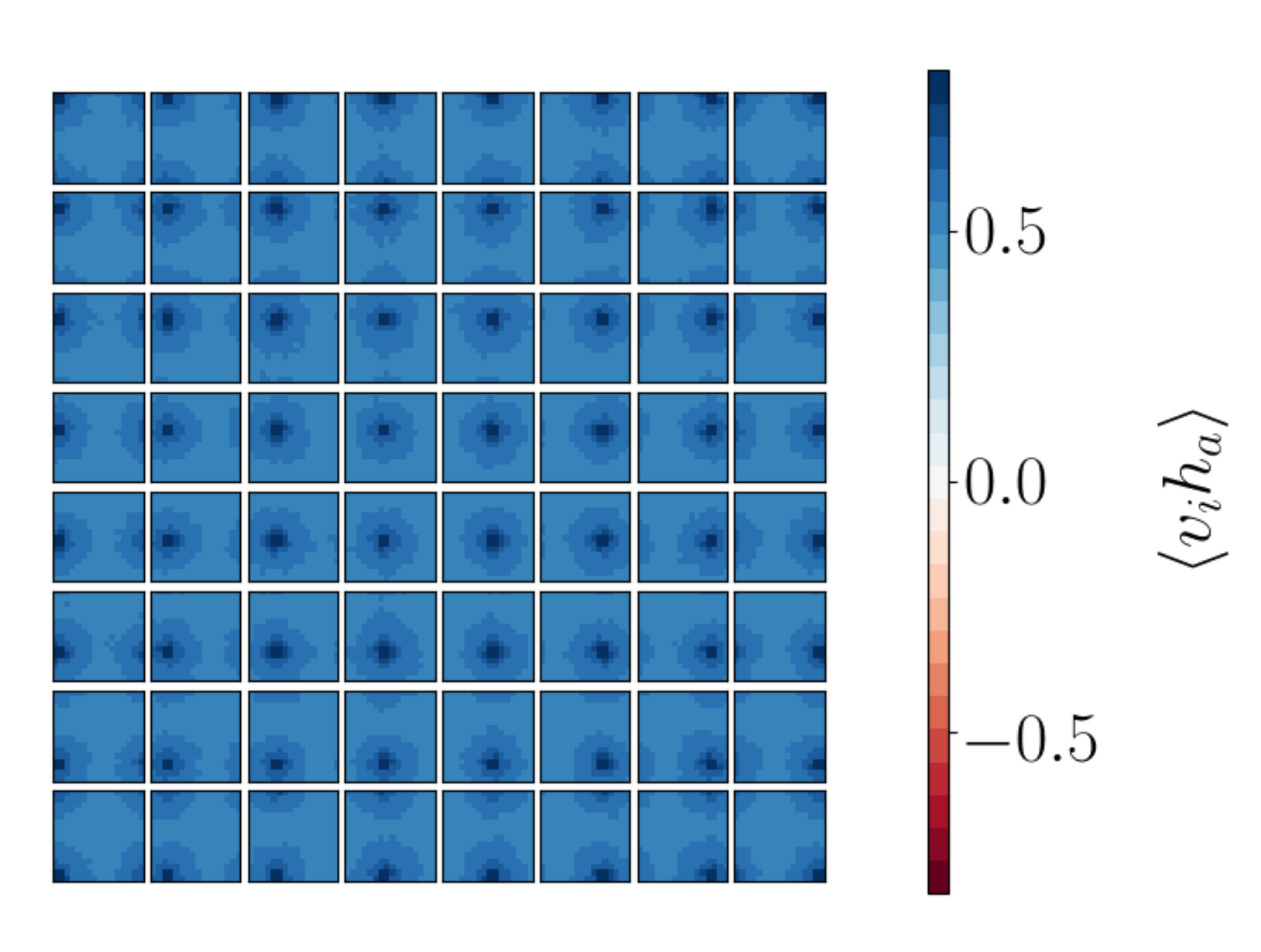}}
\renewcommand{\thesubfigure}{c}
\subcaptionbox{\label{fig:c-i-rg}}{\includegraphics[trim={0 0 0 0},clip,width=0.237\textwidth]{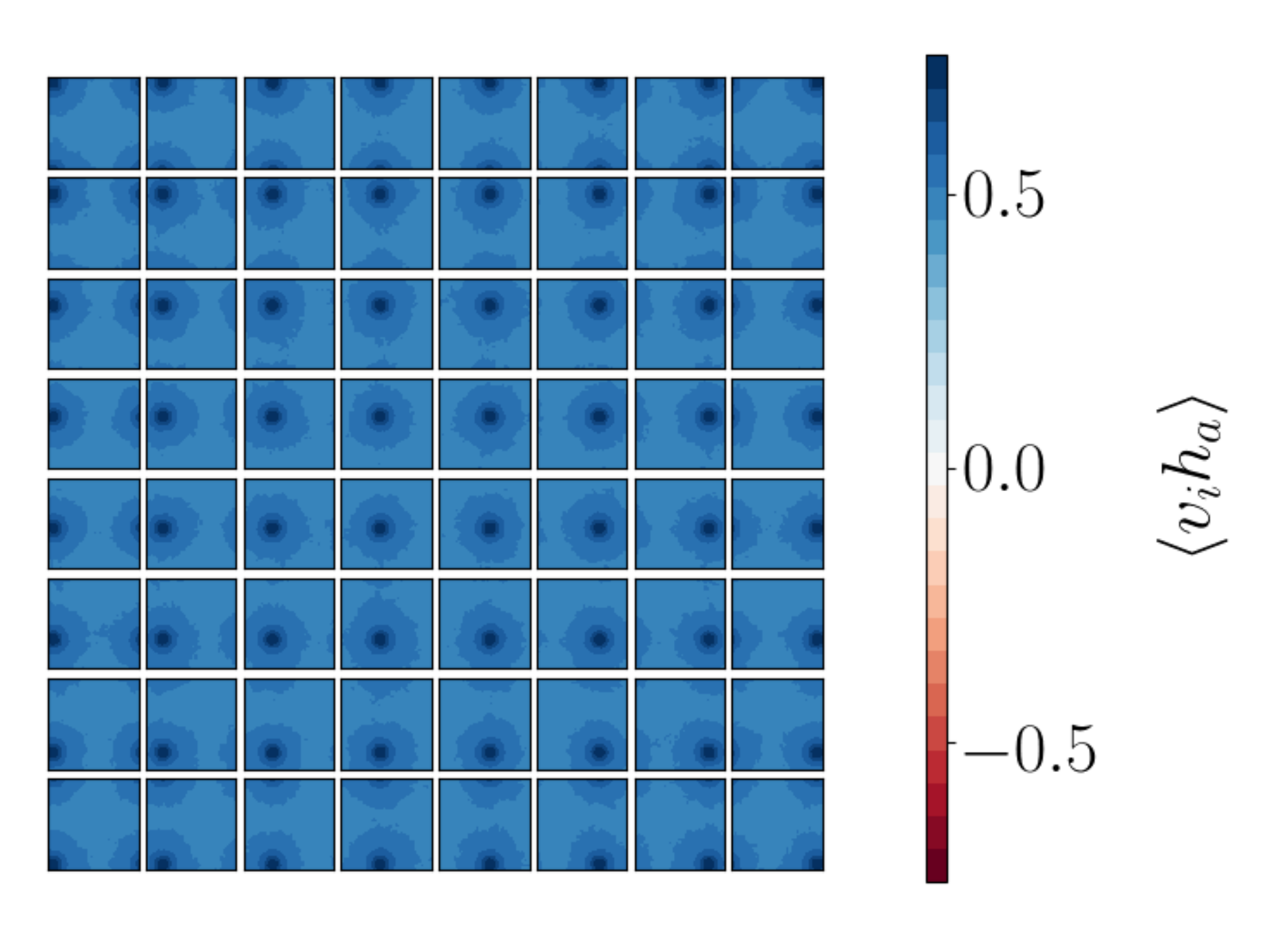}}~~
\renewcommand{\thesubfigure}{d}
\subcaptionbox{\label{fig:rg_deci}}{\includegraphics[trim={0 0 0 0},clip,width=0.2\textwidth]{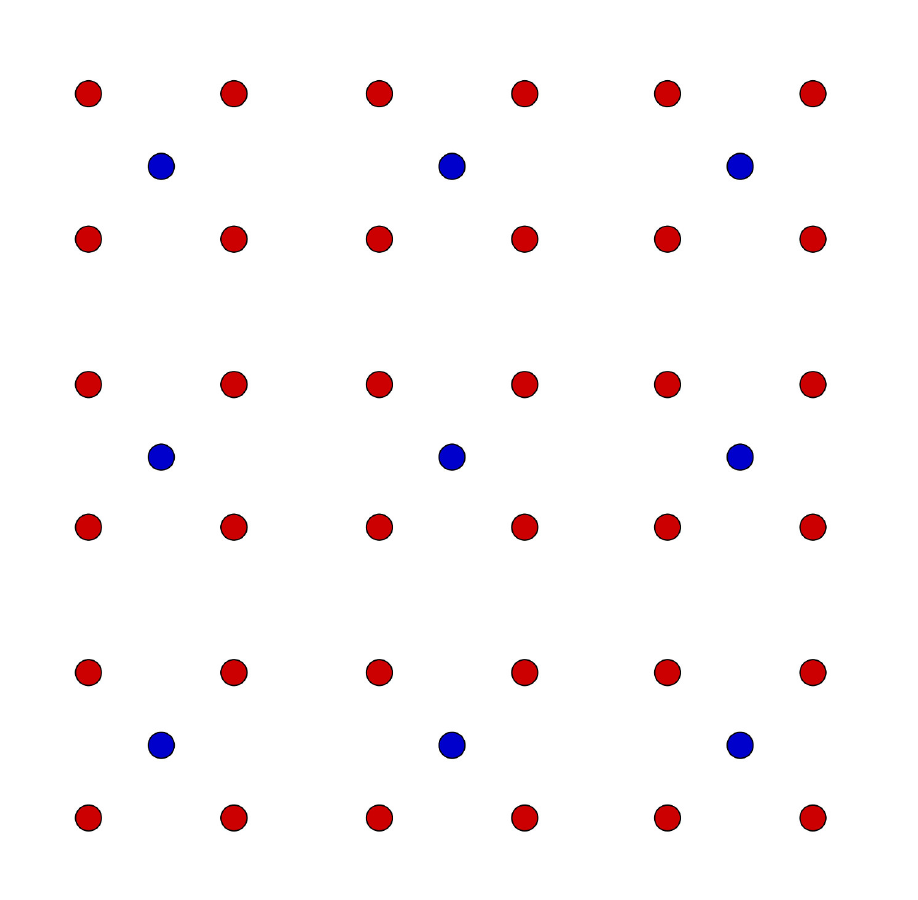}}
    \caption{Correlation plots for Ising model visible data with lattice size 32 by 32 at $T_c$ and RG decimated Ising data of sizes 16 by 16 (one step of RG) and 8 by 8 (two steps of RG). 
(a) shows visible Ising data correlated with configurations resulting after one step of RG.
(b) shows correlations between configurations resulting after one step of RG and configurations resulting after two steps of RG.
(c) shows correlations between Ising model visible data and configurations resulting after two steps of RG. 
(d) shows one step of RG. The red dots show the original visible lattice and the blue dots show the lattice obtained after one step of RG. Each blue dot is surrounded by four red dots. The value of the blue dot is determined by averaging the surrounding four red dots.}
    \label{fig:vihaplot_rg}
\end{figure}

The above argument proves an equivalence between deep learning and RG if and only if the equations (\ref{KeyRBM}) 
provide a unique characterization of the joint probability function $p_{\lambda}(\{v_i,h_a\})$. 
This is not the case: it is easy to construct functions $p_{\lambda}(\{v_i,h_a\})$ that obey (\ref{KeyRBM}), but are nothing 
like either the RBM or RG joint probability functions.
As an example, define
\bea
\rho(\{v_i\})=\frac{\tr_{h_a}\left(e^{T(\{v_i,h_a\})-H(\{v_i\})}\right)}{\mathcal{Z}},\cr\cr
\tilde\rho(\{h_a\})=\frac{\tr_{v_i}\left(e^{T(\{v_i,h_a\})-H(\{v_i\})}\right)}{\mathcal{Z}},
\eea
where
\begin{equation}
\mathcal{Z}=\sum_{v_i,h_a}e^{T(\{v_i,h_a\})-H(\{v_i\})}.
\end{equation}
We clearly have $\tr_{v_i}(\rho (\{v_i\}))=1=\tr_{h_a}(\tilde \rho(\{h_a\}))$  which implies that
\begin{equation}
A_\lambda (\{v_i,h_a\})=\tilde\rho(\{h_a\})\rho(\{v_i\}),
\end{equation}
obeys (\ref{KeyRBM}). It is quite clear that in $A_\lambda (\{v_i,h_a\})$ there are no correlations between 
the hidden and visible spins
\begin{equation}
\begin{split}
\langle v_j h_b\rangle =&\tr_{v_i,h_a}(v_j h_b\, A_\lambda (\{v_i,h_a\})\,)\\
=&\tr_{v_i} (\rho (\{v_i\}) v_j)\tr_{h_a}(\tilde\rho (\{h_a\}) h_b)\\
=&\langle v_j\rangle\langle h_b\rangle,
\end{split}
\end{equation}
so that we would reject it as a possible model of either the RG quantity
\bea
\mathcal{Z}^{-1}e^{T(\{v_i,h_a\})-H(\{v_i\})},
\eea
or of the RBM quantity
\bea
\mathcal{Z}^{-1} e^{-E(\{v_i,h_a\})}.
\eea
In addition to clarifying aspects of the argument of \cite{mehta2014exact}, the joint correlation functions between visible and 
hidden spins can be used to characterize the RG flow, as we now explain.
The RG flow ``coarse grains'' in position space: a ``block of spins'' is replaced by an effective spin, whose magnitude
is the average of the spins it replaces.
Since correlations between microscopic spins fall off with distance, an RG coarse graining implies that 
because the hidden spin is a linear combination of nearby visible spins, the correlation function between hidden and 
visible spins reflects a correlation between a hidden spin and a cluster of visible spins. 
This produces distinctive correlation functions, some examples of which are plotted in Figure \ref{fig:vihaplot_rg}.
We will search for this distinctive signal in the $\langle v_i h_a\rangle$ correlator, to find quantitative evidence that
deep learning is indeed performing an RG coarse graining.

\subsection{Numerical results}
Our numerical study aims to do two things: First, we establish whether there are RG-like patterns present within the 
correlator $\corr{v_ih_a}$, for correlators computed using the patterns generated by an RBM flow.
If these patterns are indeed present, this constitutes strong evidence in favor of the connection between RG and deep
learning.
{
The $\langle v_i h_a\rangle$ correlator is calculated using
\begin{equation}
    \langle v_i h_a\rangle = \frac{1}{N_s}\sum_{A=1}^{N_s}v_i^{(A)}h_a^{(A)},
\end{equation}
where $A=1,3,\dots,N_s$ with $N_s$ being the number of samples, $i=1,2,\dots,N_v$ labels the visible nodes within a visible vector and $a=1,2,3,\dots,N_h$ labels the nodes within a hidden vector.

For each hidden node $h_a$ we can produce a plot which shows how this hidden node is correlated to the $i=1,2,3,\dots,N_v$ visible nodes, $v_i$.
This gives us a total of $N_h$ plots.
Each panel within the plot for $h_a$ shows the $N_v$ correlation values for $\corr{v_ih_a}$.
By arranging these panels according to the lattice sites of the visible spins we get a grid of $L_v\times L_v=N_v$ values for the correlators $\corr{v_ih_a}$, where a is fixed for the given plot and $i$ runs from 1 to $N_v$.

By doing this we can determine if a given hidden node is correlated to a local patch of visible nodes which are neighbors on the original lattices produced from MC.
This local information is not encoded inherently in the RBM so learning about the nearest neighbour interactions present in the 2D Ising model would show promise that RBMs are performing a coarse graining related to that of RG.
}
We find that RG-like patterns do indeed emerge.

Second, according to the proposal of \cite{mehta2014exact}, in a deep network each layer that is stacked to produce the depth of 
the network performs one step in the RG flow.
With this interpretation in mind, it may be useful to compare how a network with multiple stacked RBMs learns as 
compared to a network with a single layer.
{ This issue is explored below.}

The training data is a set of $30000$ configurations of Ising model 32 by 32 lattices, near the critical temperature $T=2.269$.
The dataset is generated using Monte Carlo simulations.
An input lattice length of 32 allows a large enough final configuration even after two steps of RG, corresponding to 
stacking two RBMs.
In each step of the RG, the number of lattice sites is reduced by a factor of 4.
Thus, we flow from lattices with 1024 sites to lattices with 64 sites.
We enforce periodic boundary conditions.
To find signals of RG in the correlation functions the maximum distance between operators in a correlator 
must be large enough that the spin-spin correlation has dropped to zero.
We have confirmed that our lattice is large enough, judged by this criterion.

Having described the conditions of our numerical experiment, we consider the correlators $\corr{v_ih_a}$ generated when
the hidden neurons $h_a$ are generated from the visible neurons $v_i$ using RG.
Our goal is to understand the patterns appearing in correlation functions, that are a signature of the RG.
In Figure \ref{fig:rg_deci} the process of decimation used in our RG is explained.
The red dots, representing the visible lattice, are averaged (coarse grained) to produce the blue dots which define the lattice 
after a single step of the RG.
The four spin values located at the red dots surrounding each blue dot are averaged 
to obtain the value of the spin at the new (blue) lattice point.
This process clearly reduces the number of lattice sites by a factor of four.

Using the visible data which populates a 32 by 32 lattice, we populate lattices of size 16 by 16 and 8 by 8 spins by applying 
the RG and then calculate the various possible $\corr{v_ih_a}$ correlations. 

Figure \ref{fig:a-i-rg} shows the $\corr{v_ih_a}$ correlation function that results from a single RG step.
Each panel of the three Figures \ref{fig:a-i-rg}, \ref{fig:b-i-rg} and \ref{fig:c-i-rg}, shows how a given hidden spin is 
correlated with the visible spins.
We can clearly see a peak in correlation values around the spatial location of the hidden spin.
This is the signal of RG coarse graining: small spatially localized collections of spins are replaced by their average value.
We can go into a little more detail: the patches of large correlation in Figures \ref{fig:b-i-rg} and \ref{fig:c-i-rg} are 
larger in size than those of Figure \ref{fig:a-i-rg}.
This makes sense since each step of the RG implies ever larger spatial regions of the spins are being averaged to produce 
the coarse grained variables.
The fact that the spins that are averaged are spatially localized is a direct consequence of the fact that the Ising model
Hamiltonian is local in space so that spatially adjacent spins have similar behaviors.
In more general big data settings it may be harder to decide if the coarse graining is RG-like or not, since it might not
be clear what is meant by spatial locality. 
\begin{figure}[h!]
    \centering
\renewcommand{\thesubfigure}{a-i}
\subcaptionbox{\label{fig:a-i}}{\includegraphics[trim={0 0 0 0},clip,width=0.234\textwidth]{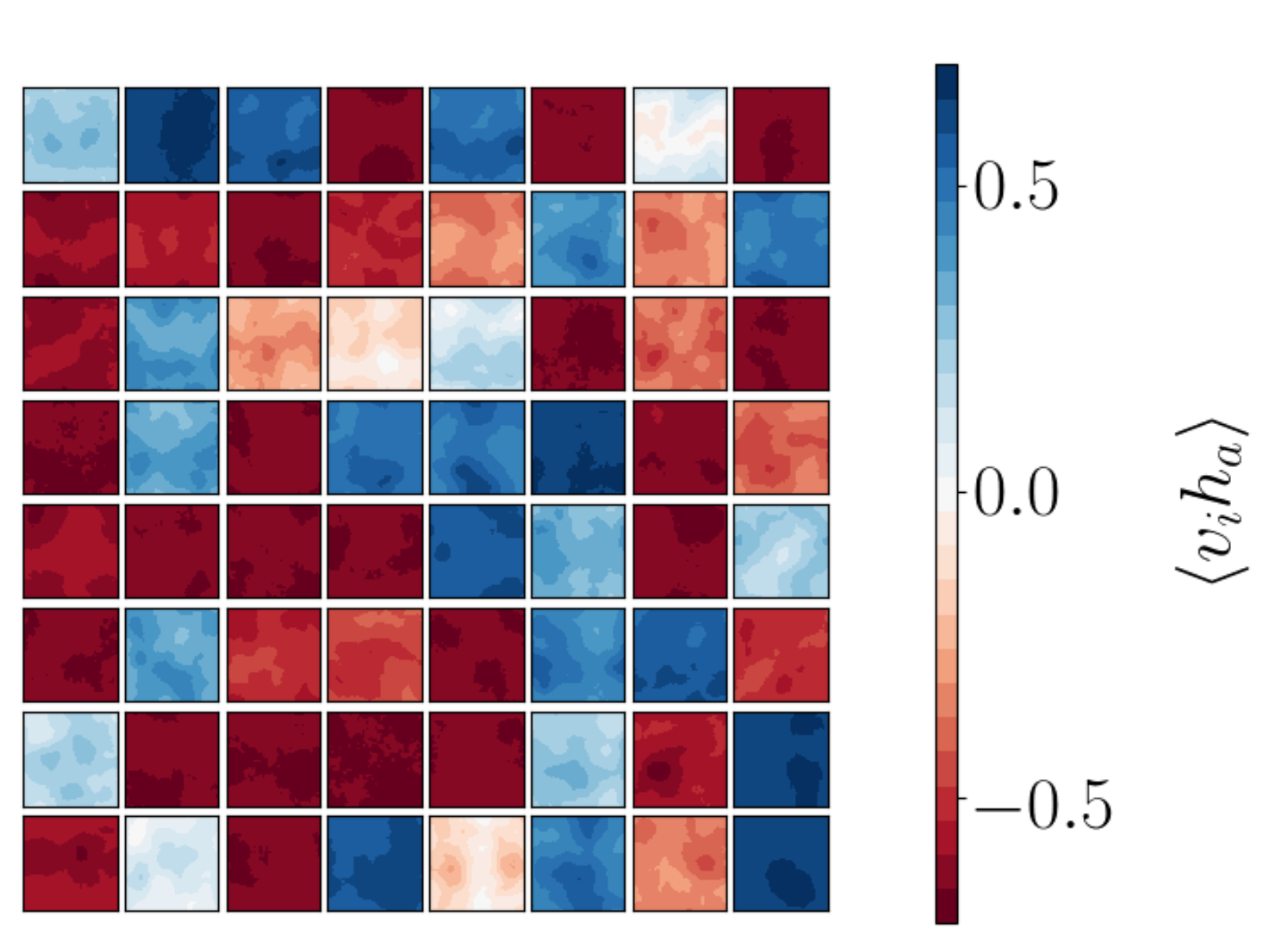}}~
\renewcommand{\thesubfigure}{a-ii}
\subcaptionbox{\label{fig:b-i}}{\includegraphics[trim={0 0 0 0},clip,width=0.234\textwidth]{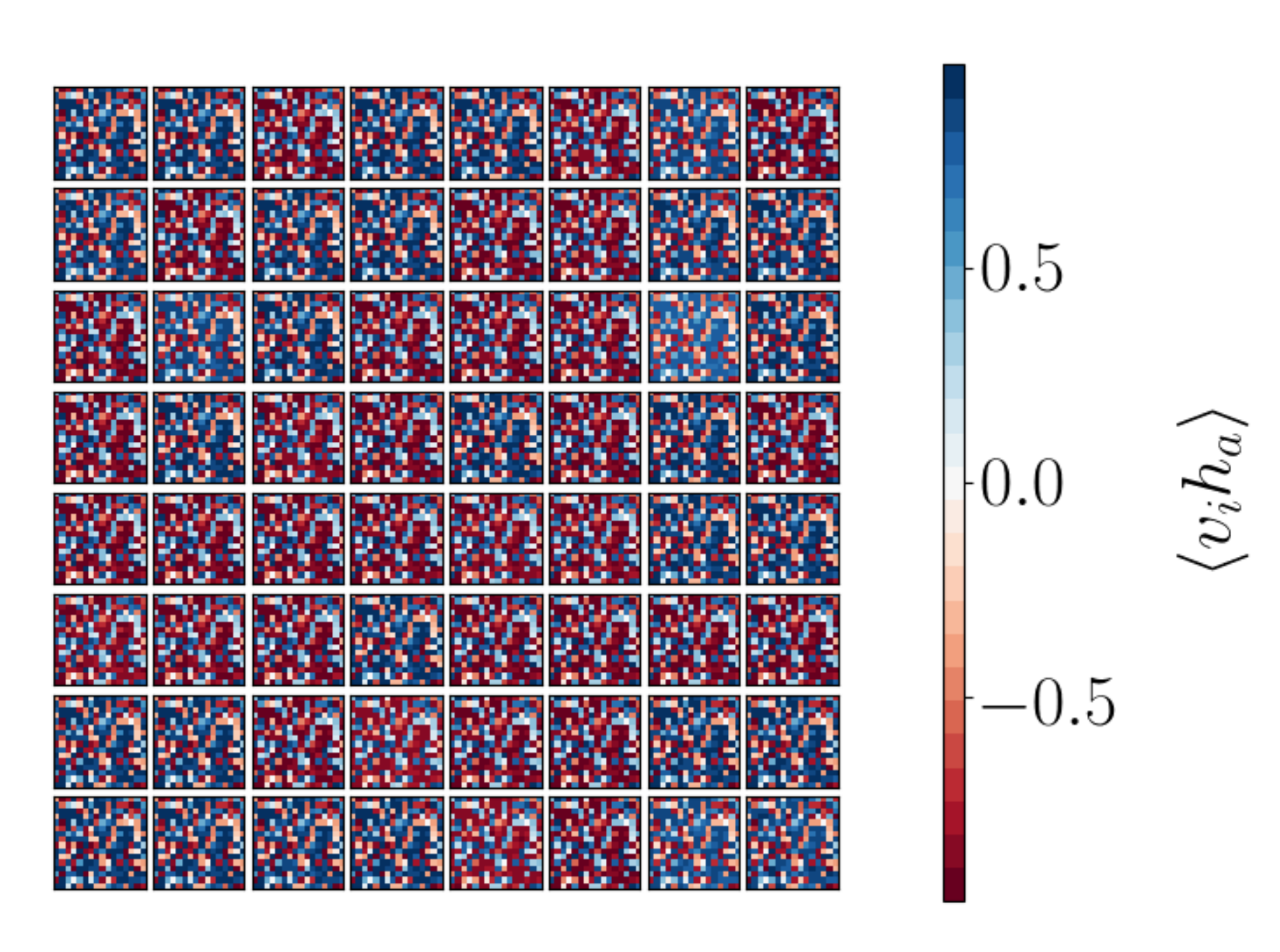}}
\renewcommand{\thesubfigure}{a-iii}
\subcaptionbox{\label{fig:c-i}}{\includegraphics[trim={0 0 0 0},clip,width=0.234\textwidth]{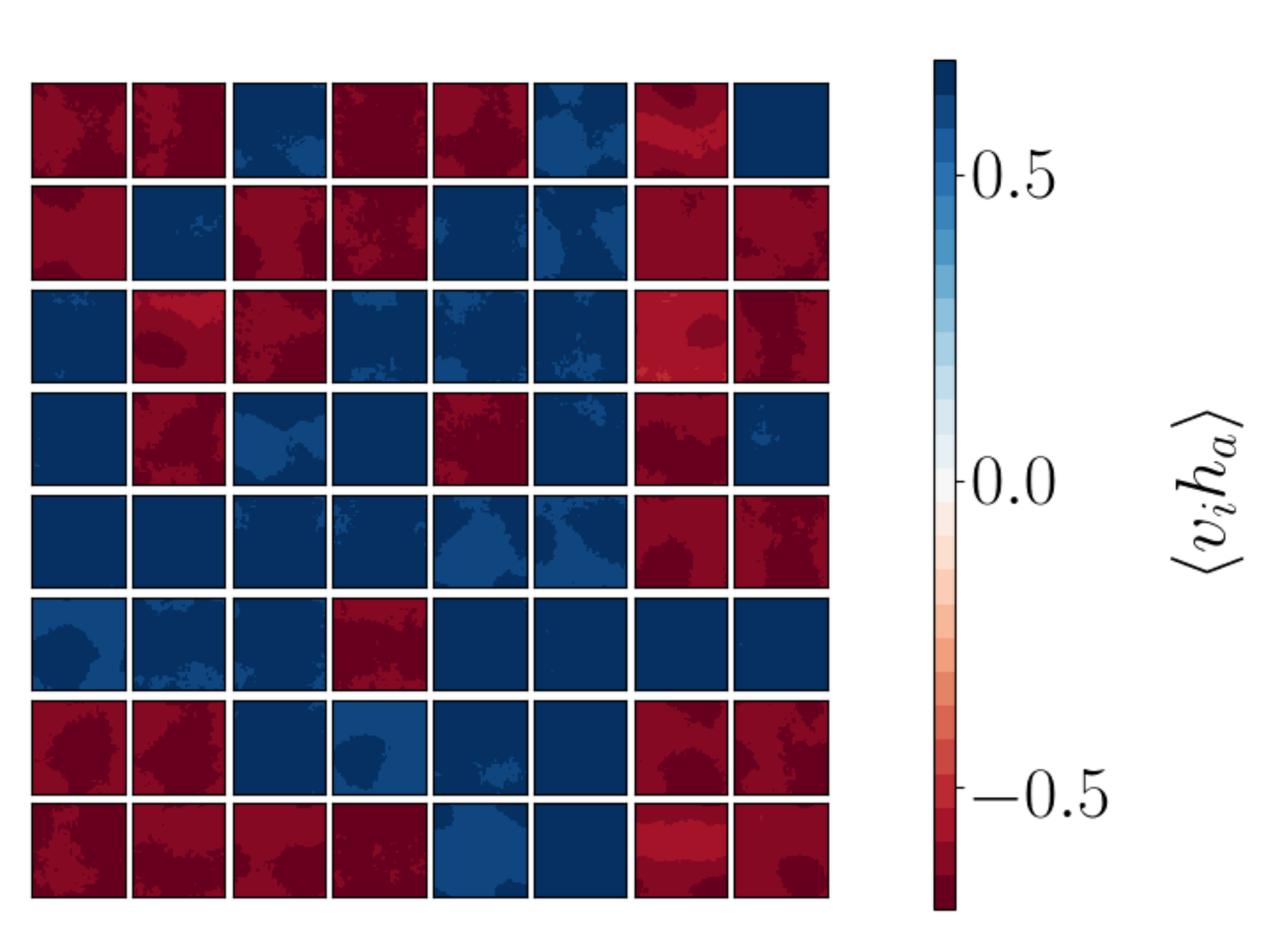}}~
\renewcommand{\thesubfigure}{b}
\subcaptionbox{\label{fig:vihaplot_rbm_single}}{\includegraphics[trim={0 0 0 0},clip,width=0.234\textwidth]{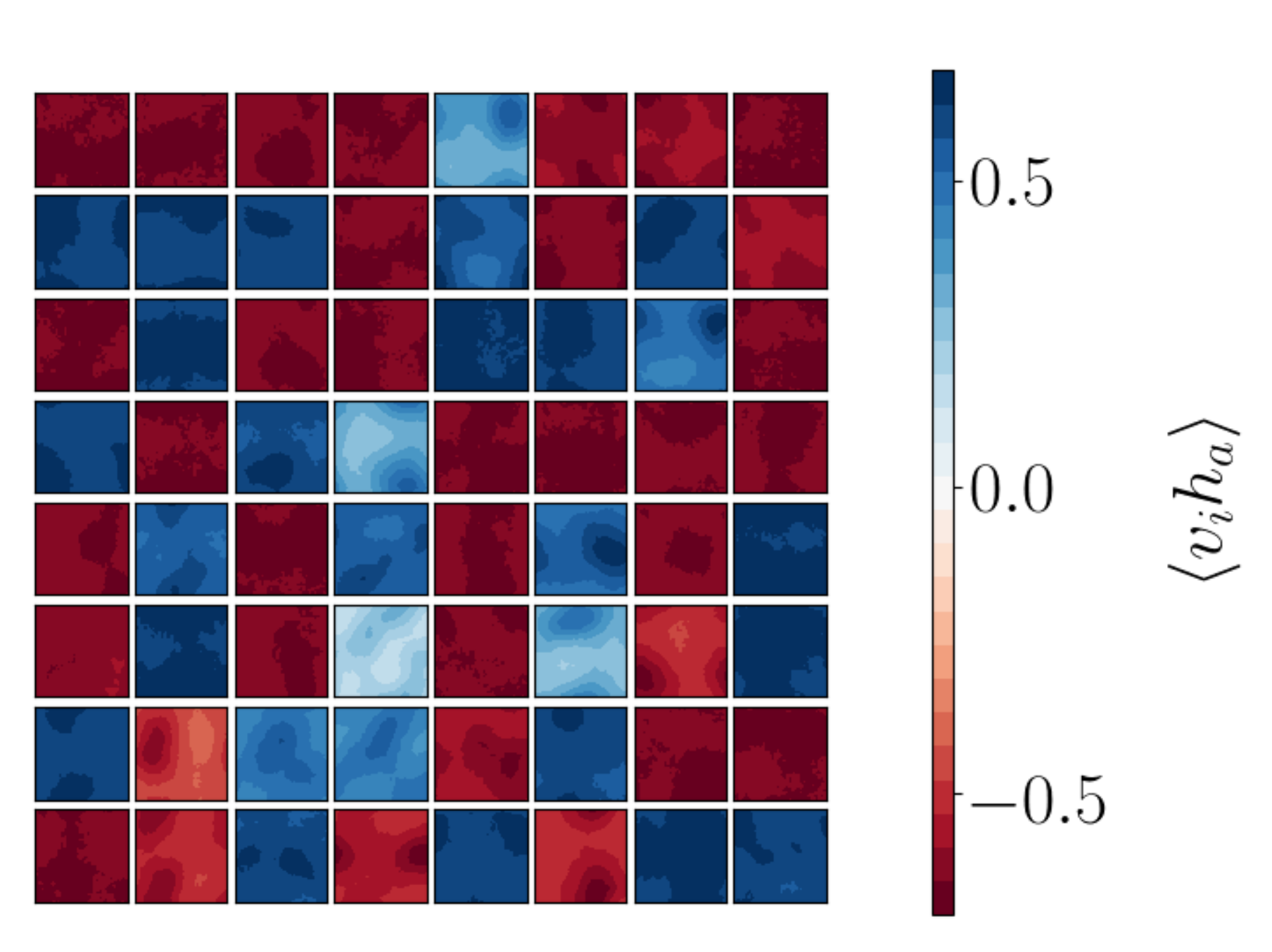}}
    \caption{Plots showing the correlation values for (a) the stacked RBMs various layers and (b) the single RBM. 
(a-i) shows correlations between visible Ising data (1024 nodes) at $T_c$ and outputs from the first stacked RBM (256 nodes).
(a-ii) shows correlations between outputs from the first stacked RBM (256 nodes) and outputs from the second stacked RBM (64 nodes).
(a-iii) shows correlations between visible Ising data (1024 nodes) at $T_c$ and outputs from the second stacked RBM (64 nodes).
(b) shows correlations between visible Ising data (1024 nodes) at $T_c$ and outputs from the single RBM (64 nodes).}
    \label{fig:vihaplot_rbm_stacked}
\end{figure}

Having established the signal characteristic of the RG flow, we will now search for this signal in the $\corr{v_ih_a}$ 
correlators computed using the configurations generated from the RBM flow. 
We consider configurations generated by a stacked network with an RBM having 1024 visible nodes and 256 hidden nodes
cascading into a second RBM having 256 visible nodes and 64 hidden nodes. 
We also consider configurations generated by a single RBM network with 1024 visible nodes and 64 hidden nodes.
The factor of $4$ relating the number of visible to hidden nodes is chosen to mimic the decimation of lattice
sites in each step of the RG.
The networks are trained on the same data used as input for the RG considered above.
Training is through $10000$ steps of contrastive divergence \cite{carreira2005contrastive}.

Figures \ref{fig:a-i} to \ref{fig:c-i} show plots for the stacked RBM and Figure \ref{fig:vihaplot_rbm_single} for the 
single RBM network.
Figure \ref{fig:c-i} shows the correlation functions between the visible vectors input to the first network in the stack and the final hidden vectors output from the stack and is to be compared to the corresponding RG result in Figure 
\ref{fig:c-i-rg}.
The two patterns are very similar suggesting that the trained RBM is indeed performing something like the RG
coarse graining.

{

To quantitatively compare the patterns we observe in the $\langle v_ih_a \rangle$ correlators produced by RG to those produced by the RBM we make use of a two point correlation function.
When we perform an RG coarse graining we average local nearby nodes from the input (visible lattice) to obtain the output (hidden lattice).
This local averaging is encoded in the $\corr{v_ih_a}$ plots by bright spots.
The bright spots correspond to a specific hidden node being highly correlated to a patch of local visible spins.
In each $\langle v_ih_a \rangle$ plot, the hidden node we consider is fixed and we plot its correlation with all visible nodes.
If we denote each value of $\langle v_ih_a \rangle$ by $x_i$ we calculate the two point correlator $\langle x_i x_j \rangle$ between values $\langle v_ih_a \rangle$ and $\langle v_jh_a \rangle$ summed over all hidden nodes

\begin{equation}
    \langle x_i x_j \rangle =\frac{1}{N_h} \sum_{a=1}^{N_h}\langle v_ih_a \rangle \times \langle v_jh_a \rangle.
\end{equation}
By calculating this quantity, we learn about the size of the correlated patches in the $\langle v_ih_a\rangle$ plots.
We can plot the value of $\langle x_i x_j \rangle$ versus the distance, $|i-j|$. 
This quantity tells us important information about the size of the correlated patches.
We average the values of $\corr{x_ix_j}$ where the distances $|i-j|$ are equal. 
The patches present in $\corr{v_ih_a}$ will thus be detected regardless of where they appear in the plot.
If we do have local patches of high correlation, $\corr{x_ix_j}$ will be peaked at short distances and as distance increases, $\corr{x_ix_j}$ will decrease in value. 

For RG, the plots seen in Figure \ref{RG_corrs} show a linear fall off in the correlator as distance increases. 
The fall off of these correlators is in the order of magnitude of $10^{-4}$.
In Figure \ref{RBM_corrs} the behavior of the RBM correlator is shown. 
Figures \ref{fig:rbm_1024_256} and \ref{fig:rbm_1024_64} show similar behavior to that seen for RG.
There is a linear decrease in $\corr{x_ix_j}$ in the same order of magnitude of $10^{-4}$.
A difference between these plots is that the RBM correlators are offset.
We do not have an explanation for this offset.

\begin{figure}[t!]
    \centering
    \begin{subfigure}{0.245\textwidth}
        \includegraphics[width=\textwidth]{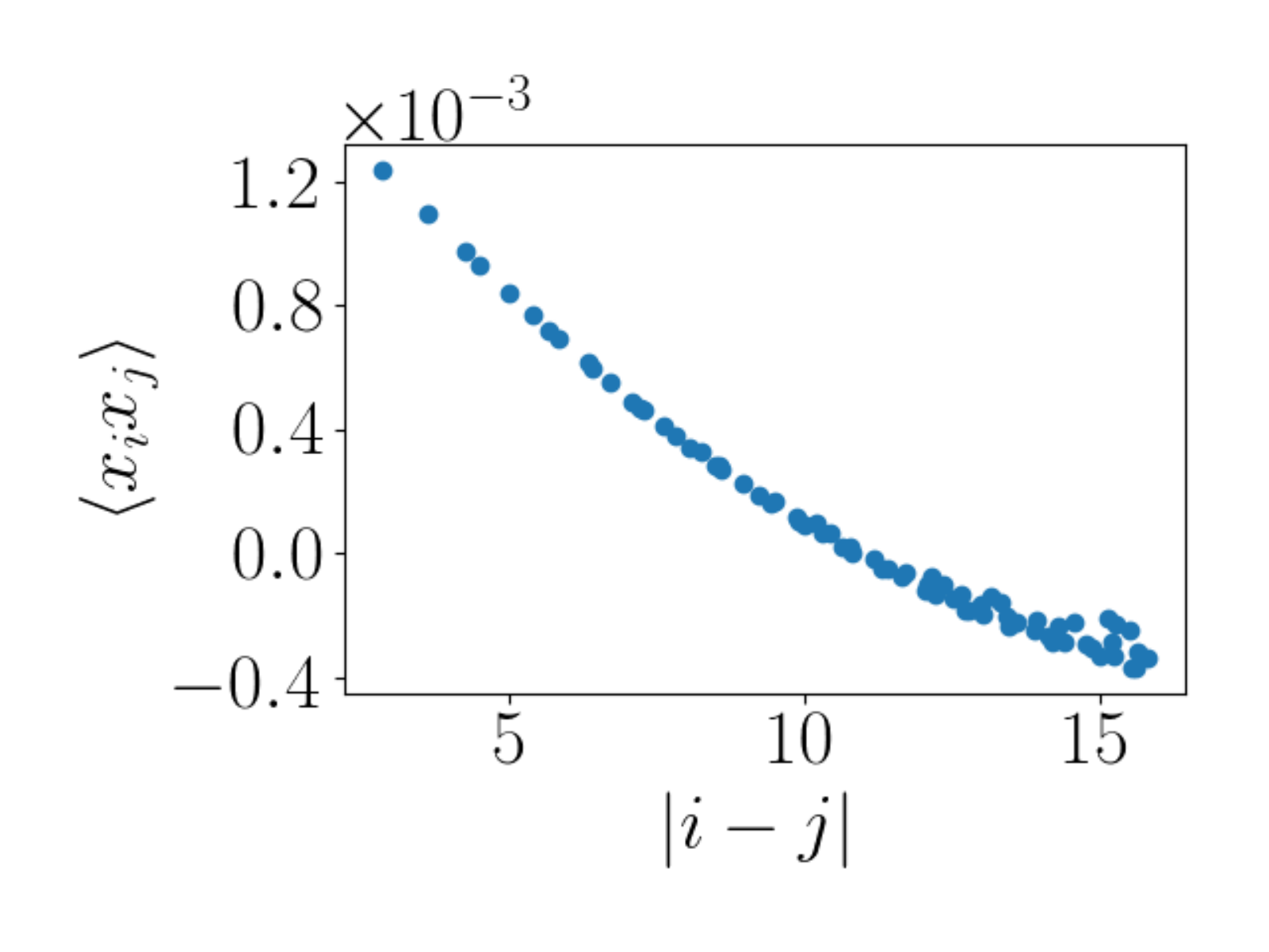}
        \caption{Two point correlator versus distance for Figure \ref{fig:a-i-rg}.}
        \label{fig:RG_1024_256}
    \end{subfigure}~~
    \begin{subfigure}{0.245\textwidth}
        \includegraphics[width=\textwidth]{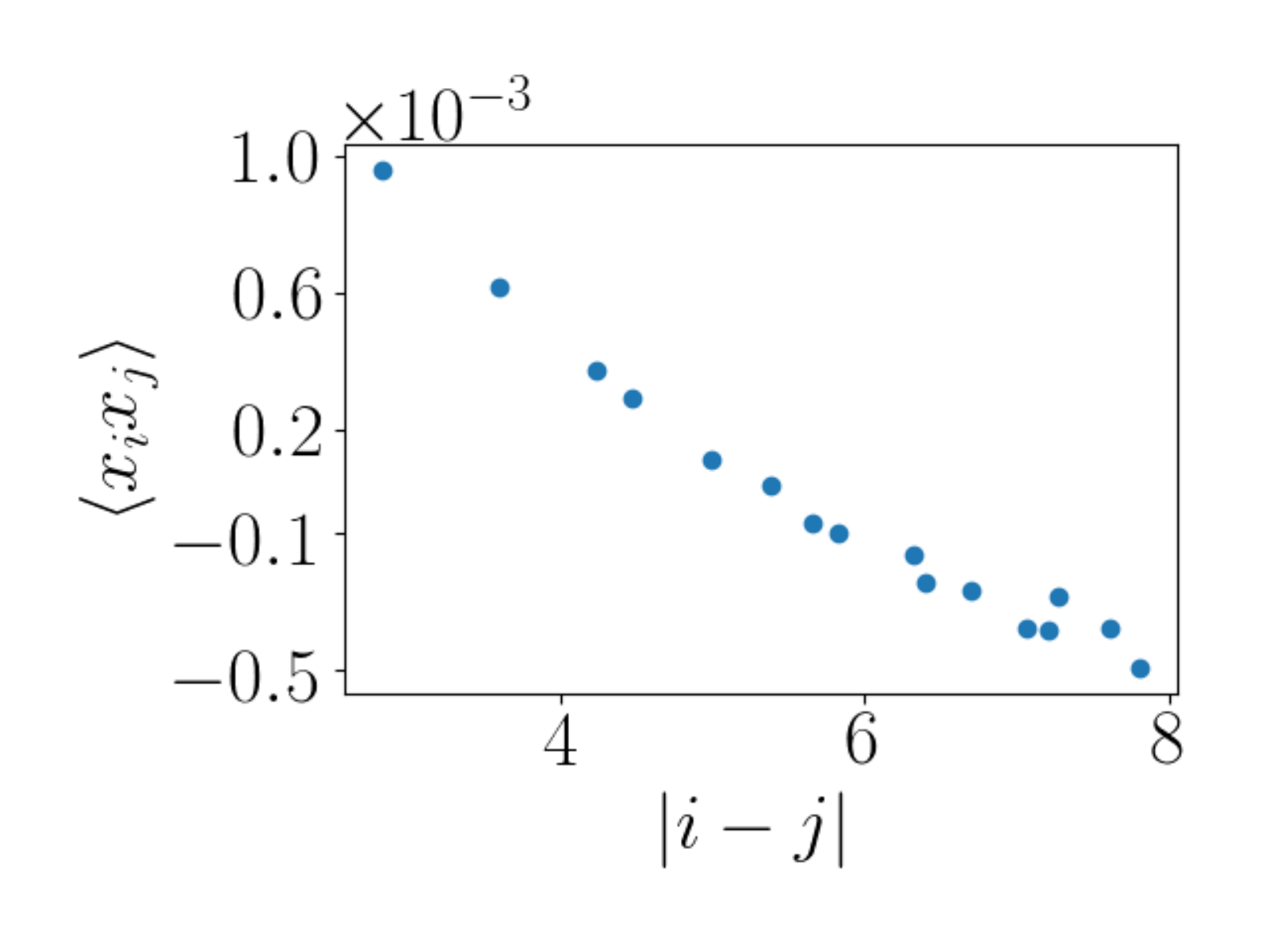}
        \caption{Two point correlator versus distance for Figure \ref{fig:b-i-rg}.}
        \label{fig:RG_256_64}
    \end{subfigure}
    \begin{subfigure}{0.245\textwidth}
        \includegraphics[width=\textwidth]{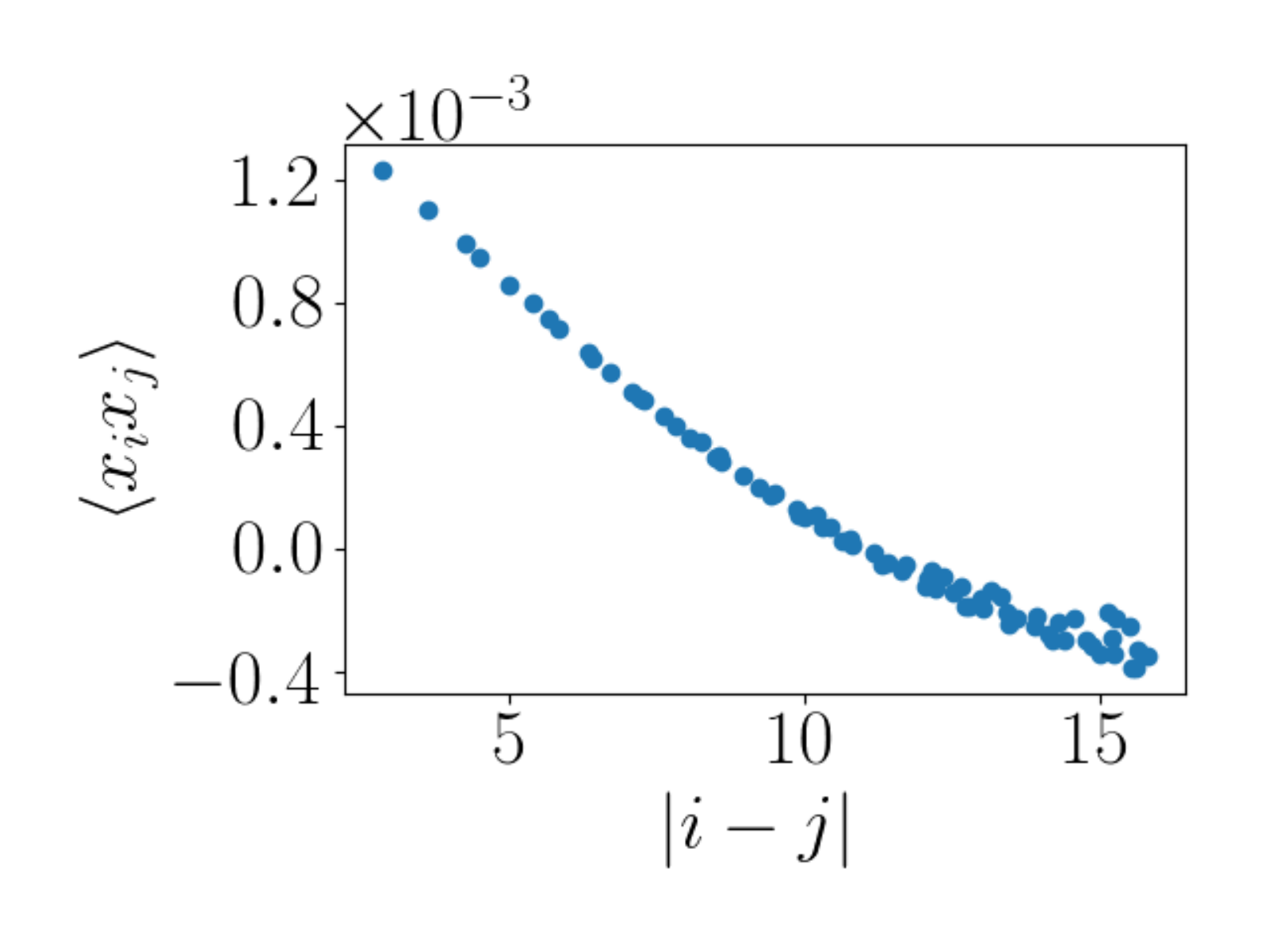}
        \caption{Two point correlator versus distance for Figure \ref{fig:c-i-rg}.}
        \label{fig:RG_1024_64}
    \end{subfigure}
    \caption{Plots showing $\langle x_i x_j \rangle $ for RG $\langle v_i h_a \rangle$ plots shown in Figures \ref{fig:a-i-rg}, \ref{fig:b-i-rg} and \ref{fig:c-i-rg}.}
    \label{RG_corrs}
\end{figure}

\begin{figure}[t!]
    \centering
    \begin{subfigure}{0.23\textwidth}
        \includegraphics[width=\textwidth]{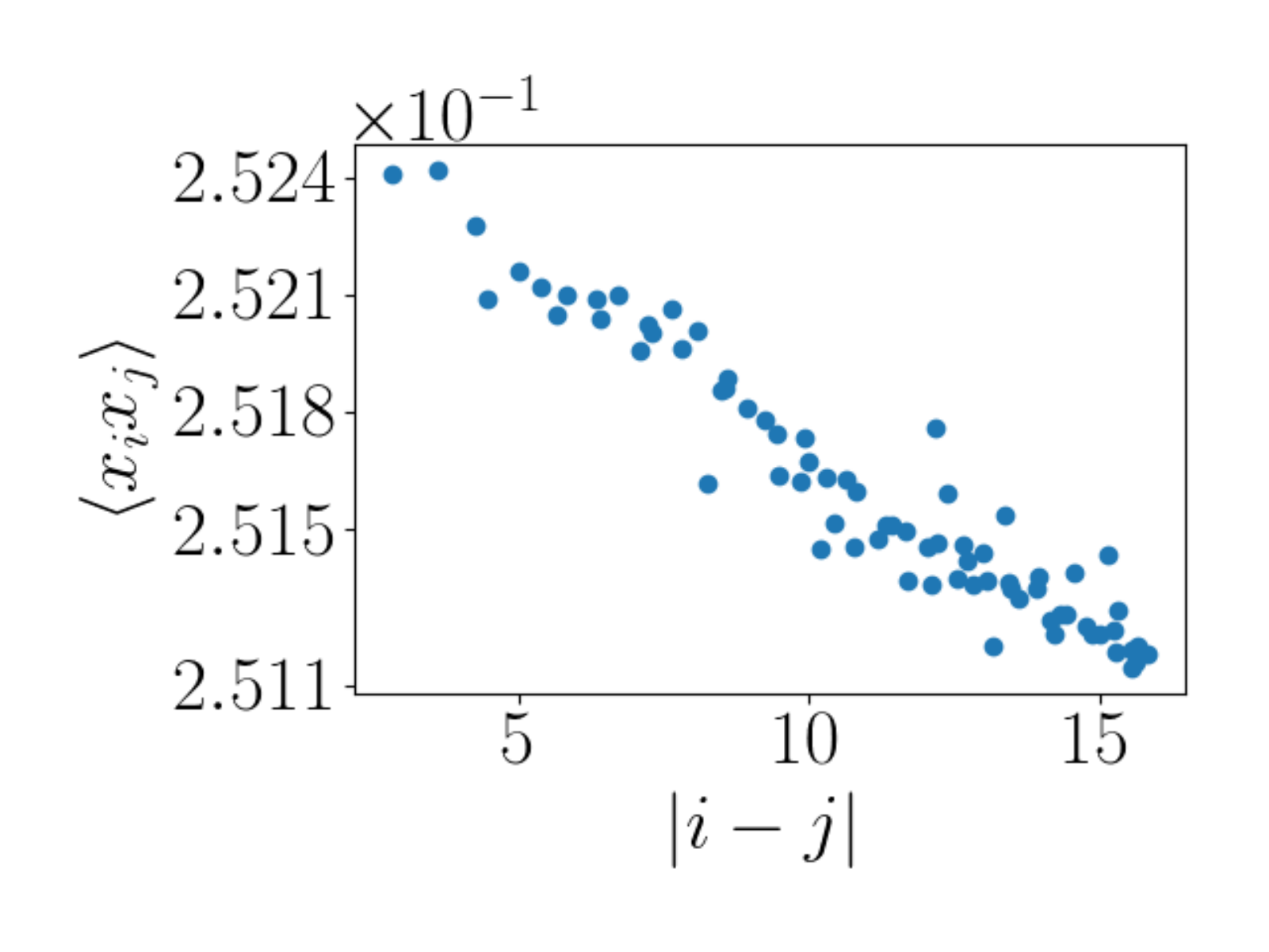}
        \caption{Two point correlator versus distance for Figure \ref{fig:a-i}.}
        \label{fig:rbm_1024_256}
    \end{subfigure}~~
    \begin{subfigure}{0.23\textwidth}
        \includegraphics[width=\textwidth]{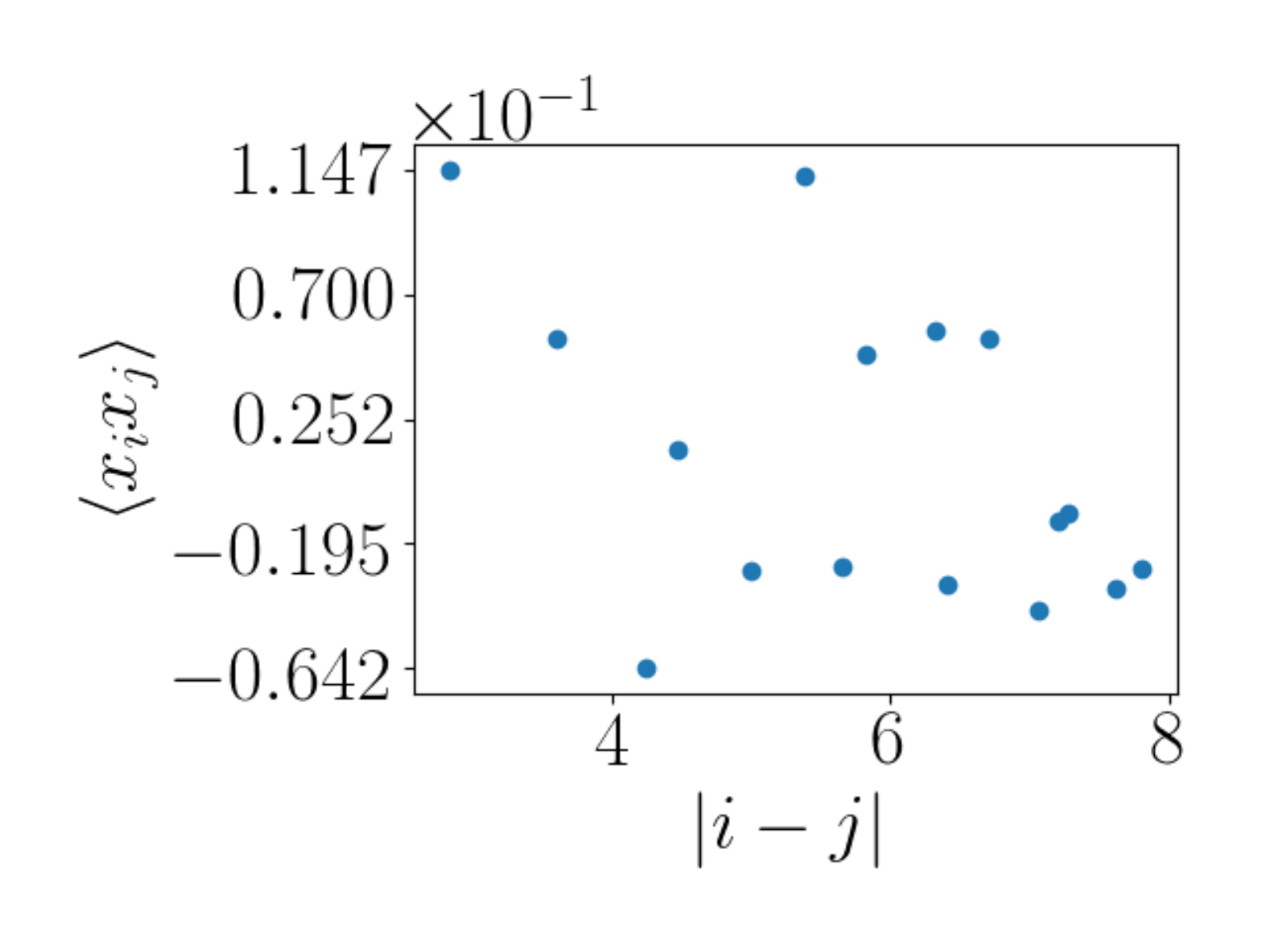}
        \caption{Two point correlator versus distance for Figure \ref{fig:b-i}.}
        \label{fig:rbm_256_64}
    \end{subfigure}
    \begin{subfigure}{0.23\textwidth}
        \includegraphics[width=\textwidth]{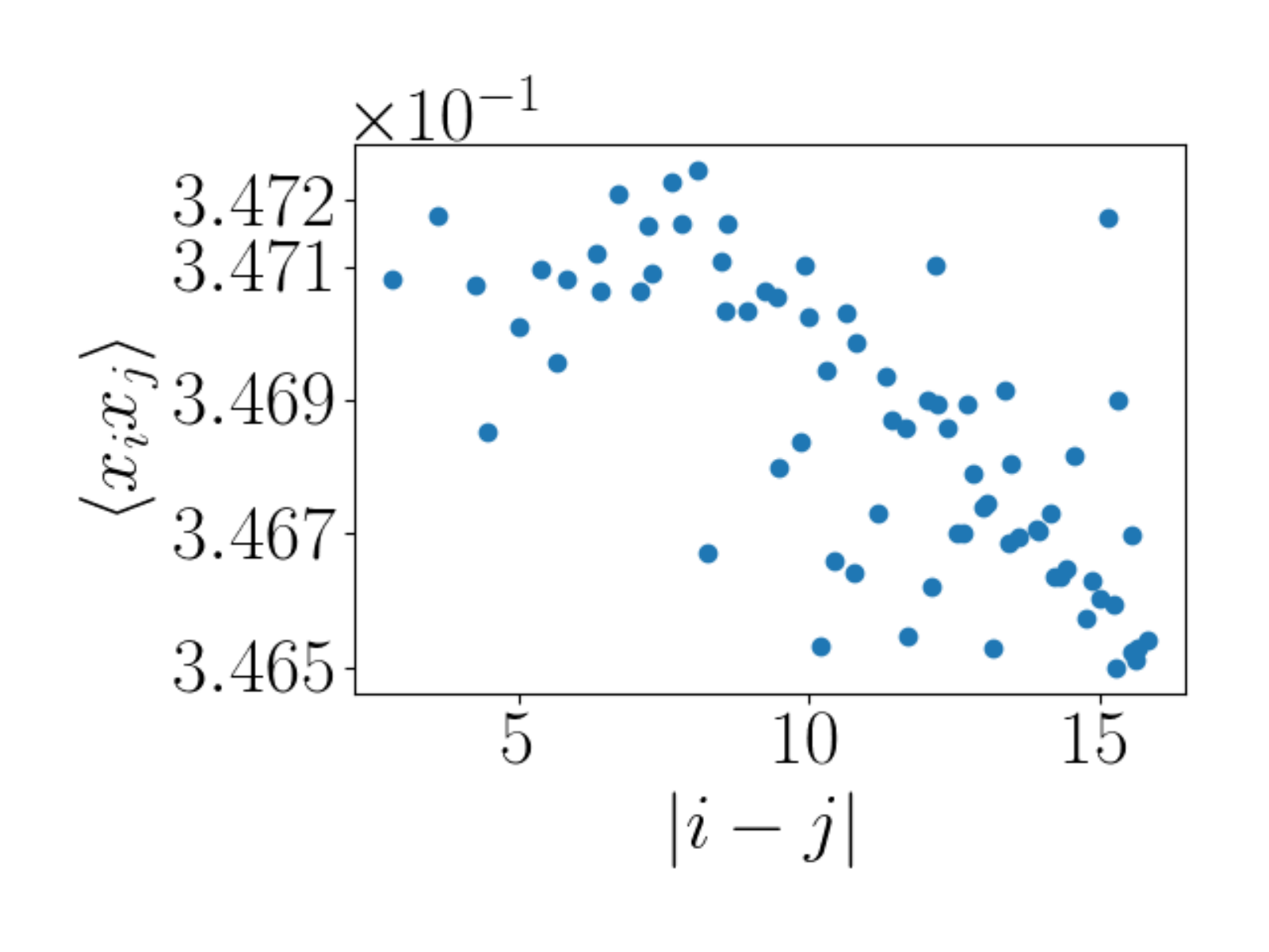}
        \caption{Two point correlator versus distance for Figure \ref{fig:c-i}.}
        \label{fig:rbm_1024_64}
    \end{subfigure}
    \caption{Plots showing $\langle x_i x_j \rangle $ for RBM $\langle v_i h_a \rangle$ plots shown in Figures \ref{fig:a-i}, \ref{fig:b-i} and \ref{fig:c-i}.}
    \label{RBM_corrs}
\end{figure}

We also study $\corr{x_ix_j}$ where the visible lattice is of size $48\times 48=2304$ and the hidden lattice is of size $24\times 24 = 576$.
The $\corr{v_ih_a}$ correlators are determined using a visible set of 40000 lattices at $T_c=2.269$ and the hidden set produced by an application of RG and by applying a trained RBM (which is trained on this same visible data set).
The results for $\corr{x_ix_j}$ are shown in Figure \ref{fig:vh_two_pt_2304_576}.
For the RBM (Figure \ref{fig:RBM_two_pt_2304_576}) the fall off of the correlator again matches the behavior of RG (Figure \ref{fig:RG_two_pt_2304_576}).
For the RBM the fall off trend is clearer with these larger lattice sizes when compared to the RBMs of smaller lattice sizes.
There is a slight increase in correlation in Figure \ref{fig:RBM_two_pt_2304_576} as the distance nears $L_v/2=24$.
This suggests that there are more than 1 local patches of correlation in the RBM $\corr{v_ih_a}$ plots.
In Figure \ref{fig:RBM_vh_2304_576} we can see that some plots show some speckle with a few highly correlated spots in a single $\corr{v_ih_a}$ plot.
We verify this observation by considering specific $\corr{v_ih_a}$ correlation patterns below.

\begin{figure}[t!]
    \centering
    \begin{subfigure}{0.25\textwidth}
        \includegraphics[width=\textwidth]{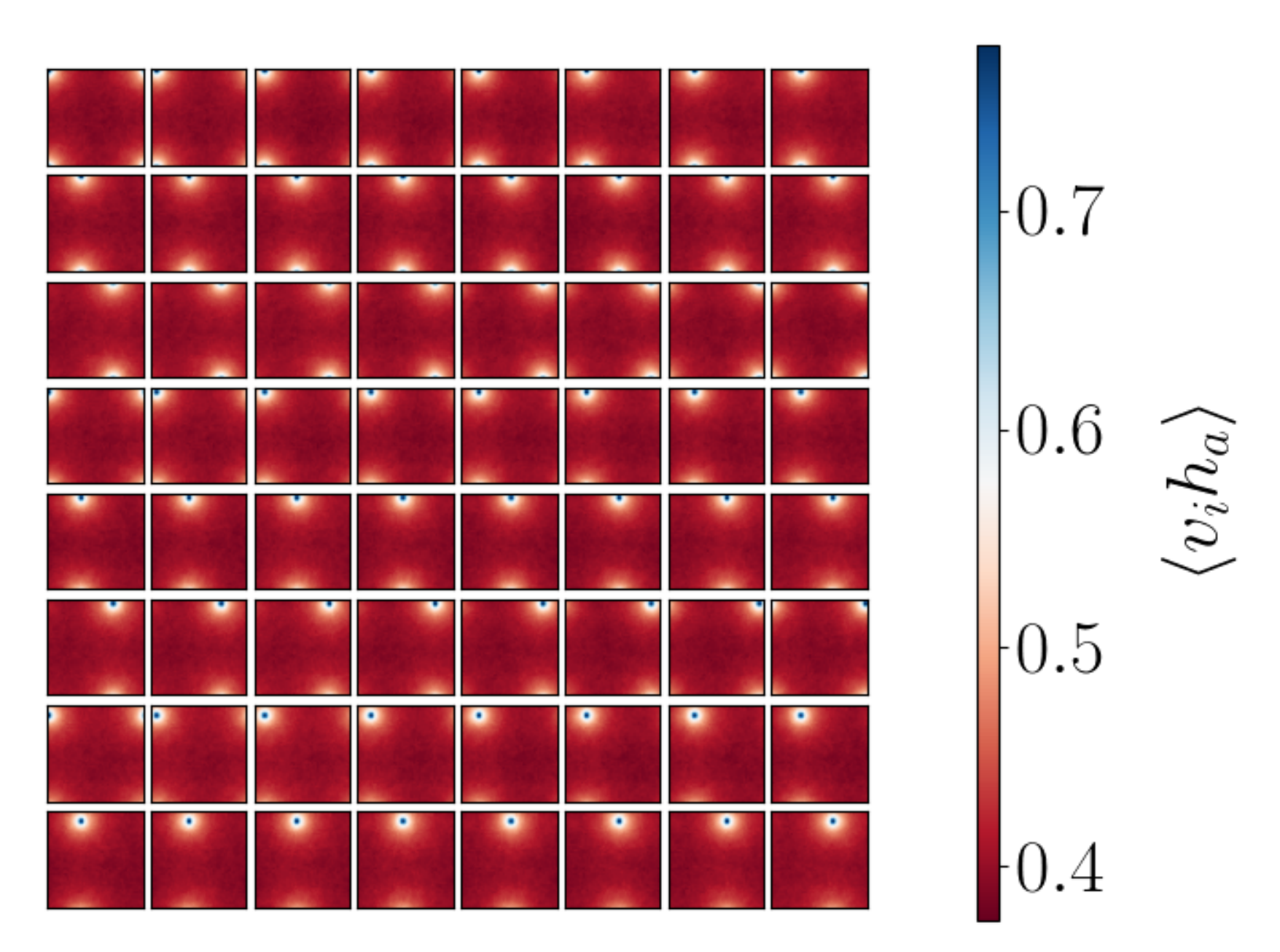}
        \caption{RG $\corr{v_ih_a}$.}
        \label{fig:RG_vh_2304_576}
    \end{subfigure}~~
    \begin{subfigure}{0.25\textwidth}
        \includegraphics[width=\textwidth]{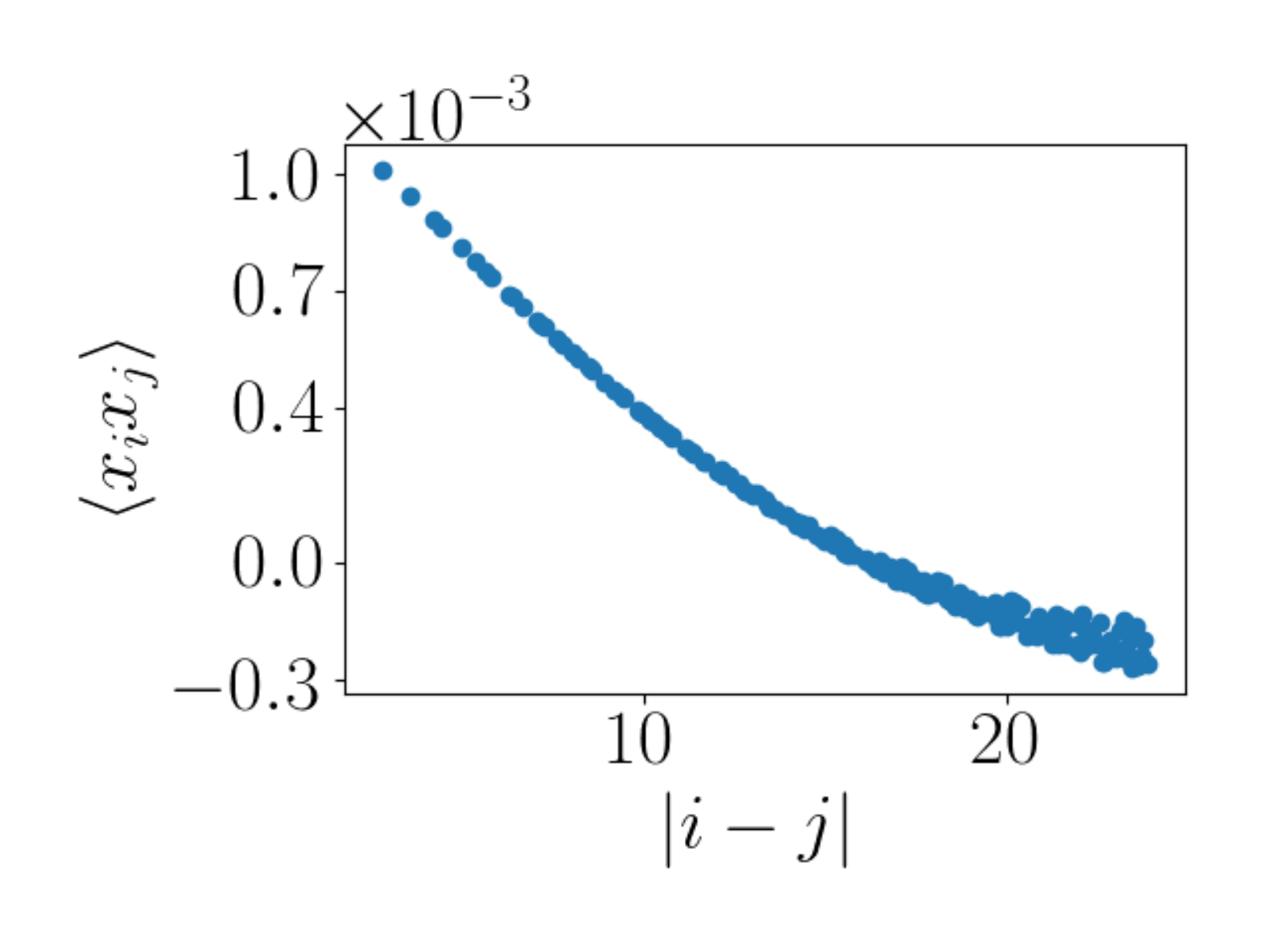}
        \caption{RG $\corr{x_ix_j}$.}
        \label{fig:RG_two_pt_2304_576}
    \end{subfigure}
    \begin{subfigure}{0.25\textwidth}
        \includegraphics[width=\textwidth]{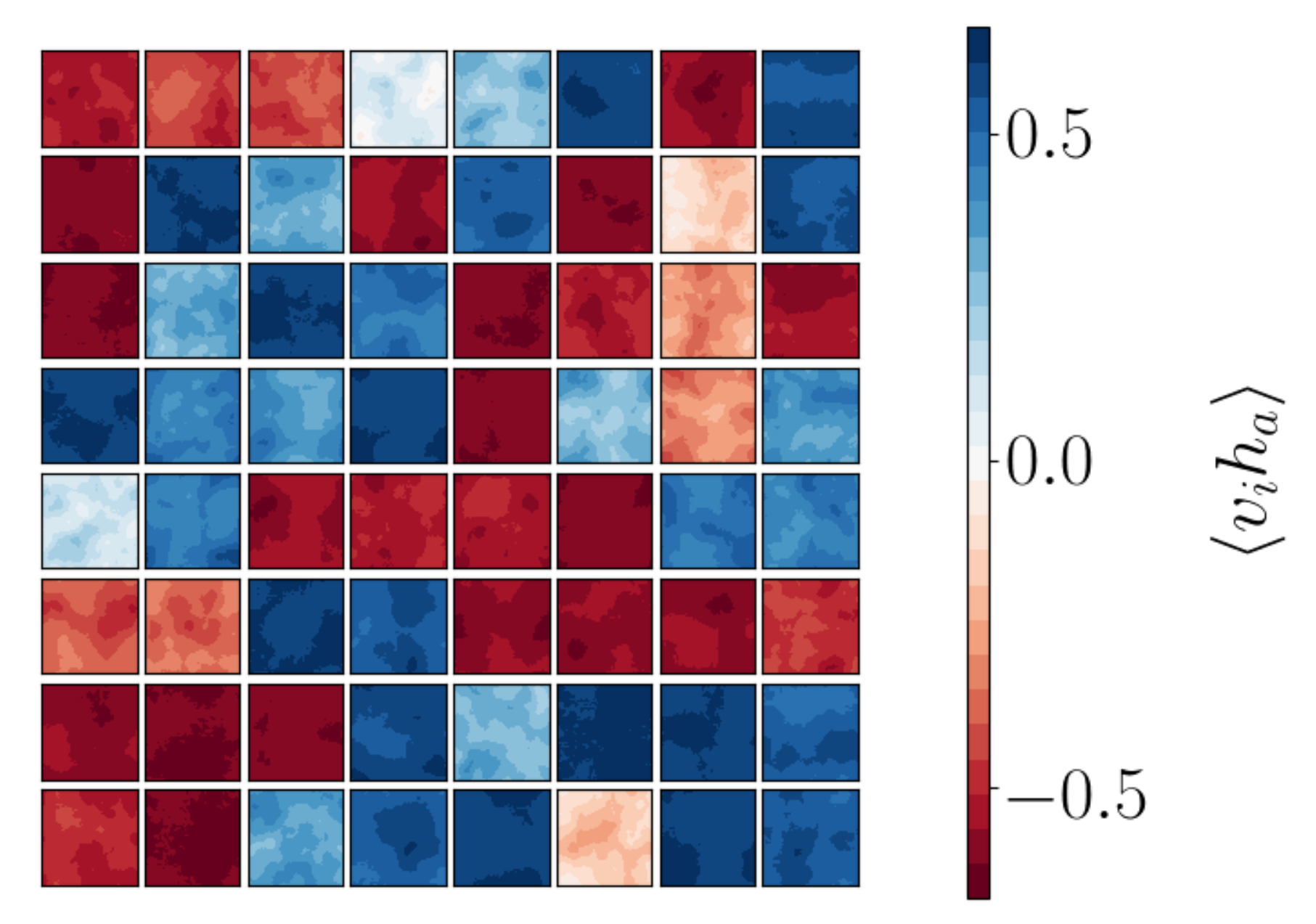}
        \caption{RBM $\corr{v_ih_a}$.}
        \label{fig:RBM_vh_2304_576}
    \end{subfigure}~~
    \begin{subfigure}{0.25\textwidth}
        \includegraphics[width=\textwidth]{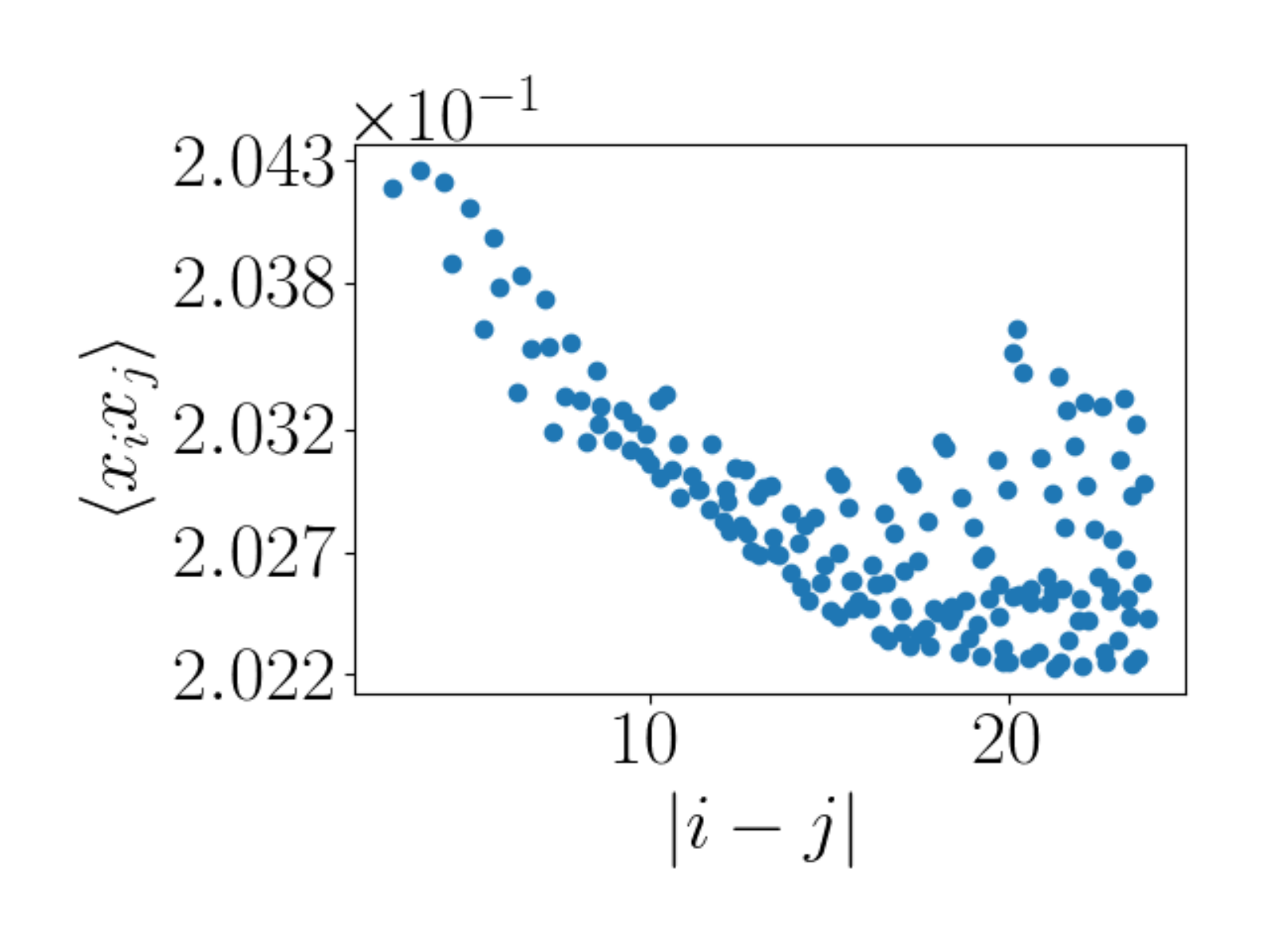}
        \caption{RBM $\corr{x_ix_j}$.}
        \label{fig:RBM_two_pt_2304_576}
    \end{subfigure}
    \caption{Plots showing the $\corr{v_ih_a}$ correlators and corresponding two point correlator $\corr{x_ix_j}$ values for one step of RG and a single RBM starting from an input lattice of size $48\times 48 = 2304$ which is reduced to an output lattice of size $24 \times 24=576$.}
    \label{fig:vh_two_pt_2304_576}
\end{figure}

To gain more understanding of the information encoded in the two point correlator we consider $\langle v_ih_a\rangle$ patterns of white noise in addition to a checkerboard shape with various sizes for the sub-blocks on the checkerboard.
This allows us to explore the benefit of studying $\langle x_ix_j\rangle$ in probing patterns present in $\langle v_ih_a \rangle$.
We show an example of a single hidden node's correlation with all visible nodes constructed using white noise in Figure \ref{fig:vh_wn}.
In Figure \ref{fig:vh_corr_wn} we can see $\corr{x_ix_j}$ calculated from the values shown in Figure \ref{fig:vh_wn}. 
We see different behavior to that observed in Figures \ref{RG_corrs} and \ref{RBM_corrs}.
As expected, there is no clear relationship between the value of the two point correlator $\corr{x_ix_j}$ and the distance between values $x_i$ and $x_j$.

\begin{figure}[t!]
    \centering
    \begin{subfigure}{0.225\textwidth}
        \includegraphics[width=\textwidth]{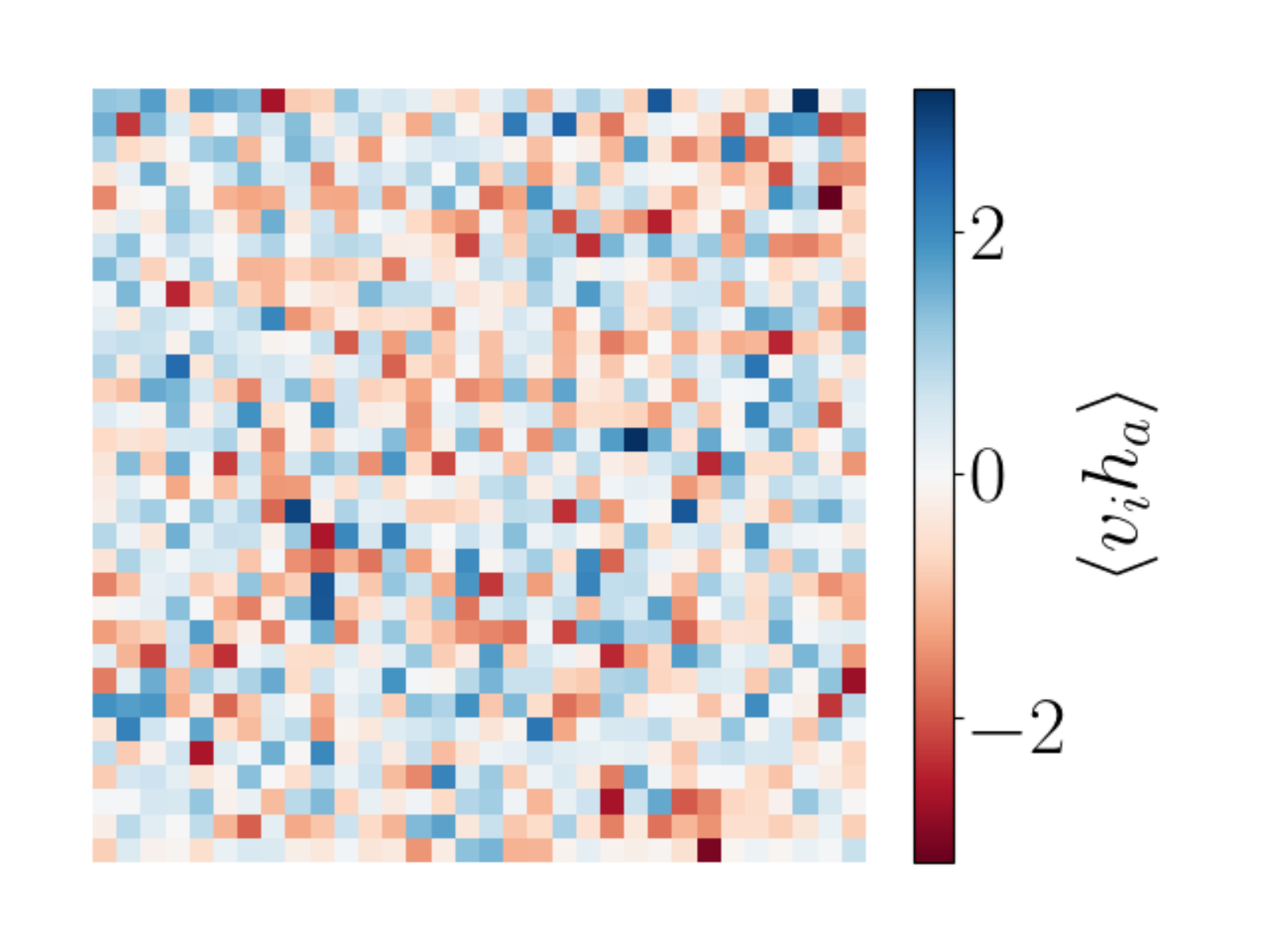}
        \caption{}
        \label{fig:vh_wn}
            \end{subfigure}~~
    \begin{subfigure}{0.225\textwidth}
\includegraphics[width=\textwidth]{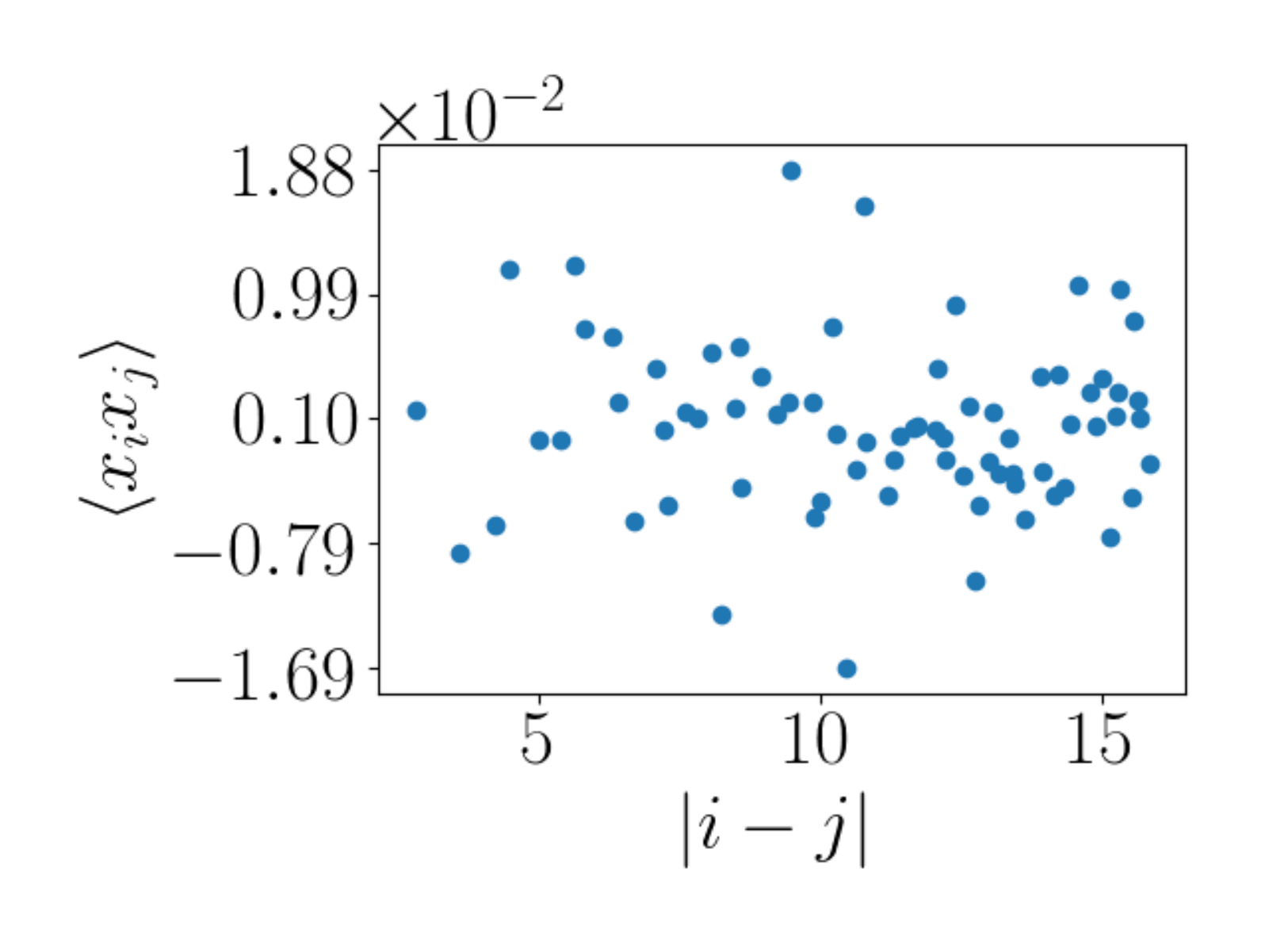}
\caption{}
\label{fig:vh_corr_wn}
    \end{subfigure}
    \caption{White noise: Plots showing (a) a hypothetical $\langle v_ih_a\rangle$ correlator (for a single hidden node with all visible nodes) consisting of white noise and (b) the two point correlator $\langle x_i x_j \rangle$ calculated from the values of $\langle v_ih_a\rangle$ in (a).}
    \label{fig:wn_corrs}
\end{figure}

We also study $\corr{v_ih_a}$ with a checkerboard pattern as shown in Figure \ref{fig:checkerboard}.
We explore various sub-block sizes within the checkerboard pattern.
In Figures \ref{fig:block_4}, \ref{fig:block_8} and \ref{fig:block_16} we show the $\corr{v_ih_a}$ plot with a checkerboard pattern on a lattice of size $32\times 32$ with sub-blocks of size 4 by 4, 8 by 8 and 16 by 16 respectively. 
The corresponding two point correlators are shown in Figures \ref{fig:block_4_two_point}, \ref{fig:block_8_two_point} and \ref{fig:block_16_two_point}.
We can see from these plots that having many correlated patches in $\corr{v_ih_a}$ which are of size $<L_v/2$, produces a two point correlator which is peaked at a number of points.
In the case of Figure \ref{fig:block_16_two_point}, where the sub-block sizes equal $L_v/2$ we see similar behavior to that seen in the RBM and RG correlator plots.
The additional peaks seen in Figures \ref{fig:block_4_two_point} and  \ref{fig:block_8_two_point} are due to multiple patches in the image being correlated.
This behavior is not characteristic of the RG local patches as a single highly correlated patch is present in the RG $\corr{v_ih_a}$ plots.

\begin{figure}[t!]
    \centering
    \begin{subfigure}{0.25\textwidth}
        \includegraphics[width=\textwidth]{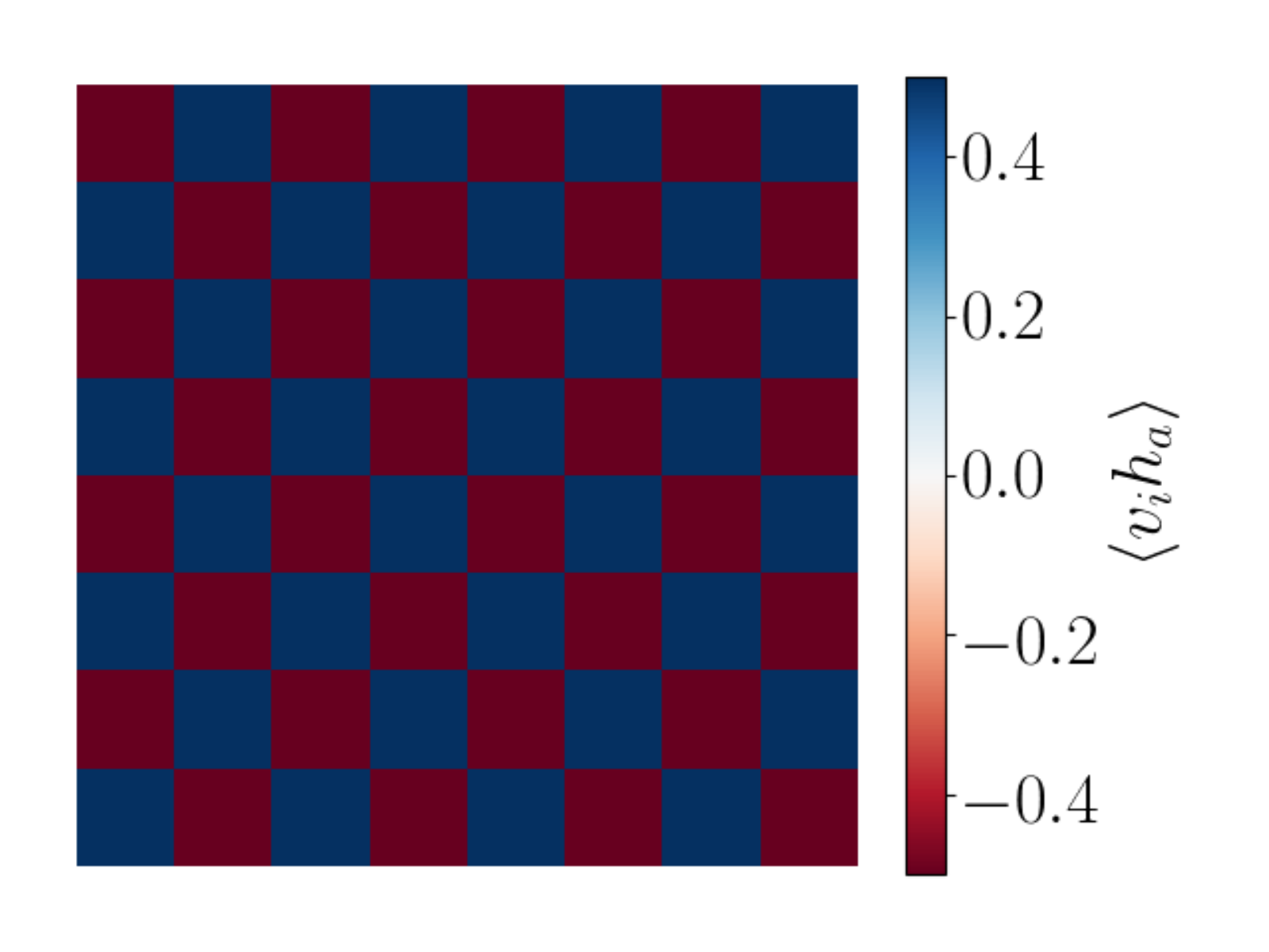}
        \caption{$\langle v_ih_a\rangle$ grid of $32 \times 32$ nodes with sub-blocks of size 4.}
        \label{fig:block_4}
            \end{subfigure}~~
    \begin{subfigure}{0.25\textwidth}
\includegraphics[width=\textwidth]{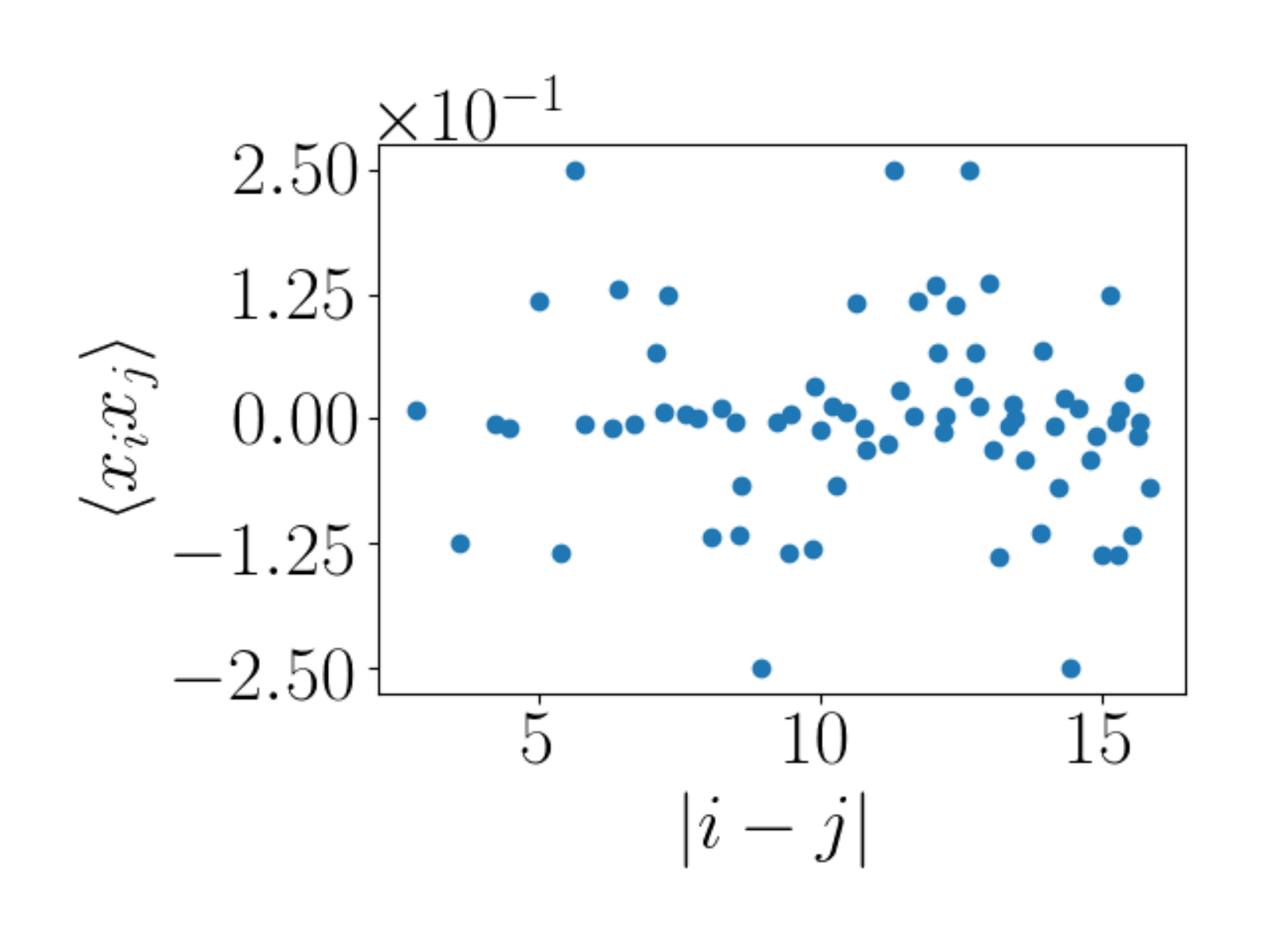}
\caption{$\corr{x_ix_j}$: sub-blocks of size 4.}
\label{fig:block_4_two_point}
    \end{subfigure}
    \begin{subfigure}{0.25\textwidth}
        \includegraphics[width=\textwidth]{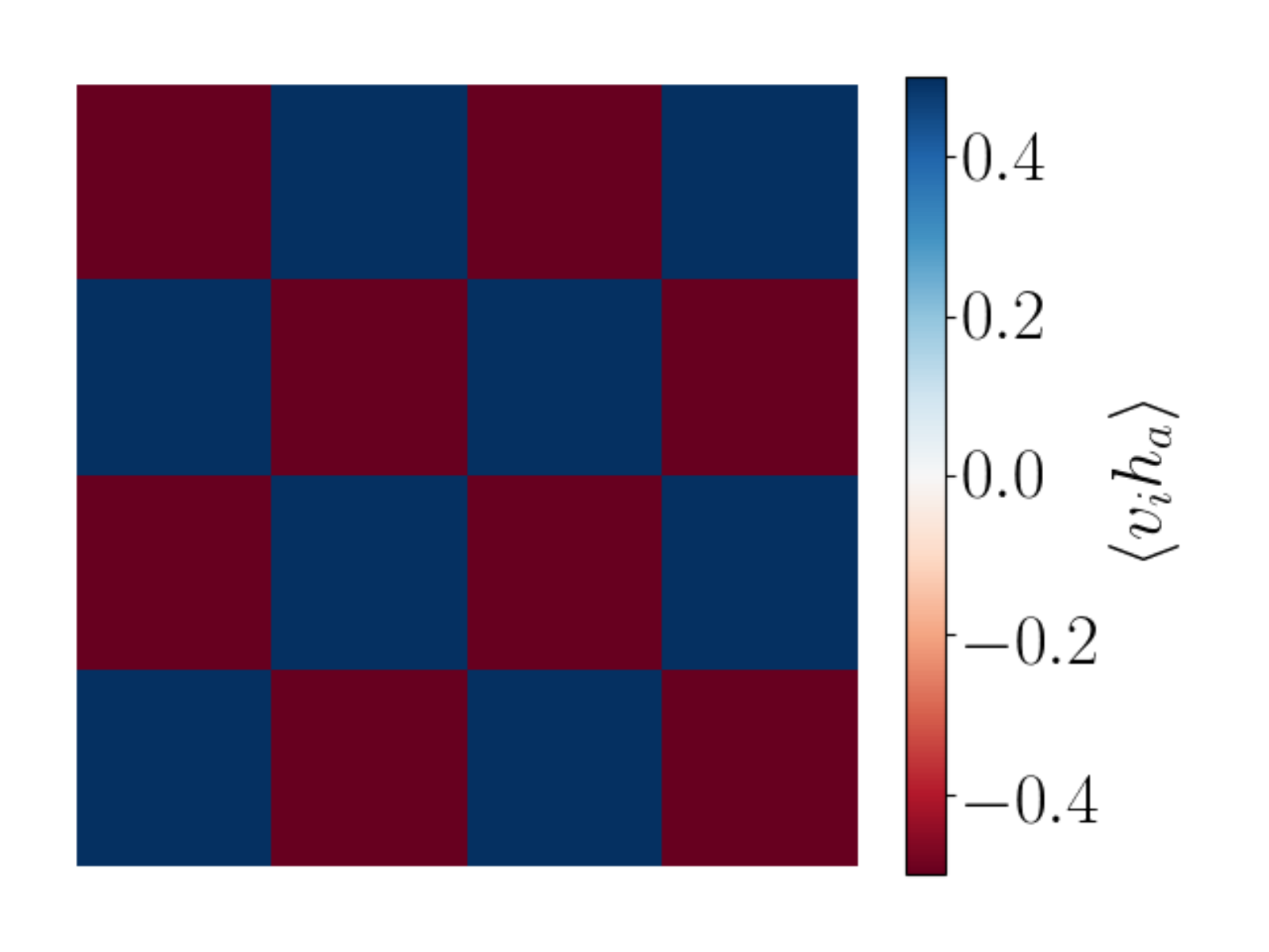}
        \caption{$\langle v_ih_a\rangle$ grid of $32 \times 32$ nodes with sub-blocks of size 8.}
        \label{fig:block_8}
            \end{subfigure}~~
    \begin{subfigure}{0.245\textwidth}
\includegraphics[width=\textwidth]{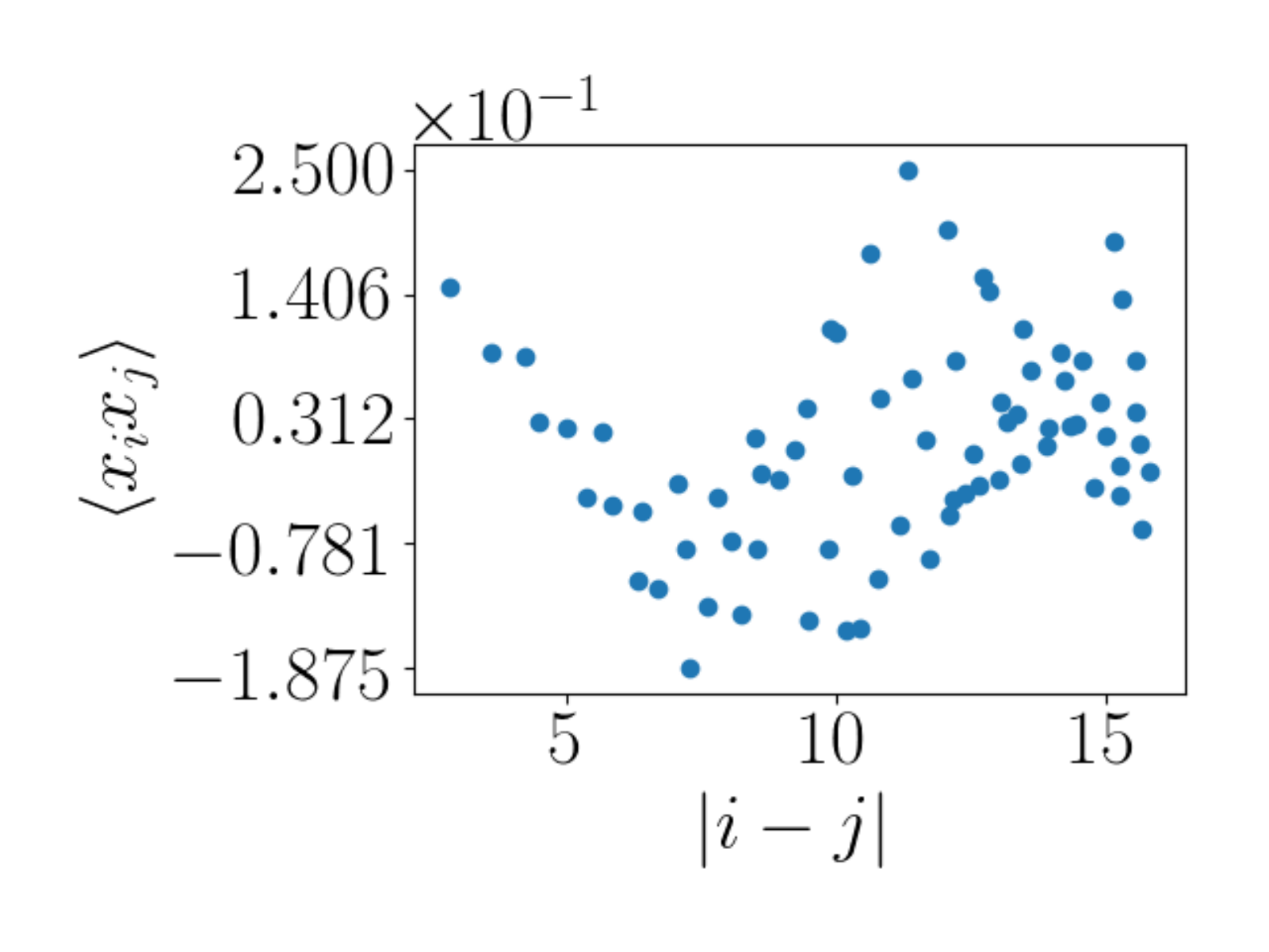}
\caption{$\corr{x_ix_j}$: sub-blocks of size 8.}
\label{fig:block_8_two_point}
    \end{subfigure}
    \begin{subfigure}{0.245\textwidth}
        \includegraphics[width=\textwidth]{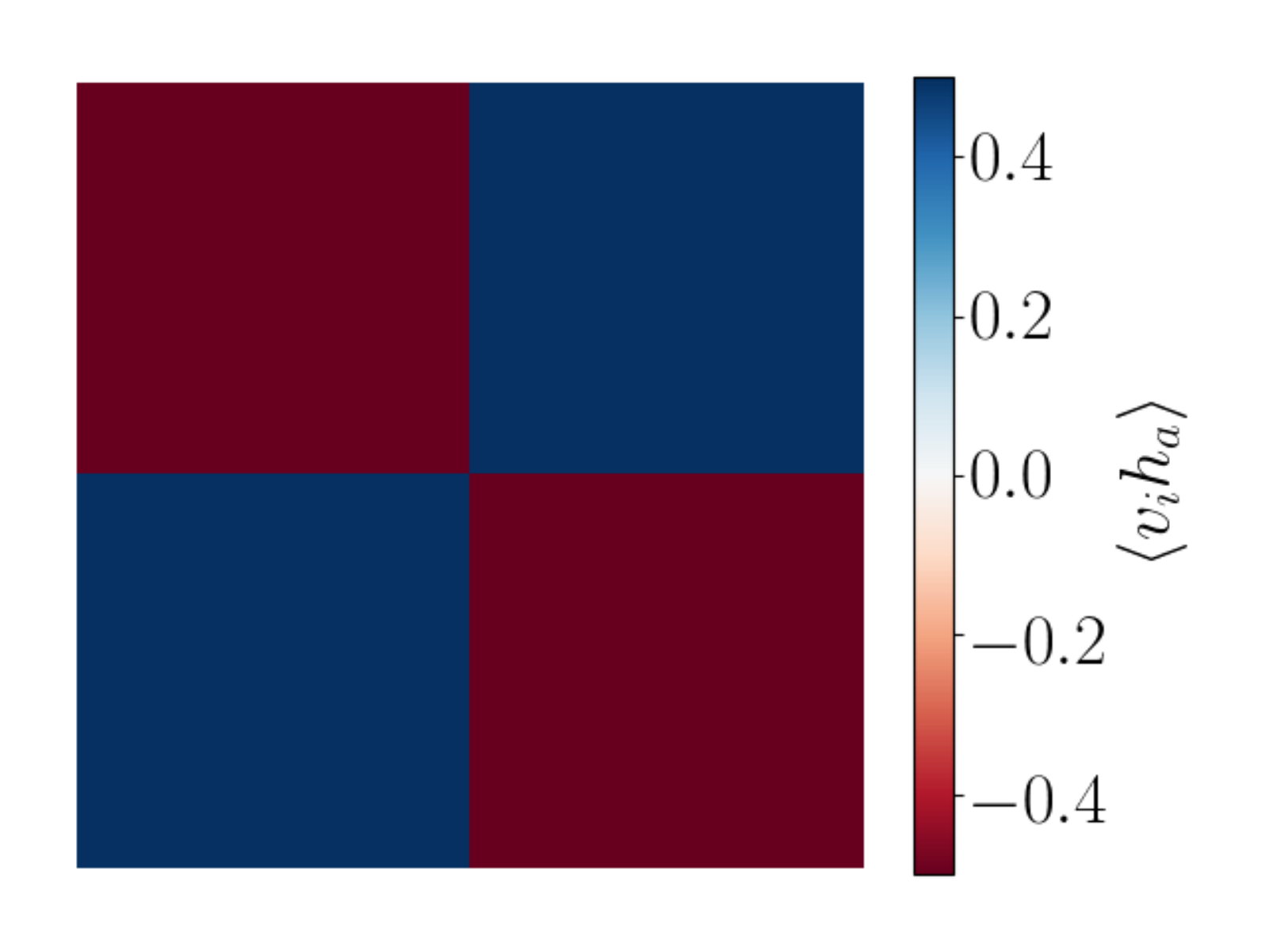}
        \caption{$\langle v_ih_a\rangle$ grid of $32 \times 32$ nodes with sub-blocks of size 16.}
        \label{fig:block_16}
            \end{subfigure}~~
    \begin{subfigure}{0.245\textwidth}
\includegraphics[width=\textwidth]{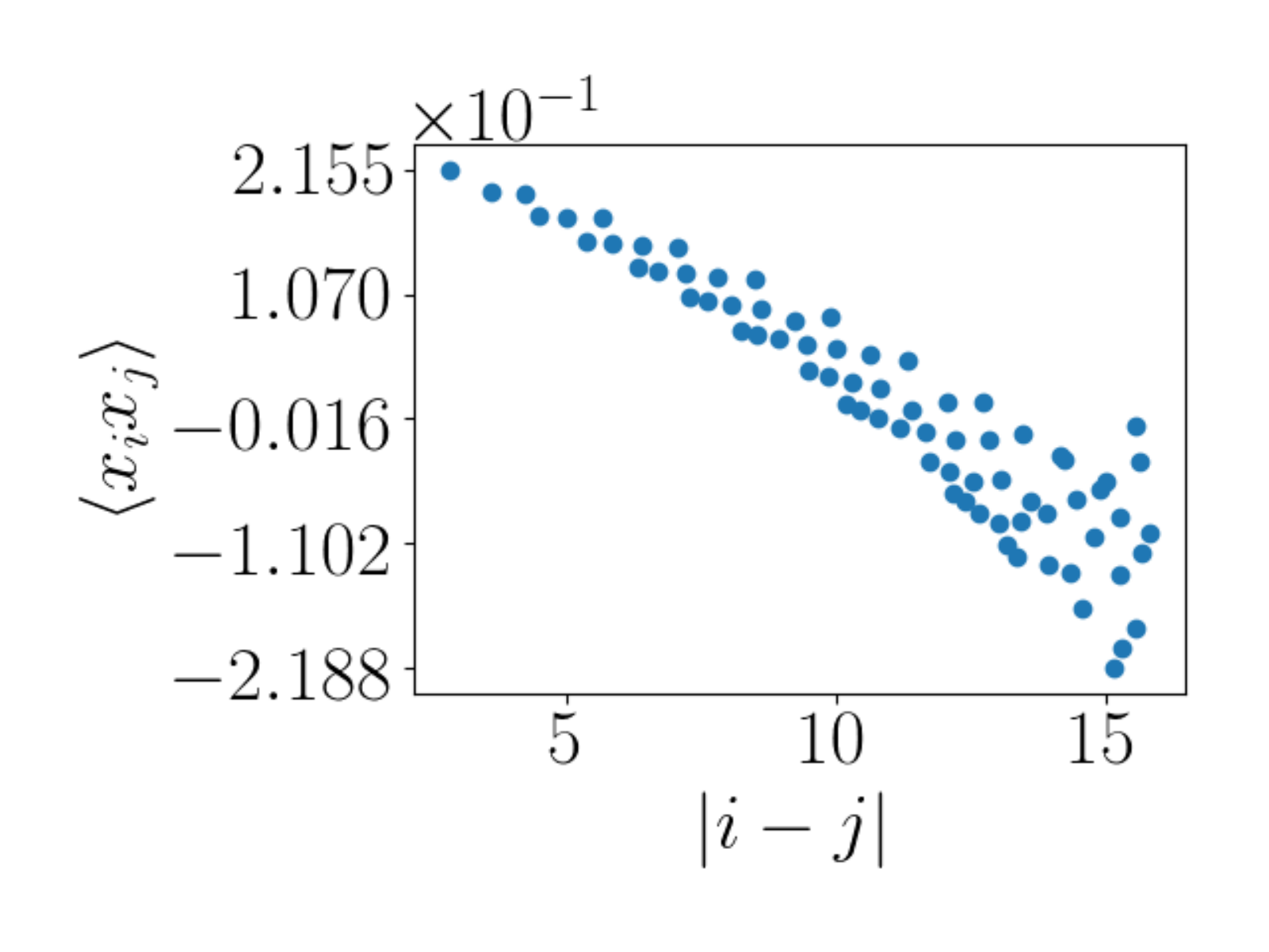}
\caption{$\corr{x_ix_j}$: sub-blocks of size 16.}
\label{fig:block_16_two_point}
    \end{subfigure}
    \caption{Checkerboard: Plots showing $\langle v_ih_a\rangle$ correlators generated to depict a checkerboard with varying block sizes as well as the two point correlator $\langle x_i x_j \rangle$ corresponding to the given $\langle v_ih_a\rangle$ plots. Plot (b) corresponds to plot (a), plot (d) corresponds to plot (c) and plot (f) corresponds to plot (e).}
    \label{fig:checkerboard}
\end{figure}
}

\begin{figure}[t!]
    \centering
\renewcommand{\thesubfigure}{a}
\subcaptionbox{\label{fig:rg_temps}}{\includegraphics[trim={0 0 0 0},clip,width=0.23\textwidth]{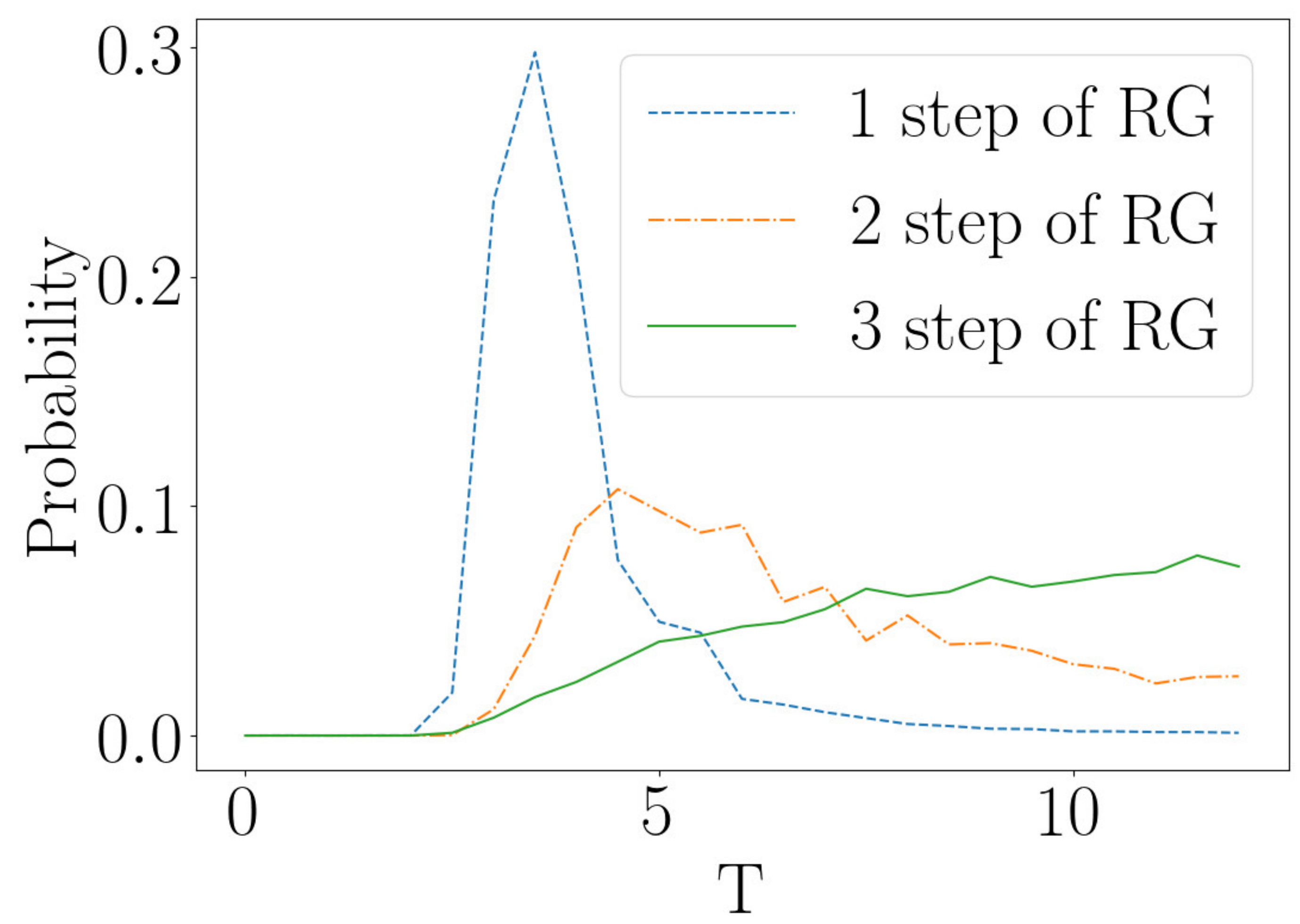}}~
\renewcommand{\thesubfigure}{b-i}
\subcaptionbox{\label{fig:rbm_temps_tc}}{\includegraphics[trim={0 0 0 0},clip,width=0.23\textwidth]{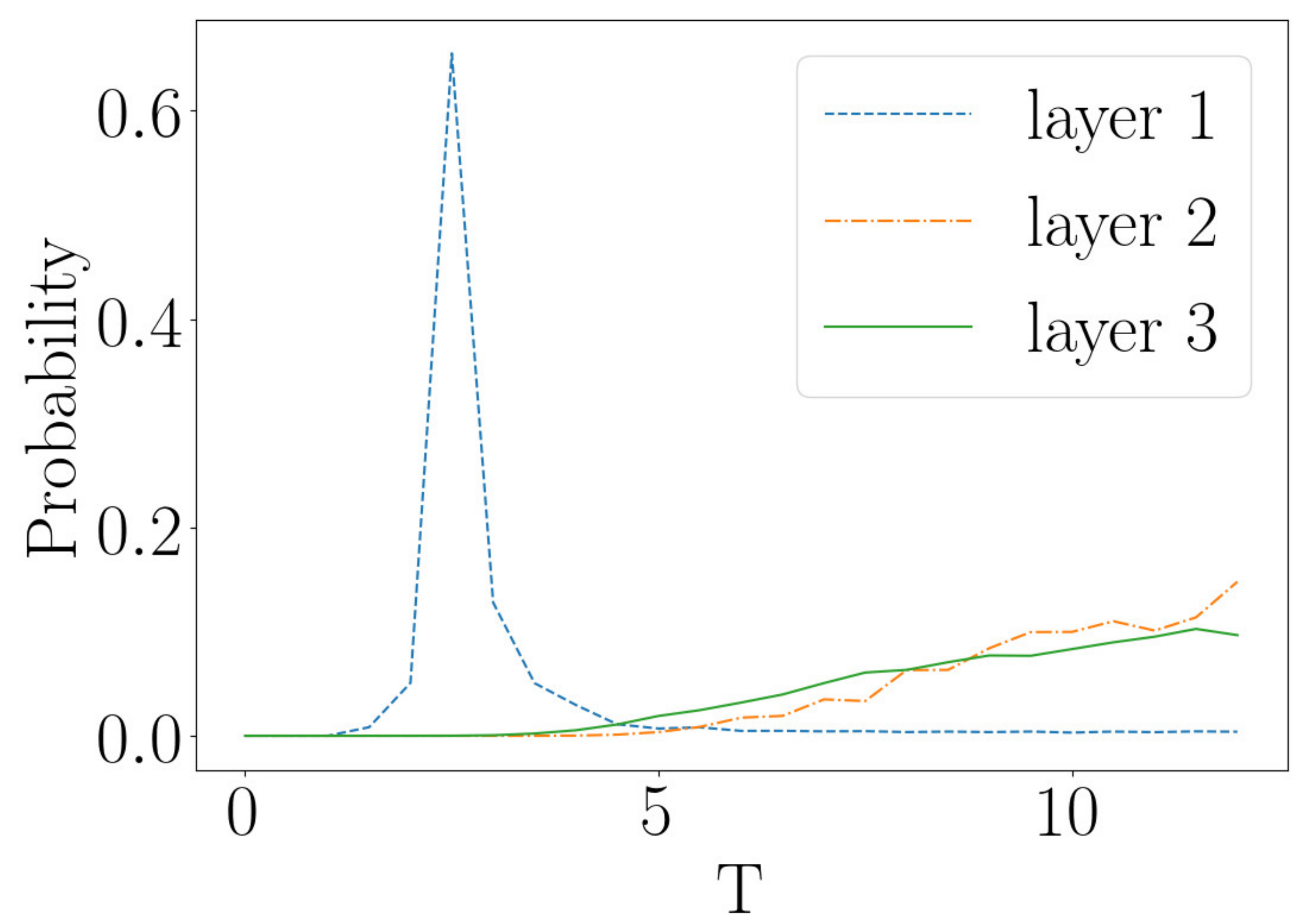}}
\renewcommand{\thesubfigure}{b-ii}
\subcaptionbox{\label{fig:rbm_temps_low}}{\includegraphics[trim={0 0 0 0},clip,width=0.23\textwidth]{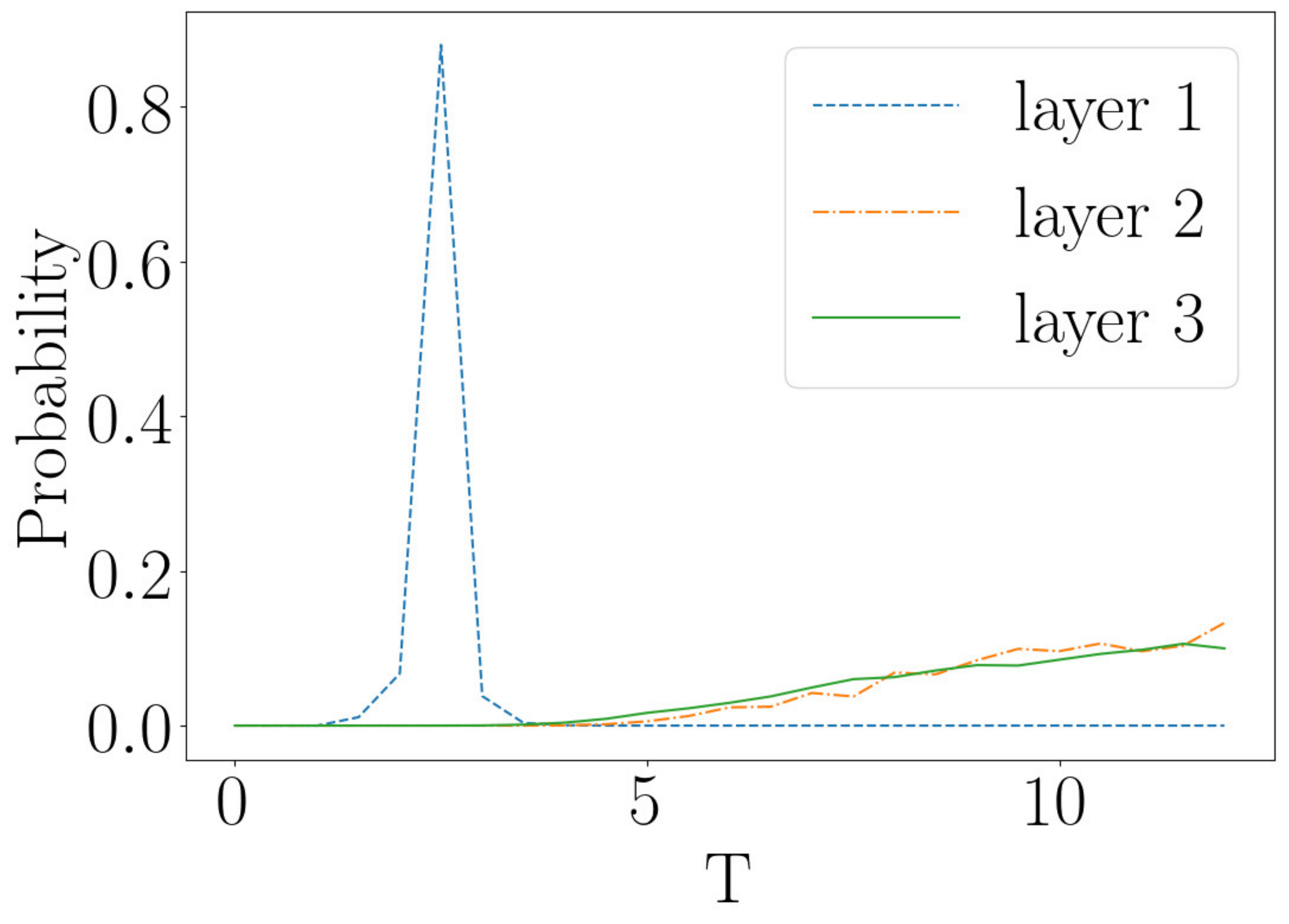}}~
\renewcommand{\thesubfigure}{b-iii}
\subcaptionbox{\label{fig:rbm_temps_high}}{\includegraphics[trim={0 0 0 0},clip,width=0.23\textwidth]{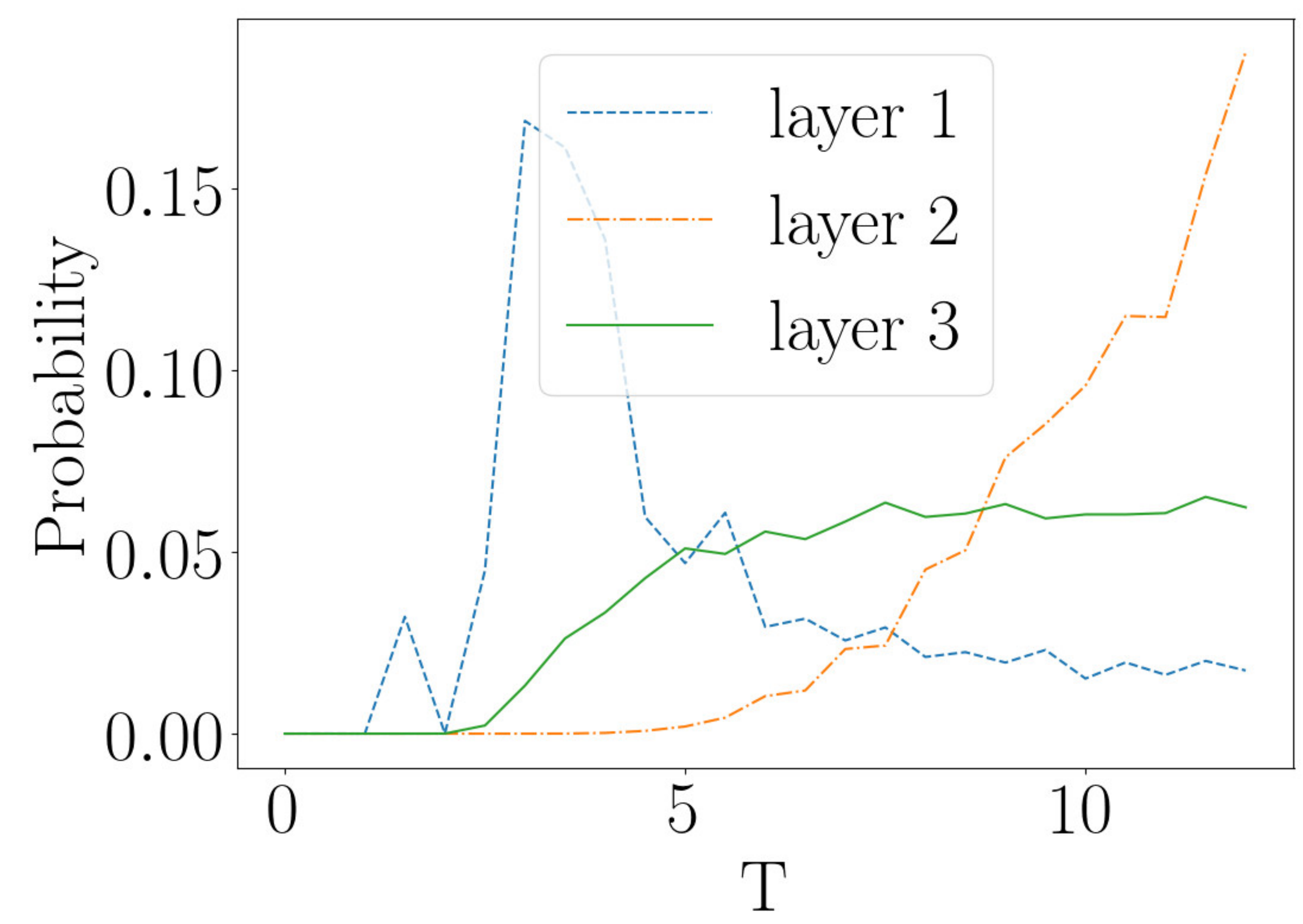}}
\caption{(a) shows the average probability of the measured temperature of lattices resulting after 3 steps of RG, applied to an input lattice at $T_c$ with 4096 sites. (b) shows the average probability plot of the measured temperature of outputs produced by a stacked RBM with 4096 input nodes, 1024 nodes in the first layer, 256 nodes in the second layer and 64 nodes in the output layer. (b-i) is given input Ising samples at $T=2.269$, (b-ii) is given input Ising samples at $T=2$ and (b-iii) is given input Ising samples at $T=2.7$.}
\label{fig:rbm_rg_temp_layers}
\end{figure}

There is one more interesting comparison that can be carried out and it quantitatively tests the flow.
The temperature is a relevant coupling so it grows as the flow proceeds.
In the block spin RG that we are considering, the length of the lattice keeps halving.
{ Thus, after 7 steps our unit of length is $2^7=128\approx 100$ times larger than it was.
To get some insight into the effect of this change of units, imagine we change units from centimeters to meters.
In the new units, a length of $100$cm is now 1m.
Anything with the units of length will roughly halve with each step of the flow.} 
In contrast to this, the temperature of the system, which in suitable units has a dimension of inverse length, will 
roughly double.
There will be small departures from precise doubling due to interactions, but the temperature must increase
by roughly a factor of 2 as each new layer is stacked.
If the RBM is performing an RG-like coarse graining, the temperature should grow in a similar way as we pass through 
the layers of the deep network.
Figure \ref{fig:rg_temps} plots the temperature of coarse grained lattices, generated by applying three steps of RG to an 
input lattice of size 64 by 64, at a temperature of $T=2.7$ .
There is a clear increase in the measured temperature as the number of RG steps increase.
The temperature of each layer is roughly $T=2.3$, $4.8$ and $11$ for layers 1, 2 and 3 respectively, which is indeed 
consistent with the rough rule that the temperature doubles with each step.

Now consider a deep network made by stacking three RBMs.
The first network has 4096 visible nodes and 1024 hidden nodes, the second 1024 visible nodes and 256 hidden nodes 
and the third 256 visible nodes and 64 hidden nodes.
The network is trained on Ising data at the critical temperature, as described above.
Figures \ref{fig:rbm_temps_tc}, \ref{fig:rbm_temps_low} and \ref{fig:rbm_temps_high} give the temperatures of the 
outputs of the layers of the RBM, given input lattices at temperatures of $T=2.269$, $2$ and $2.7$ respectively.
Temperatures of $T=2$ and $T=2.269$ lead to the same behavior for the temperature flow, as exhibited in 
Figures \ref{fig:rbm_temps_tc} and \ref{fig:rbm_temps_low}.
The temperature jumps rapidly to a high temperature in the first step of the flow, and remains fixed when the second step
is taken.   
This is an important difference that deserves to be understood better. 
It questions the identification of layers of a deep network with steps in an RG flow. 

Figure \ref{fig:rbm_temps_high} shows different characteristics to those of \ref{fig:rbm_temps_tc} and \ref{fig:rbm_temps_low}. 
Here the temperature of the input is above $T_c$ at $2.7$.
Layer 1 is not as sharply peaked near $T_c$ as observed in Figures \ref{fig:rbm_temps_tc} and \ref{fig:rbm_temps_low}.
In addition to this, layers 2 and 3 are not at the same temperature but rather layer 2 is at a higher temperature than layer 3.
This differs to the RG flow, where temperature increases along the flow.
Figure \ref{fig:rbm_temps_high} shows a decrease in temperature from layer 2 to layer 3 rather than an increase.
These plots demonstrate that the flow defined by multiple layers in a ``deep'' network show important differences to the RG flow.
The discrepancies we have uncovered are important and precise quantitative mismatches that may provide useful clues in
understanding the relationship between unsupervised deep learning by an RBM and the RG flows.

%


The results above have shown that the correlator $\corr{v_ih_a}$ exhibits RG-like characteristics. 
This is evident from the comparison between the $\corr{v_ih_a}$ plots from RG, a stacked RBM network and a network with a single RBM. 
We can see RG-like patterns in the correlators produced by the two RBM networks. 
This is a promising result that demonstrates that a form of coarse graining is taking place when networks are stacked.

\section{Conclusions and Discussion}\label{Conc}

Our main goal has been to explore the possibility that RG provides a framework within which a theoretical understanding
of deep learning can be pursued. 
{ We have focused on a single model, the Ising model, which is naturally related to RBMs. Thus, at best our
conclusions and discussion can only suggest interesting avenues for further study. We are not able to draw general definite
conclusions about the applicability of RG as a framework within which a theoretical understanding of deep learning can be
achieved.}
Our data set contains the possible states of an Ising magnet, generated using Monte Carlo simulation.
This is an interesting data set, since we know that there is a well defined theory for the magnet defined on large length scales.
The existence of this long distance theory guarantees that there is some emergent order for the unsupervised learning to
identify.
Another point worth stressing is that the RG treatment of this system is well understood and is easily implemented
numerically.
It is therefore an ideal setting in which both deep learning and RG can be implemented and their results can be compared.
At the critical temperature, where the system is on the verge of spontaneous magnetization, there is an interesting scale
invariant theory which is well understood.
By working at this critical point, we have managed to probe the patterns generated by the RBM at different length scales and to
compare it to the expected results from an RG treatment.
 
Our first set of numerical results compare the RBM flow introduced in \cite{iso2018scale} and further pursued in \cite{funai2018thermodynamics}.
From a theoretical point of view the RBM flow looks rather different to RG since the RBM flow appears to drive configurations towards
the critical temperature.
The RG would drive configurations to ever higher temperatures due to the fact that the temperature corresponds to a
relevant perturbation.
Another important difference between the RBM flow and RG is that the number of spins is a constant of the RBM flow, but
decreases with the RG flow.
Our numerical results confirm that the RBM flow does indeed generate RG-like Ising configurations and we have reproduced
the scaling dimension of the spin variable from the spatial statistics of the patterns generated by the RBM.
This is a remarkable result and it extends and supports results reported and discussed in \cite{iso2018scale,funai2018thermodynamics}.
The spin variable has the smallest possible scaling dimensions and consequently probes the largest possible scales 
in the pattern.
When considering correlation functions of the next primary operators we find that the RBM data does not reproduce the
correct scaling dimension, proving that the spatial statistics of the patterns generated by the RBM flow and those generated
by RG start to differ as smaller scales are tested.
We therefore conclude that the RBM flow and RG are distinct, but they do agree on the largest scale structure of the 
generated patterns.
This is a hint into the mechanism behind the RBM flow and it deserves an explanation.

Our second numerical study has explored the idea that deep learning is an RG flow with each stacked layer performing a
step of RG.
We have explained why correlation functions between the visible and hidden neurons, $\corr{v_ih_a}$ are capable of
diagnosing RG-like coarse graining and we have computed these correlation functions using the patterns generated by the RBM.
The basic signal of RG coarse graining is a ``bright spot'' in the $\corr{v_ih_a}$ correlation function, since this indicates that spins in a localized region were averaged to produce the coarse grained spin.
The numerical results do indeed show a dark background with emerging bright spots.
It would be interesting if the emergent patterns again guarantee agreement on the largest length scales, similar to what was
found for the RBM flows, but we can not confidently make this assertion yet.

Our final numerical study considered the flow of the temperature, a relevant operator according to the RG.
We find three distinct behaviors.
Section \ref{sec:rbm_flows_numeric} reviewed that RBM flows converge to the critical temperature.
This is borne out in our results.
The RG flows to ever higher temperatures, with (roughly) a doubling in temperature for each step.
Again, this is precisely what we observe.
Finally, for a deep network made by stacking three RBMs, the temperature appears to flow when moving between 
the first and second layers of a deep network, but is fixed when moving between the second and third layers.
This is an important difference that deserves to be understood better. 
It questions the identification of layers of a deep network with steps in an RG flow.
 
Our results are encouraging.
There are enough similarities between unsupervised learning by an RBM and the RG flow that the relationship
between the two should be developed further.
Regarding future studies, it maybe useful to explore models other than Ising.
The Ising model has an unstable fixed point due to the presence of relevant operators.
Consequently, finite flows starting near the critical point all terminate on different models.
In this case its not easy to know if the RBM has flowed to the ``right answer'' because there are many possible right answers!
The stable fixed point of the model is at infinite temperature and the configurations at this fixed point are random with
correlators that have a correlation length of zero.
This is hardly a promising answer to shoot for.
It maybe more instructive to study models that have an attractive RG fixed point.
In this case the minimum that the RBM is looking for would be unique and the connection between the two may be easier
to recognize.
{
We have in mind systems that exhibit self organized criticality \cite{diaz1994dynamic}, 
including models constructed to understand the spread of 
forest fires \cite{loreto1995renormalization} and models for the spread of 
infectious diseases \cite{clar1996forest}.}

By using Ising model data, generated by Monte Carlo simulation, starting from a local Hamiltonian we know
how a coarse graining capable of identifying emergent patterns should proceed: spatially neighboring
spins should be averaged.
For more general data sets, this may not be the case.
It is fascinating to ask what the rules determining the correct coarse graining are and in fact, with respect to this question,
deep learning has the potential to shed light on RG.

{ Another interesting comparison worth mentioning is the similarity between an average pooling layer within a convolutional neural network (CNN) and the averaging performed in variational RG.  
CNNs are known for their excellent performance in image recognition and classification tasks \cite{krizhevsky2012imagenet,huang2017densely,khan2019survey}.
CNNs have a number of layers which act on groups of nearby pixels in the image. 
One of these layers which is similar to the coarse graining performed in variational RG is called a pooling layer.
The pooling layer performs a down-sampling on the data it receives from previous layers in the network \cite{lee2016generalizing,yu2014mixed,zhou2016learning}.
The down-sampled data is more robust to changes in position of features and gives the network the property of local translation invariance.
One way in which the pooling operation is implemented is by averaging all values in the given patch of data it acts on to obtain a new value to replace these values.
Pooling usually averages blocks of data which are of size $2\times 2$.
This results in an input block of data being reduced by a factor of 2 in length and by a factor of 4 in the number of values which is the same factor of rescaling which occurs in variational RG.}

{ In recent years a connection between the renormalization group and tensor networks \cite{vidal2007entanglement} 
has been discovered, providing a connection to the field of quantum information. 
The discovered connection demonstrates that the multi-scale entanglement renormalization ansatz (MERA) tensor networks
carry out a coarse graining that agrees in many ways with the coarse graining performed by the renormalization group \cite{evenbly2015tensor} .
This suggests that there maybe a link between tensor networks and deep learning.
For related ideas see \cite{cichocki2016tensor,cichocki2017tensor}.
Since tensor networks have been extensively studied for calculations the connection may prove to be useful for better
understanding deep learning.}

Apart from the exciting possibility that the link to RG might contribute towards a theoretical understanding of deep learning,
one might also ask if the connection would have any practical applications.
One possibility that we are currently pursuing, is a Callan-Symanzik like equation governing the learning process.
Roughly speaking, one might mimic RG by dividing the weights to be learned into relevant, marginal and irrelevant
parameters, depending on gross statistical properties of the training data.
If this classification is itself not too expensive, one could pursue a more efficient approach towards training, since the
classification of weights would provide an understanding of which weights are important, and which can simply be set to zero. 
We hope to report on this possibility in the future.

\section*{Acknowledgement}

This work is supported by the South African Research Chairs
Initiative of the Department of Science and Technology and National Research Foundation
as well as funds received from the National Institute for Theoretical Physics (NITheP).
We are grateful for useful discussions to Mitchell Cox and Dimitrios Giataganas.

\appendices
\section{RBM expectation values}\label{EVs}
The expectation values quoted in equations \eqref{eq:dklw}, \eqref{eq:dklbv} and \eqref{eq:dklbh} are derived 
using \eqref{eq:model}.
Data expectation values are evaluated by summing over all samples, $\hat{v_i}^{(A)}$ in the training set. 
On the other hand model expectation values employ sums over the entire space of visible and hidden vectors.
This is such an enormous sum that its numerically intractable.
Consequently, the approximations described in Section \ref{RBM} are used.
The complete set of expectation values needed to describe the RBM are given by
\bea
\corr{v_ih_a}_{data}=\frac{1}{N_s}\sum_{A=1}^{N_s}\hat{v_i}^{(A)}\tanh\left(\sum_kW_{{k}a}\hat{v_{k}}^{(A)}+b_a^{(h)}\right)\cr,
\eea
\be
\begin{split}
\corr{v_ih_a}_{model}=&\sum_{\{{\bf v},{\bf h}\}}
\tanh\left(\sum_{{k}}W_{{k}a}v_{k}+b_a^{(h)}\right)\cdot\\
&\tanh\left(\sum_{j}W_{i{j}}h_{j}+ +b_i^{(v)}\right),
\end{split}
\ee
\bea
\corr{v_i}_{data}=\frac{1}{N_s}\sum_{A=1}^{N_s}\hat{v_i}^{(A)},
\eea
\bea
\corr{v_i}_{model}=\sum_{\{\bf h\}}\tanh\left(\sum_aW_{ia}h_a+b_i^{(v)}\right),
\eea
\bea
\corr{h_a}_{data}=\frac{1}{N_s}\sum_{A=1}^{N_s}\tanh\left(\sum_iW_{ia}\hat{v_i}^{(A)}+b_a^{(h)}\right),
\eea
\bea
\corr{h_a}_{model}=\sum_{\{\bf v\}}\tanh\left(\sum_iW_{ia}v_i+b_a^{(h)}\right),
\eea
with $\hat{v}_i^{(A)}$ the $A$th sample of the data set, $\bf{\hat{v}}$.

\section{Two Versions of RG}\label{varrg}

In this section we review two versions of the RG that are needed in this article.
The first of these, the variational renormalization group, was introduced by Kadanoff \cite{kadanoff2000statistical,kadanoff1976variational,efrati2014real}
 as a method to 
approximately perform the renormalization group in practice.

\subsection{Variational RG}\label{VarRG}

Consider a system of $N$ spins $\{v_i\}$ which each take the values $\pm 1$. 
The partition function describing the system is given by
\bea
Z = \sum_{v_i} e^{-H(\{v_i \})}.
\eea
Here the sum is over all possible configurations of the system of spins and the function $H(\{v_i\})$, called the
Hamiltonian, gives the energy of the system.
This would include the energy of each individual spin as well as the energy associated to the fact that the collection
of spins is interacting.
The Hamiltonian $H(\{v_i\})$ can be an arbitrarily complicated function of the spins
\bea
H(\{v_i\}) =-\sum_i K_iv_i-\sum_{i,j}K_{ij}v_iv_j\cr
-\sum_{i,j,k} K_{ijk}v_iv_jv_k +\cdots.
\eea
The RG flows maps the original Hamiltonian to a new Hamiltonian with a different set of coupling constants.
The new Hamiltonian 
\bea
H(\{h_a\}) =-\sum_a K_a' h_a-\sum_{a,b}K'_{ab}h_ah_b\cr
-\sum_{a,b,c} K'_{abc}h_ah_bh_c +\cdots,
\eea
gives the energy for the coarse grained spins $h_a$.
After many RG iterations many coupling constants (the so called irrelevant terms) flow to zero.
A much smaller number may remain constant (marginal terms) or even grow (relevant terms).
To implement this conceptual framework a concrete RG mapping is needed. 
Variational RG provides a mapping which is not exact but can be implemented numerically. 
It does this by introducing an operator $T_\lambda(\{v_i, h_a\})$ which is a function of a set of parameters 
$\{\lambda\}$. 
The Hamiltonian after a step of RG flow is
\bea
e^{-H_{RG}(\{h_a\})} = \sum_{v_i} e^{T_\lambda(\{v_i,h_a\})-H(\{v_i\})}.
\eea
The form of $T_\lambda (\{v_i,h_a\})$ must be chosen cleverly, for each problem we consider.
This is the tough step in variational RG and it is carried out using physical intuition, but essentially on a trial and
error basis.
Once a given $T_\lambda (\{v_i,h_a\})$ has been chosen, we minimize the following quantity by choosing the parameters
$\{\lambda\}$
\bea
\log( \sum_{v_i} e^{-H(\{v_i\})})-\log(\sum_{h_a} e^{-H_{RG}(\{h_a\})}).
\label{target}
\eea
The minimum possible value for this quantity is zero.
Notice that when 
\bea
\sum_{h_a}e^{T_\lambda(\{v_i,h_a\})}=1,
\label{eq:varrg_sumha}
\eea
(\ref{target}) attains its minimum value of 0 and the RG transformation is called exact.

\subsection{Block Spin Averaging}\label{BSA}

Block spin averaging is a pedagogical version of RG.
To illustrate the method, consider a rectangular lattice of interacting spins.
Divide the lattice into blocks of $2\times 2$ squares.
Block spin averaging describes the system in terms of {\it block variables}, which are variables describing the average 
behavior of each block. 
The ``block spin'' is literally the average of the four spins in the block.
The plots shown in Figures \ref{fig:a-i-rg} use block spin averaging.
The block spins $h_a$ are each an average of four visible spins $v_i$. 

\bibliographystyle{unsrt}

\begin{IEEEbiography}[{\includegraphics[width=1in,height=1.25in,clip,keepaspectratio]{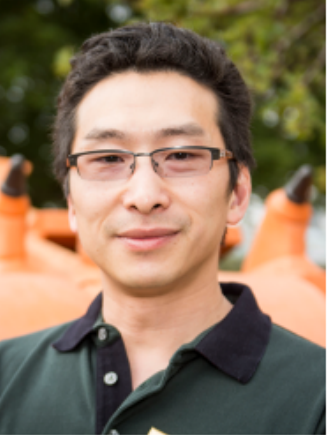}}]{Ling Cheng}

(M’10-SM’15) received the degree B. Eng. Electronics and Information (cum laude) from Huazhong University of Science and Technology (HUST) in 1995, M. Ing. Electrical and Electronics (cum laude) in 2005, and D. Ing. Electrical and Electronics in 2011 from University of Johannesburg (UJ). His research interests are in Telecommunications and Artificial Intelligence. In 2010, he joined University of the Witwatersrand where he was promoted to Associate Professor in 2015. He has served as the Vice-chair of IEEE South African Information Theory Chapter. He has been a visiting professor at five universities and the principal advisor for over forty full research post-graduate students. He has published more than 80 research papers in journals and conference proceedings. He was awarded the Chancellor’s medals in 2005, 2019 and the National Research Foundation rating in 2014. The IEEE ISPLC 2015 best student paper award was made to his Ph.D. student in Austin.\end{IEEEbiography}

\begin{IEEEbiography}[{\includegraphics[width=1in,height=1.25in,clip,keepaspectratio]{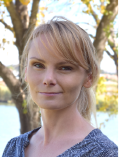}}]{Ellen de Mello Koch}
obtained her BSc(Eng) degree in 2014 and MSc(Eng) degree in 2016 in Electrical Engineering at the University of the Witwatersrand.
She is currently a Ph.D. student and lecturer in the department of Electrical and Information Engineering.
Her doctoral research investigates the link between deep learning and the renormalization group, as an attempt to develop a theoretical framework for deep learning.
Her research interests lie in unsupervised learning algorithms and their application to the real world.
\end{IEEEbiography}


\begin{IEEEbiography}[{\includegraphics[width=1in,height=1.25in,clip,keepaspectratio]{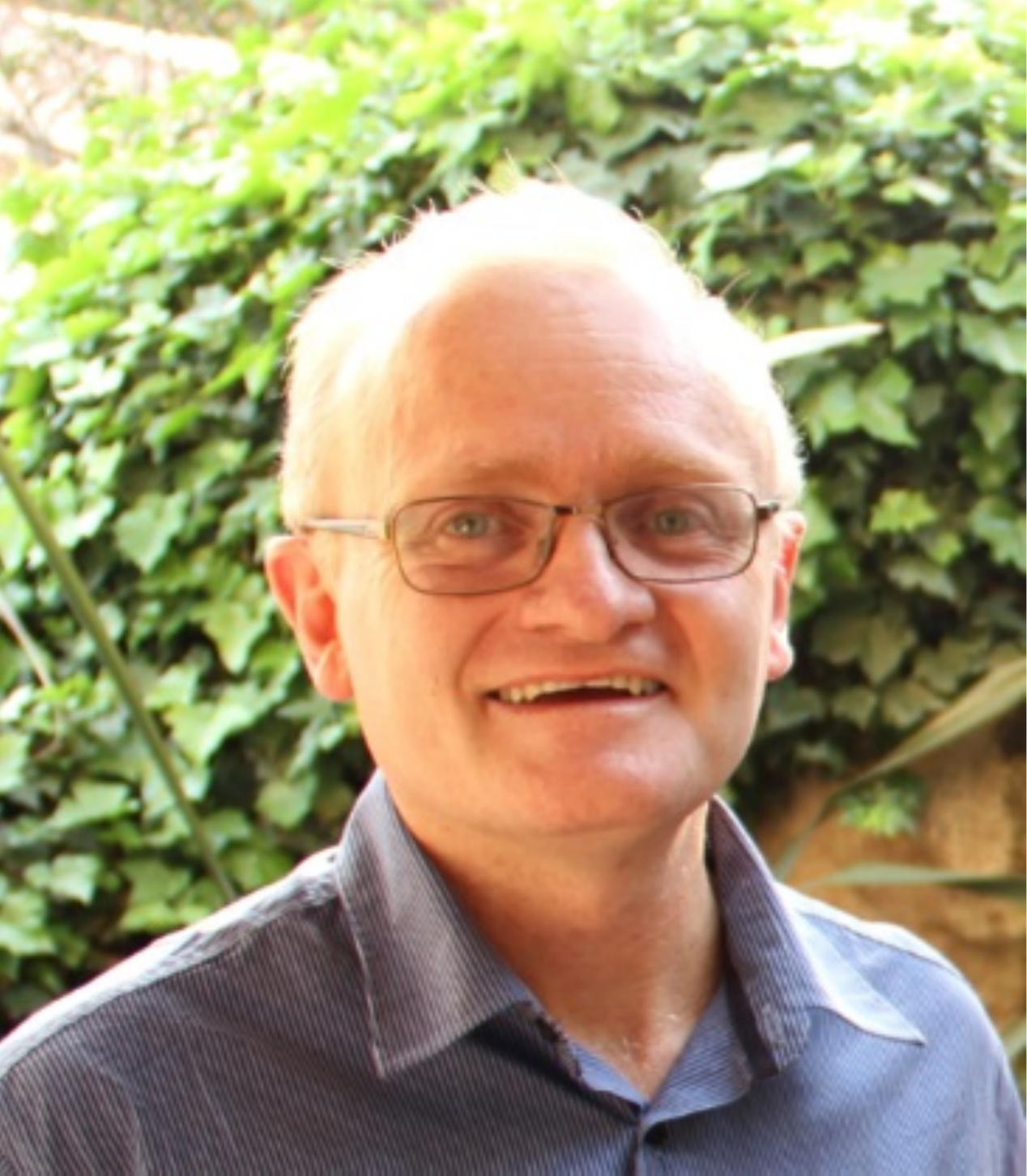}}]{Robert de Mello Koch}
obtained his BSc(Eng) degree in 1992, his BscHon(Phys) in 1993, his MSc(Phys) in 1994 and his PhD (Phys) in 1998,
all from the University of the Witwatersrand. 
He is a Professor at the University of the Witwatersrand where he holds the DST/NRF Research Chair in Fundamental
Physics and String Theory and is a Distinguished Visiting Professor at South China Normal University. He is a fellow
of the Durham Institute for Advanced Studies, the Stellenbosch Institute for Advanced Studies and the Academy of
Science of South Africa.
His most recent research interests include the gauge theory/string theory duality, the application of representation
theory of discrete groups and Lie groups to quantum field theory and deep learning.  
\end{IEEEbiography}

\EOD

\end{document}